\documentclass[
12pt, 
oneside, 
english, 
doublespacing, 
liststotoc, 
]{main} 

\usepackage[utf8]{inputenc} 
\usepackage[T1]{fontenc} 
\usepackage{palatino} 
\usepackage[backend=bibtex,style=numeric,natbib=true,sorting=none]{biblatex} 
\usepackage[autostyle=true]{csquotes} 
\usepackage{amsmath}
\usepackage{amssymb}
\usepackage{graphicx} 
\usepackage{times} 
\usepackage{amsmath} 
\usepackage{amssymb}  
\usepackage{bm}
\graphicspath{{./Figures/}}
\usepackage[caption=false,font=footnotesize]{subfig}
\usepackage{braket}
\usepackage{floatrow}
\usepackage{tikz}
\usepackage{floatrow}
\usepackage{svg}
\usepackage{physics}
\usepackage{pdfpages}

\usepackage{booktabs,ragged2e}
\usepackage[flushleft]{threeparttable}

\usepackage{xcolor}

\definecolor{orange}{rgb}{1.0, 0.22, 0.0}

\thesistitle{Development of a Novel Impedance-Controlled Quasi-Direct-Drive Robot Hand} 

\supervisor{Dr. Amin Fakhari}  
\examinerO{Dr. Nilanjan Chakraborty} 
\examinerTh{Dr. Shanshan Yao} 
\examinerJr{4th Examiner's Name} 
\degree{Master of Science} 
\author{Jay Best} 
\addresses{} 

\subject{Mechanical Engineering} 
\university{\href{http://www.stonybrook.edu/}{Stony Brook University}} 
\department{{Mechanical Engineering}} 
\group{{Your Lab}} 
\faculty{{Faculty name}} 

\hypersetup{pdftitle=\ttitle} 
\hypersetup{pdfauthor=\authorname} 
\hypersetup{pdfkeywords=\keywordnames} 

\usepackage{enumitem}

\begin{document}
\begin{singlespace}
\frontmatter 

\pagestyle{plain} 


\begin{titlepage}
\begin{center}


\HRule \\[0.4cm] 
{\huge \bfseries \ttitle\par}\vspace{0.4cm} 
\HRule \\[1.5cm] 
 
\large A Thesis Presented \\[0.5cm] 
by \\[0.5cm] 
\textbf{{\authorname}} \\[0.5cm] 
to \\[0.5cm] 
The Graduate School \\[0.5cm] 
in Partial Fulfillment of the \\[0.5cm]
Requirements \\[0.5cm]
for the Degree of \\[0.5cm] 
\textbf{ \degreename}\\[0.5cm] 
in\\[0.5cm]
\textbf{\deptname} \\[1cm]
\univname\\[1cm] 
 
\textbf{\large May 2023}\\[4cm] 

\vfill
\end{center}
\end{titlepage}


\pagenumbering{gobble}


\setcounter{page}{2}
\includepdf{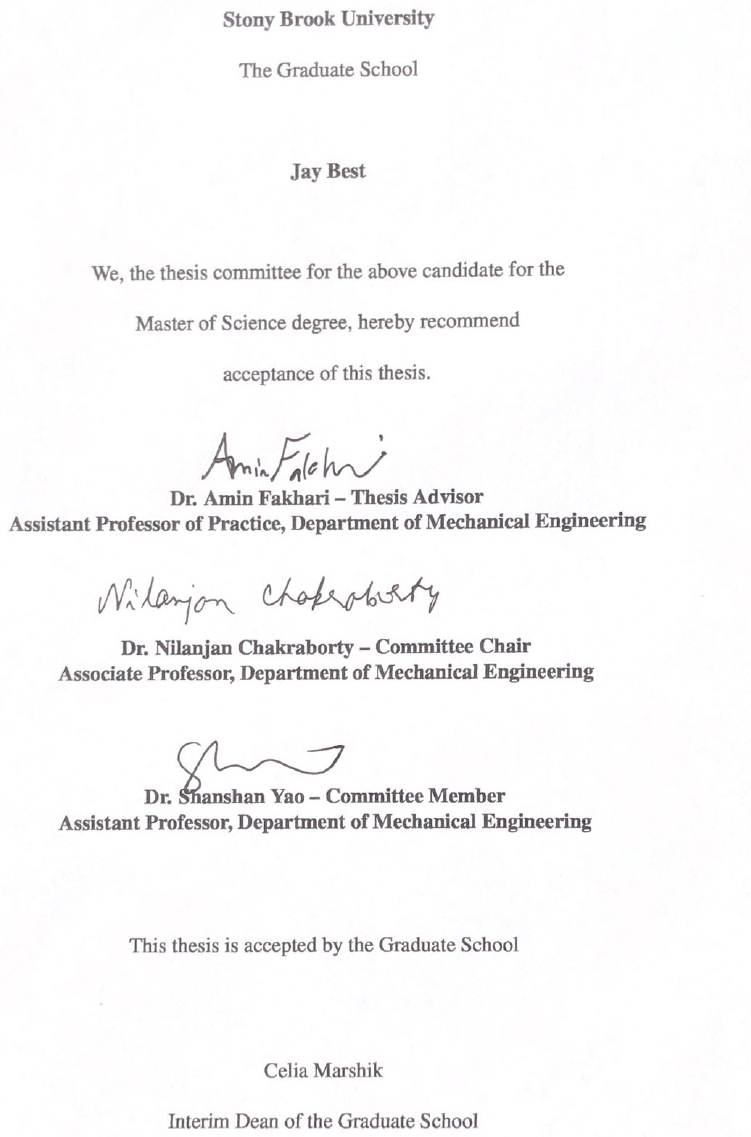}
	
	
	
	



\renewcommand{\abstractname}{Abstract of the Thesis}
\begin{center}
	\section*{\abstractname}

		\textbf{\ttitle} \\ [0.5cm]
		by  \\ [0.5cm]
		\textbf{\authorname} \\ [0.5cm]
		\textbf{\degreename} \\ [0.5cm]
		in \\ [0.5cm]
		\textbf{\deptname}  \\ [0.5cm]
		\univname  \\ [0.5cm]
		2023  \\ [0.5cm]
\end{center}

Most robotic hands and grippers rely on actuators with large gearboxes and force sensors for controlling gripping force. However, this might not be ideal for tasks which require the robot to interact with an unstructured and/or unknown environment. We propose a novel quasi-direct-drive two-fingered robotic hand with variable impedance control in the joint space and Cartesian space. The hand has a total of four degrees of freedom, a backdrivable gear train, and four brushless direct current (BLDC) motors. Field-Oriented Control (FOC) with current sensing is used to control motor torques. Variable impedance control allows the hand to perform dexterous manipulation tasks while being safe during human-robot interaction. The quasi-direct-drive actuators enable the fingers to handle contact with the environment without the need for complicated tactile or force sensors. A majority 3D printed assembly makes this a low-cost research platform built with affordable off-the-shelf components. The hand demonstrates grasping with force-closure and form-closure, stable grasps in response to disturbances, tasks exploiting contact with the environment, simple in-hand manipulation, and a light touch for handling fragile objects.
\end{singlespace}
\newpage



\vspace*{0.125\textheight}
\begin{center}
\includegraphics[scale = 0.12]{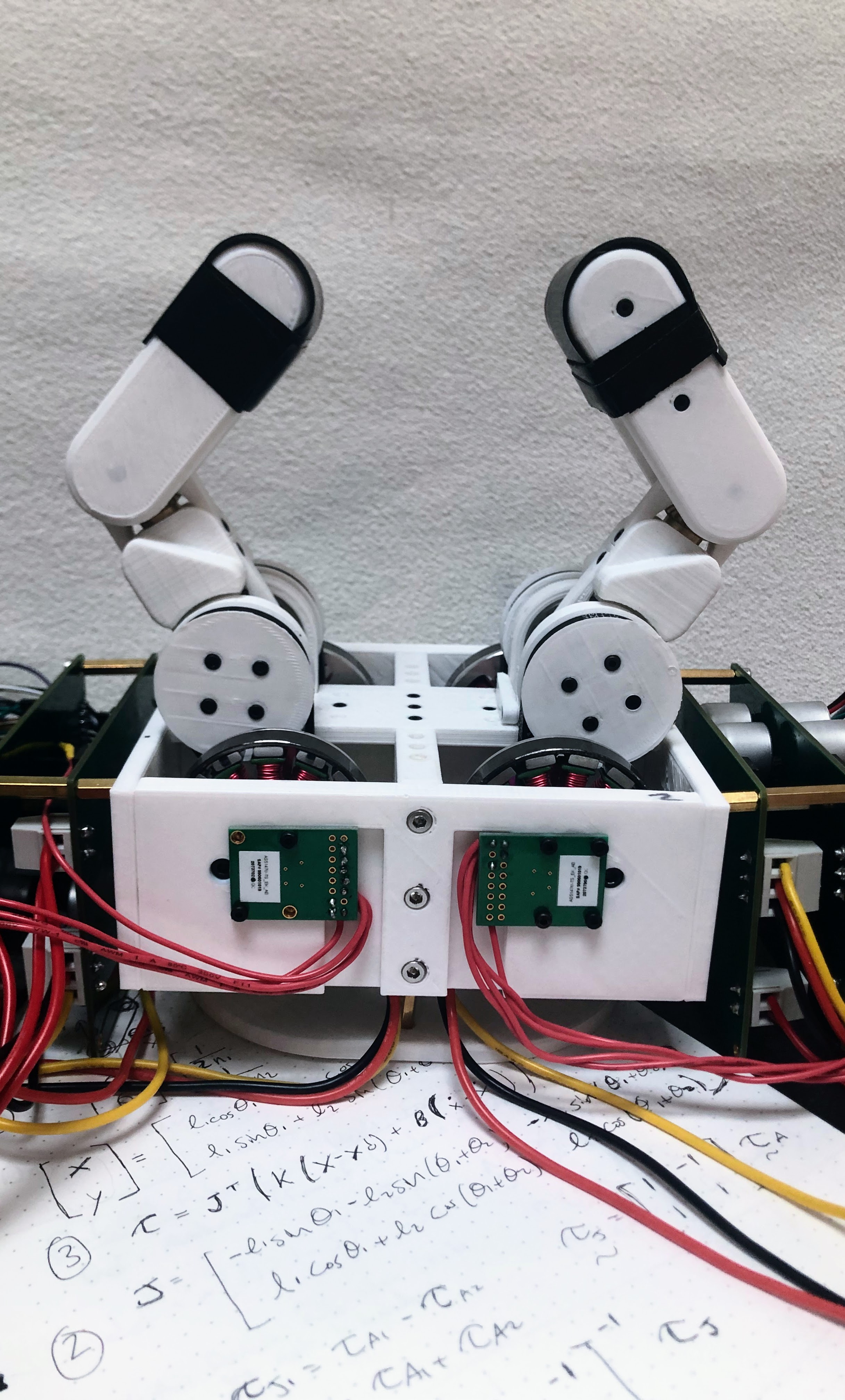}
\end{center}


\begin{spacing}{1.5}
\tableofcontents 
\end{spacing}

\listoffigures 

\listoftables 


\begin{abbreviations}{ll} 

\textbf{BLDC}
& \textbf{B}rush\textbf{l}ess \textbf{D}irect \textbf{C}urrent
\\
\textbf{FOC}
& \textbf{F}ield \textbf{O}riented \textbf{C}ontrol
\\
\textbf{QDD}
& \textbf{Q}uasi \textbf{D}irect \textbf{D}rive
\\
\textbf{DD}
& \textbf{D}irect \textbf{D}rive
\\
\textbf{CPR}
& \textbf{C}ycles \textbf{P}er \textbf{R}evolution
\\
\textbf{UART}
& \textbf{U}niversal \textbf{A}synchronous \textbf{R}eceiver/\textbf{T}ransmitter
\\
\textbf{MCU}
& \textbf{M}icro \textbf{C}ontroller \textbf{U}nit
\\
\textbf{DOF}
& \textbf{D}egree(s) \textbf{O}f \textbf{F}reedom
\\
\textbf{PWM}
& \textbf{P}ulse \textbf{W}idth \textbf{M}odulation

\end{abbreviations}


\begin{acknowledgements}
\addchaptertocentry{\acknowledgementname} 

I would like to thank Dr. Fakhari, for agreeing to be my advisor, and for all your help and wisdom this past year to make this project a success. Thank you to Dr. Chakraborty and Dr. Yao for agreeing to be committee members for my thesis defense. Thank you to all my friends and classmates who have made my time here at Stony Brook so enjoyable and memorable. Thank you to my parents, my siblings, and Mel for always supporting me in my educational endeavors.

\end{acknowledgements}


%
%
%


%
%
%
%


\dedicatory{Dedicated to my parents.}


\mainmatter 

\pagestyle{thesis} 

\chapter{Introduction}

\begin{figure}[!hbtp]
\centering
\includegraphics[scale=0.17]{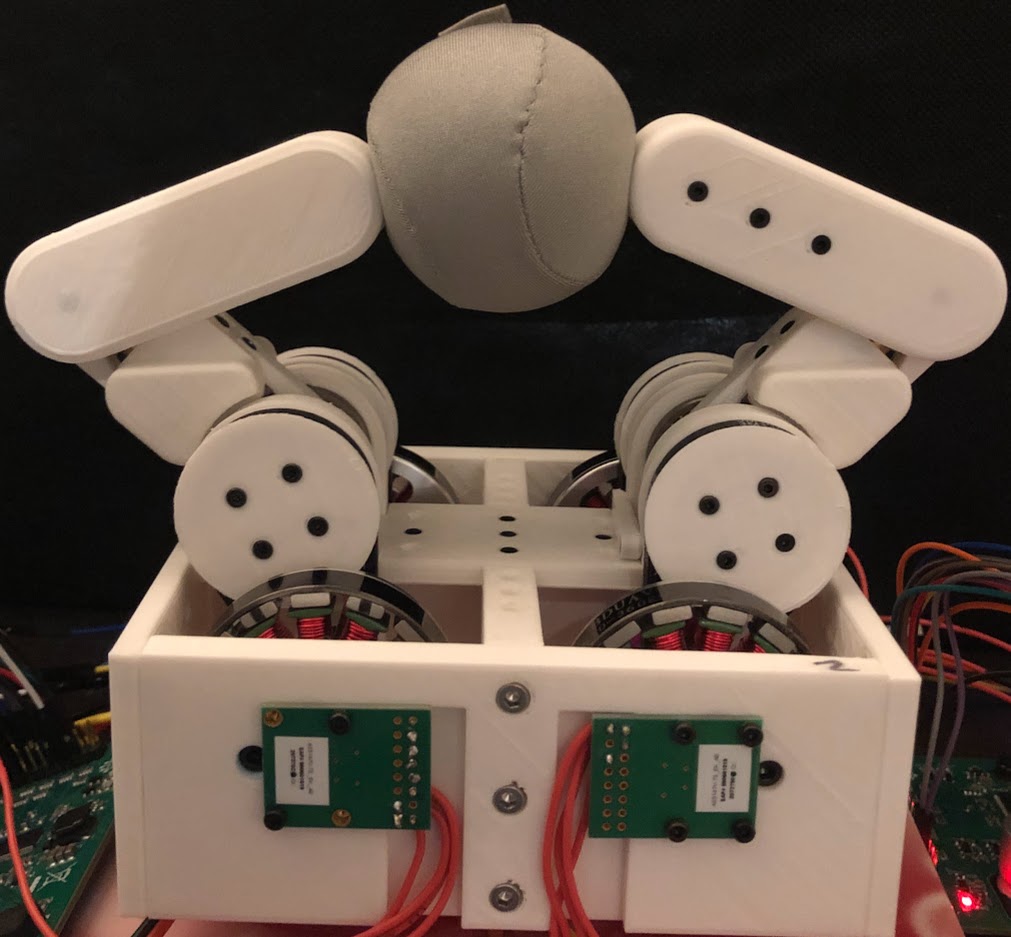}
\caption{Our quasi-direct-drive robotic hand holding a ball.}
\label{fig:intro_pic}
\end{figure}

Robotic grippers and hands have come a long way, however, there are still challenges when they have to significantly interact with an environment, e.g., picking a small object like a coin from an edge of a table, or rapid/dynamic grasping of a small object off an unstructured environment where the impact between the gripper/hand and the environment is inevitable. In these contact-rich grasping and manipulation scenarios, we have to prioritize the adaptivity and stiffness variation capability of the gripper/hand over large gripping forces. Therefore, the gripper will be able to comply with the environment and avoid damaging itself, the object, and possibly the environment.
Different strategies have been employed to handle and control contact of the gripper/hand with the environment during manipulation to ensure a safe and stable grasp and manipulation. One of the strategies which might be useful to explore further is impedance control combined with direct-drive (DD) or quasi-direct-drive (QDD) actuators (Fig.~\ref{fig:intro_pic}).

Impedance control differs from more traditional manipulation approaches in that the relationship between the force and position of the end-effector can be dynamically controlled \cite{Hogan1984}. This is especially important for grasping and dexterous manipulation tasks where contacts with the environment must be effectively dealt with. For example, if we were to apply force to the fingers of a robot hand, we would want to be able to control how the fingers respond to this force. Impedance control models this system as a simple linear mass-spring-damper system. In this way, instead of rigidly moving the fingers through a set of prescribed positions, we can treat the fingers as a ``virtual spring" with a desired starting position, stiffness, and damping behavior. Displacing the finger from its desired position will create a force in accordance with the behavior of the mass-spring-damper system. This contact force will be used to perform various manipulation tasks.

\begin{figure}[!htbp]
    \centering
    \sidesubfloat[]{\includegraphics[scale=0.5]{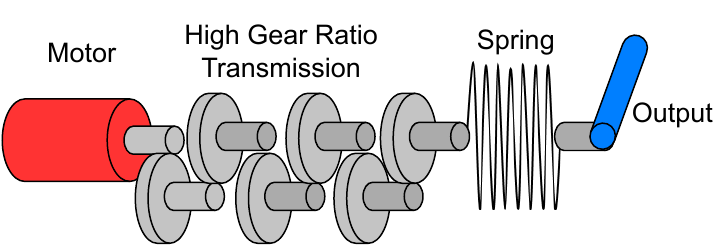}} \quad
    \sidesubfloat[]{\includegraphics[scale=0.5]{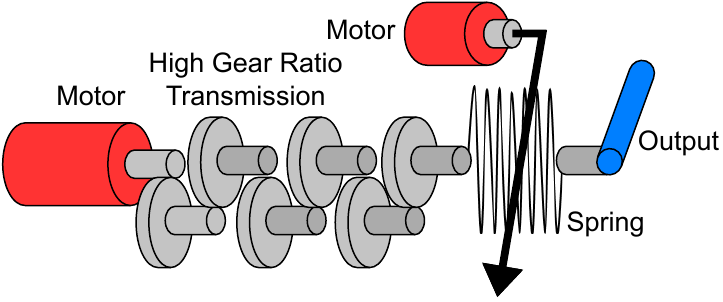}} \\
    \sidesubfloat[]{\includegraphics[scale=0.5]{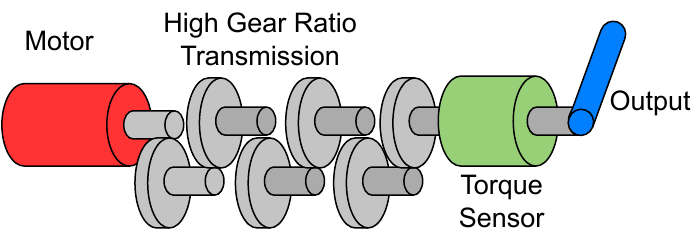}} \quad 
    \sidesubfloat[]{\includegraphics[scale=0.5]{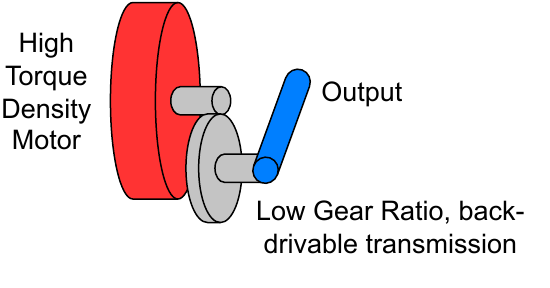}} \\
    \caption{Classification of robotic hands/grippers based on the actuator type, (a) Series elastic actuators, (b) Variable stiffness actuator, (c) Torque sensor, (d) Proprioceptive actuators (direct drive and quasi-direct-drive).}
\label{fig:classification}
\end{figure}

One way to classify previous robotic hand/gripper designs is on the basis of the type of actuator used, specifically how the actuator controls its output torque. These can be summarized in Fig.~\ref{fig:classification}.

(a) Series elastic actuators work by using a motor with a large gear reduction and a spring placed in between the motor and output shaft. By measuring the deformation of the spring with a sensor, the output torque can be calculated and controlled. (b) Variable stiffness actuators are a variation of this concept with the addition of an additional motor to control the stiffness of the spring. Both of these methods can accurately control torque, but at the expense of additional mechanical complexity. Maximum torque is also limited by the stiffness of the spring being used. (c) The most common actuator type used in previous robotic hands and grippers is a motor with a large gear reduction and a force/torque sensor used to control the output force of the end-effector. The advantage of this is that by using a large gear reduction, the maximum output force of the end-effector can be very high, along with the overall actuator torque density. However, force control bandwidth is limited by the torque sensor bandwidth. A large gear reduction also introduces additional complexity in the mechanical design and control of the actuator. Often, motors with a large gear reduction need to incorporate friction compensation into the controller, to compensate for the friction in the gears.

(d) Proprioceptive is a term that includes both direct-drive and quasi-direct-drive actuators. They differ from the previous actuators in that they use a higher torque density motor with a smaller gear reduction. By reducing friction in the gear train, the mechanism can be highly backdrivable, and motor phase currents can be used to control the output torque. Backdriveabile refers to how easily the output of a particular mechanism can be moved around when the motors are powered off. Due to their high bandwidth force control and impact resistance, this category of actuator is used heavily in walking robots such as the MIT Cheetah which experience highly dynamic interaction between the robot legs and the ground \cite{7827048}. For this reason, we believe that they would also be well suited for handling the contact forces between an object and the fingers in a robotic hand. Compared with other methods, these actuators suffer from reduced torque density, however, by using brushless direct current (BLDC) motors with more sophisticated methods of control such as Field-Oriented Control (FOC), we can control torque accurately and more efficiently while maintaining a high enough force output for most basic tasks. The comparative lack of friction in direct-drive and quasi-direct-drive robots makes them a better candidate for applications requiring variable force and compliance control while not requiring external force sensors or springs \cite{10.5555/27675}. However, for quasi-direct-drive actuators to control torque effectively, the gear reduction should in general have a reduction of less than 10:1 with negligible friction \cite{manipulation}.

\section{Motivation}

In a 2018 review entitled \textit{Towards Robotic Manipulation} \cite{doi:10.1146/annurev-control-060117-104848}, robotics pioneer Mattew Mason describes a phenomenon known as Moravec’s paradox, which he states is part of what makes robotic manipulation such a difficult task. The example he uses is the game of chess. In the field of artificial intelligence, the world's best chess computer defeated the world's best chess player for the first time in 1997 \cite{CAMPBELL200257}. Today, as technology has progressed, a chess program running on a smartphone would easily crush the world's best players. The paradox is that while computers are significantly better than even the most elite humans at playing chess, nearly every human can move the pieces on the chess board better than a robot. It turns out, the simple act of moving pieces on the chess board is actually a harder problem in robotics and artificial intelligence than knowing the right moves to play, a task that top human players devote their lives to. Years later, humans can still move chess pieces better than robots, as well as many other seemingly commonplace tasks. In a world where robots will likely play a larger role in our lives and interact closely with humans, it is important to continue to think of new ways robotic manipulators and end-effectors can be improved in order to make them more useful, cheaper, and safer. This includes coming up with new ideas in both hardware and control.

\section{Literature Review}

One of the common forms of variable compliance control in robotic hands and grippers is through force and torque sensors. This includes tactile force sensors at the griper/fingertips and torque sensors in the actuator gear train. Salisbury was one of the first to create a multi-fingered robot hand with stiffness control using force sensors \cite{salisbury, doi:10.1177/027836498200100102}. The hand had three fingers with 3 DOF each driven by cables enabling the hand to grasp objects in its fingertips and complete some basic in-hand manipulation tasks. Stiffness control as used here is a subset of impedance control which neglects controlling the damping behavior of the end-effector. Another example is the DLR/HIT Hand II \cite{DLR_Hit_II} which used strain gauge-based torque sensors integrated with harmonic gears to measure joint torques used in the control scheme. Joint and Cartesian level impedance controllers are implemented to control finger compliance, while an extended Kalman filter is used for friction estimation in the finger gears. The Ishikawa hand \cite{IshikawaHand} also uses strain gauge-based torque sensors with harmonic gearboxes for force control, and it is able to achieve highly dynamic motion for tasks such as throwing, catching, and dynamic regrasping \cite{Ishikawa_Regrasping}. The two prior examples achieve high performance, however, the introduction of harmonic gearboxes with strain gauges significantly increases cost and mechanical complexity which might not be desirable in certain applications.

Park \textit{et al.} \cite{Park_VariableGraspingStiffness} developed an anthropomorphic robotic hand using Series Elastic Actuator (SEA) modules and fingertip force sensors for tactile feedback. Surface hardness of the object being grasped was estimated by the controller, and finger stiffness was adjusted accordingly through modulating impedance parameters. By this method, a variety of both hard and fragile objects were grasped successfully without damage. Hu \textit{et al.} \cite{hu_liu_xie_yao_liu_2022} created a design for dexterous finger with antagonistic variable stiffness actuators which varied the stiffness of the finger by modulating spring tension in the actuators. This pre-loaded tension on the springs in the finger actuators could be changed to obtain a desired stiffness behavior. The finger achieves 2 DOF by using a differential drive gear system whereby one finger joint is driven by the sum of two actuator torques, and the other finger joint is driven by the difference in actuator torques. This finger mechanism helped serve as an inspiration for our QDD hand.

There are a number of direct-drive and quasi-direct-drive hands and grippers developed in recent years. DDHand \cite{Bhatia2019DirectDH,9981569} uses direct-drive linkages to achieve high bandwidth force control without force sensors. The linkages are easily back-driven, enabling the actuators to effectively act as sensors that sense the environment while picking up an object at an unknown height. They demonstrated the effectiveness of their direct-drive mechanism in a manipulation task called ``smack and snatch" where an object is rapidly grasped off of an unstructured environment. However, direct-drive hands lack the high gripping force compared to traditional hands and grippers. Quasi-direct-drive hands use a gear ratio of less than 10:1 in order to retain the benefits of direct-drive with increased torque and force output. Lin \textit{et al.} \cite{QuasiDirectDrive_Lin} developed a novel QDD linkage-based robotic hand with a 4:1 gear reduction. With a QDD actuator, they were able to replicate the "smack and snatch" task first performed by the DDHand. They also integrated a hybrid-drive system with a secondary linear actuator which could be used to provide additional force when the QDD actuators were not sufficient. Another example of QDD manipulation is the TriFinger an inexpensive manipulation platform with three fingers designed from modified quasi-direct-drive actuators from the robot quadruped, Solo \cite{TriFinger}. The fingers are driven by BLDC motors with a 9:1 gear reduction allowing the mechanism to be backdrivable and torque controlled with current feedback. The transparent belt reduction allows the fingers to be impact-resistant and run for long periods without damaging hardware. One last example is the manipulation platform Blue, a quasi-direct-drive 7 DOF robot arm driven by low gear reduction BLDC motors complete with a parallel jaw gripper \cite{QuasiDirectDrive_Gealy, 8843134}. While parallel jaw grippers can perform well, they may have limited functionality/dexterity when it comes to certain tasks, such as in-hand manipulation when compared to robot hands with multi-degree-of-freedom fingers.

Our design presents design alternatives and potential improvements to existing robotic hands and grippers within the proprioceptive actuator category. Compared with the other direct-drive and quasi-direct-drive hands, our finger links are more anthropomorphic in nature. This is beneficial for performing a wide variety of grasps such as both force closure and form closures around an object. This is not possible with hands resembling parallel jaw grippers. Our design is capable of performing in-hand manipulation tasks by manipulating objects in the fingertips. Implementing impedance control in the joint and Cartesian space allows the hand to safely handle contacts with both objects and the environment, while also maintaining stable grasps in response to external wrenches on the object being grasped. For these reasons, we believe this design may serve as an inspiration for future quasi-direct-drive general-purpose robotic hands for use in industry, humanoid robots, prostheses, and potentially other applications.

\section{Contributions}
The contributions of our novel robotic hand design are as follows:
\begin{itemize}[noitemsep,topsep=0pt]
  \item Integration of a quasi-direct-drive actuation scheme with a differential drive gear train to control movements of 2 DOF fingers.
  \item Implementation of impedance control in joint and Cartesian spaces using field-oriented control to control torque for grasping objects.
  \item Design, manufacturing, and experimental verification of a quasi-direct-drive robot hand capable of force closure grasp, form closure grasp, in-hand manipulation, and safe contact with an unstructured environment.
  \item All electrical and mechanical systems are designed around affordable, readily available off-the-shelf components and a majority 3D printed structures.
\end{itemize}

\section{Thesis Outline}
The remainder of the thesis will be organized as follows: Chapter 2 will go over the necessary theoretical background for our implementation of impedance control and basic concepts in grasping and manipulations. Chapter 3 goes into detail on the mechanical and electrical design of the QDD robot hand. In Chapter 4 experimental results and performance of the QDD hand are presented. Lastly, Chapter 5 includes a final discussion and plans for future work.
\chapter{Theoretical Background}

\section{Impedance Control Methodology}
Impedance control models the actuator as a mass-spring-damper system with the adjustable parameters being desired position, stiffness coefficient, and damping coefficient \cite{b3c2715a826847b9b8b5282c93c1f325}. If we were to model a 1 DOF actuator with impedance control, we would get the following equation:
\begin{equation}
m\ddot{x} + b\dot{x} + kx = F
\label{impedance}
\end{equation}
where $x$ is a displacement from a desired position, $m$ is mass, $b$ is the damping coefficient, $k$ is the actuator stiffness, and $F$ is the force applied to the actuator by the environment. Low impedance is defined as when the parameters $m$, $k$, and $b$, are low, while for high impedance the inverse is true \cite{b3c2715a826847b9b8b5282c93c1f325}. Usually, impedance can be changed by adjusting the stiffness and damping coefficients of the system. A system with very accurate position control will have very high impedance. Small errors in the desired position will result in a large corrective force to maintain the desired position. By contrast, an accurate force-controlled robot will have low impedance, displaying small changes in force output from a given positional displacement.

If we take the Laplace transform of equation~\eqref{impedance}, we get the following equation:
\begin{equation}
(ms^2 + bs + k)X(s) = F(s)
\label{laplace}
\end{equation}
We can then define \textit{impedance} as $Z(s)$, the transfer function for this system, as
\begin{equation}
Z(s)=F(s)/X(s)
\label{transfer}
\end{equation}

Writing the equation in this way is useful because we can see how impedance is defined as a particular force output for a given disturbance in position. Rather than controlling force or position separately, impedance control defines a relation between the two which we can use to attempt to perform tasks with a robot, in our case a robot hand. Taking the inverse of equation~\eqref{transfer} gives us what is known as \textit{admittance} \cite{b3c2715a826847b9b8b5282c93c1f325}. The transfer function for admittance is
\begin{equation}
Y(s)=Z^{-1}(s)=X(s)/F(s)
\end{equation}

Moving beyond the simple 1 DOF example, we can design a controller for an $m$-dimensional task-space as
\begin{equation}
\boldsymbol{M}\ddot{\boldsymbol{x}} + \boldsymbol{B}\dot{\boldsymbol{x}} + \boldsymbol{K}\boldsymbol{x} = \boldsymbol{F}_\mathrm{external}
\label{impedance2}
\end{equation}
where $\boldsymbol{x}\in{\mathbb{R}^m}$ is a displacement from a desired pose, $\boldsymbol{M}\in{\mathbb{R}^{m\times m}}$,$\boldsymbol{B}\in{\mathbb{R}^{m\times m}}$, and $\boldsymbol{K}\in{\mathbb{R}^{m\times m}}$ are the mass, damping, and stiffness matrices, respectively, defined by the user according to the particular task, and $\boldsymbol{F}_\mathrm{external}\in{\mathbb{R}^m}$ is the generalized end-effector force. In terms of the actual implementation of the impedance control, the robot uses some form of position sensor to determine the end-effector position $\boldsymbol{x}$, and then, calculates the required joint torques for the robot to impart the force $\boldsymbol{F}_\mathrm{external}$ to the environment. This is the opposite of the admittance-based control scheme, which uses a force sensor to sense $\boldsymbol{F}_\mathrm{external}$, and then, calculates the appropriate joint motion. Both methods simulate a virtual mass-spring-damper system as described in equation~\eqref{impedance2}, however, impedance-based control is better for our application as our robot hand does not include force sensors at its fingertips.

We can calculate joint torques $\boldsymbol{\tau}\in{\mathbb{R}^n}$ for a impedance-controlled $n$-DOF robot manipulator/finger using 
\begin{equation}
\boldsymbol{\tau} = \boldsymbol{J}^\mathrm{T}(\boldsymbol{\theta})\left(\tilde{\boldsymbol{\Lambda}}(\boldsymbol{\theta})\ddot{\boldsymbol{x}}+\tilde{\boldsymbol{\eta}}(\boldsymbol{\theta},\dot{\boldsymbol{x}})-(\boldsymbol{M}\ddot{\boldsymbol{x}} + \boldsymbol{B}\dot{\boldsymbol{x}} + \boldsymbol{K}\boldsymbol{x})\right)
\label{impedance3}
\end{equation}
where $\boldsymbol{\theta}\in{\mathbb{R}^n}$ is joint positions, $\boldsymbol{J}(\boldsymbol{\theta})\in{\mathbb{R}^{m \times n}}$ is the Jacobian matrix, $\tilde{\boldsymbol{\Lambda}}(\boldsymbol{\theta})\ddot{\boldsymbol{x}}+\tilde{\boldsymbol{\eta}}(\boldsymbol{\theta},\dot{\boldsymbol{x}})$ is the robot dynamics compensation, and $\boldsymbol{M}\ddot{\boldsymbol{x}} + \boldsymbol{B}\dot{\boldsymbol{x}} + \boldsymbol{K}\boldsymbol{x}$ is the external force acting on the end-effector. Often, due to noisy acceleration measurements and low mass of the robot end-effector, the terms $\tilde{\boldsymbol{\Lambda}}(\boldsymbol{\theta})\ddot{\boldsymbol{x}}$ and $\boldsymbol{M}\ddot{\boldsymbol{x}}$ are neglected. Additionally, the nonlinear dynamics compensation model $\tilde{\boldsymbol{\eta}}(\boldsymbol{\theta},\dot{\boldsymbol{x}})$ can be replaced by a gravity compensation term at small velocities \cite{b3c2715a826847b9b8b5282c93c1f325}. By a minor abuse of notation, we can substitute $\boldsymbol{x}$ (displacement from a desired pose) by $\boldsymbol{x} - \boldsymbol{x}_d$ where $\boldsymbol{x}$ is now the current pose of the end-effector frame and  $\boldsymbol{x}_d \in{\mathbb{R}^m}$ is the desired pose of the end-effector frame. Therefore, equation \eqref{impedance3} can be written as
\begin{equation}
\boldsymbol{\tau} = \boldsymbol{J}^\mathrm{T}(\boldsymbol{\theta})\left(\boldsymbol{K}(\boldsymbol{x}_d - \boldsymbol{x}) - \boldsymbol{B}\dot{\boldsymbol{x}}\right) + \tilde{\boldsymbol{g}}(\boldsymbol{\theta})
\label{impedance4}
\end{equation}
where $\tilde{\boldsymbol{g}}(\boldsymbol{\theta}) \in \mathbb{R}^n$ is the gravitational term. Note that $\dot{\boldsymbol{x}}_d=\ddot{\boldsymbol{x}}_d=\boldsymbol{0}$.

\subsection{Application to the Robot Finger}
\label{sec:Application to the Robot Finger}
\begin{figure}[!htbp]
\centering
\input{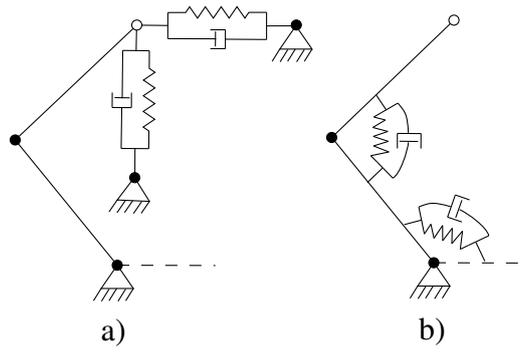}
\caption{Simplified model of a 2-DOF finger, (a) Impedance control in planar Cartesian space, (b) Impedance control in joint space.
}
\label{fig:Models}
\end{figure}

\begin{figure}[!htbp]
\centering
\input{Figures/gear_diagram.tikz} \quad
\includegraphics[scale=0.17]{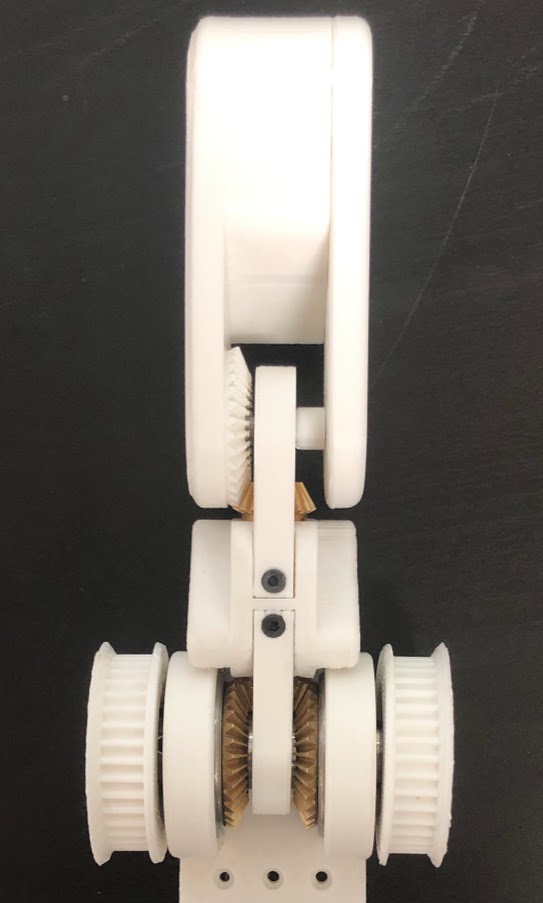} \\[2mm]
\includegraphics[scale=0.6]{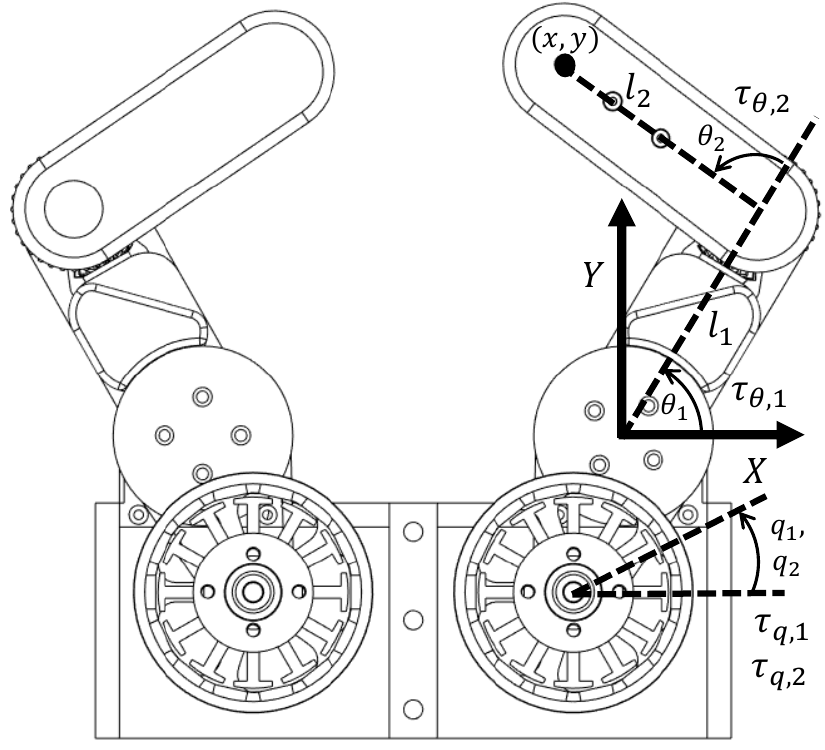}
\caption{Diagram of the gear train and kinematic model of each finger.}
\label{fig:gears}
\end{figure}

With a planar 2-DOF finger mechanism, as we have designed (Fig.~\ref{fig:gears}), impedance control can be realized in either the 2D Cartesian space or joint space. These respective methods are summarized in Fig.~\ref{fig:Models}.
In planar Cartesian space, the impedance controller attempts to simulate a mass-spring-damper system in the $X$ and $Y$ directions of the task space. The joint space impedance controller simulates a torsional mass-spring-damper system at each of the finger joints.

A schematic of the gear mechanism used in each finger and the kinematic model of the finger are shown in Fig.~\ref{fig:gears}.
Since our mechanism uses a differential gear drive to drive the respective joints of each 2-DOF finger (see Sec.~\ref{sec:Mechanical Design}), joint angles $\boldsymbol{\theta}=(\theta_1,\theta_2)$ for each finger can be calculated with motor encoder values $\boldsymbol{q}=(q_1,q_2)$ using
\begin{equation}
\boldsymbol{\theta}
=
\begin{bmatrix}
\frac{1}{2n_1} & \frac{-1}{2n_1} \\
\frac{1}{2n_1n_2} & \frac{1}{2n_1n_2}
\end{bmatrix}
\boldsymbol{q}
\label{theta}
\end{equation}
where $n_1$ is the motor belt reduction, which in our case $n_1 = 2.57$, and $n_2$ is the bevel gear reduction in the second joint, which in our case $n_2 = 1$. Moreover, the joint torques $\boldsymbol{\tau}_\theta=(\tau_{\theta,1},\tau_{\theta,2})$ can be converted to the required actuator torques $\boldsymbol{\tau}_q=(\tau_{q,1},\tau_{q,2})$ using
\begin{equation}
\boldsymbol{\tau}_q=\begin{bmatrix}
1/n_1 & 1/n_1\\ 
-1/n_1 & 1/n_1
\end{bmatrix}
\boldsymbol{\tau}_\theta.
\label{ta}
\end{equation}



Implementation of 2D Cartesian space impedance control requires the calculation of actuator torques $\boldsymbol{\tau}_q$ for each cycle of the control loop using equation \ref{ta}. But first, the finger joint torques must be calculated. Due to the low mass of our 3D-printed fingers, we can neglect the gravity compensation term in equation \eqref{impedance4}. Therefore, the joint torques $\boldsymbol{\tau}_\theta$ for an impedance-controlled planar 2-DOF finger can be computed by 
\begin{equation}
\boldsymbol{\tau}_\theta = \boldsymbol{J}^\mathrm{T}(\boldsymbol{\theta})\left(\boldsymbol{K}_{\boldsymbol{x}}(\boldsymbol{x}_d - \boldsymbol{x}) - \boldsymbol{B}_{\boldsymbol{x}}\dot{\boldsymbol{x}}\right)
\label{impedance_finger}
\end{equation}
where $\boldsymbol{K}_{\boldsymbol{x}}$ is the desired stiffness matrix and $\boldsymbol{B}_{\boldsymbol{x}}$ is the desired damping matrix. We consider these matrices to be diagonal as
\begin{equation}
\boldsymbol{K}_{\boldsymbol{x}}=\begin{bmatrix}
k_x & 0 \\ 
0 & k_y
\end{bmatrix};\\
\boldsymbol{B}_{\boldsymbol{x}}=\begin{bmatrix}
b_x & 0 \\ 
0 & b_y
\end{bmatrix}\\
\label{k and b}
\end{equation}
where $k_x$, $k_y$, $b_x$, and $b_x$ represent the desired stiffness and damping behavior in the $X$ and $Y$ directions. 


The fingertip position is calculated using the forward kinematic model as 
\begin{equation}
\boldsymbol{x}=
\begin{bmatrix}
x\\ 
y
\end{bmatrix}
=
\begin{bmatrix}
l_1\cos(\theta_1)+l_2\cos(\theta_1+\theta_2)\\
l_1\sin(\theta_1)+l_2\sin(\theta_1+\theta_2)
\end{bmatrix}
\label{fk}
\end{equation}
Moreover, the Jacobian $\boldsymbol{J}$ where $\boldsymbol{\dot{x}}=\boldsymbol{J}(\boldsymbol{\theta})\boldsymbol{\dot{\theta}}$ and $\boldsymbol{\dot{x}}$ is the velocity of the fingertip is calculated as 
\begin{equation}
\boldsymbol{J}(\boldsymbol{\theta})=\begin{bmatrix}
-l_1\sin(\theta_1)-l_2\sin(\theta_1+\theta_2) & -l_2\sin(\theta_1+\theta_2)\\ 
l_1\cos(\theta_1)+l_2\cos(\theta_1+\theta_2) & l_2\cos(\theta_1+\theta_2)
\end{bmatrix}
\label{j}
\end{equation}





\begin{figure}[!htbp]
\centering
\tikzset{every picture/.style={line width=0.75pt}} 

\begin{tikzpicture}[x=0.75pt,y=0.75pt,yscale=-1,xscale=1]

\draw  [dash pattern={on 4.5pt off 4.5pt}]  (239.89,135.31) -- (200.8,172.91) ;
\draw [color={rgb, 255:red, 208; green, 2; blue, 27 }  ,draw opacity=1 ]   (200.8,172.91) -- (141.45,232.79) ;
\draw [color={rgb, 255:red, 208; green, 2; blue, 27 }  ,draw opacity=1 ]   (151.21,112) -- (200.8,172.91) ;
\draw    (90.5,169.79) -- (132.38,129.92) -- (151.21,112) ;
\draw    (90.5,169.79) -- (141.45,232.79) ;
\draw   (141.45,232.79) -- (149.5,244.5) -- (133.4,244.5) -- cycle ;
\draw    (133.4,244.5) -- (129.58,250.67) ;
\draw    (137.4,244.75) -- (133.58,250.92) ;
\draw    (141.15,244.5) -- (137.33,250.67) ;
\draw    (149.5,244.5) -- (145.68,250.67) ;
\draw    (145.65,244.75) -- (141.83,250.92) ;

\draw  [fill={rgb, 255:red, 255; green, 255; blue, 255 }  ,fill opacity=1 ] (148.68,111.98) .. controls (148.69,110.58) and (149.84,109.46) .. (151.23,109.47) .. controls (152.63,109.48) and (153.75,110.63) .. (153.73,112.02) .. controls (153.72,113.42) and (152.58,114.54) .. (151.18,114.52) .. controls (149.79,114.51) and (148.67,113.37) .. (148.68,111.98) -- cycle ;
\draw  [fill={rgb, 255:red, 0; green, 0; blue, 0 }  ,fill opacity=1 ] (87.97,169.76) .. controls (87.99,168.37) and (89.13,167.25) .. (90.52,167.26) .. controls (91.92,167.27) and (93.04,168.41) .. (93.03,169.81) .. controls (93.01,171.2) and (91.87,172.32) .. (90.48,172.31) .. controls (89.08,172.3) and (87.96,171.16) .. (87.97,169.76) -- cycle ;
\draw  [fill={rgb, 255:red, 0; green, 0; blue, 0 }  ,fill opacity=1 ] (138.92,232.76) .. controls (138.94,231.37) and (140.08,230.25) .. (141.47,230.26) .. controls (142.87,230.27) and (143.99,231.41) .. (143.98,232.81) .. controls (143.96,234.2) and (142.82,235.32) .. (141.43,235.31) .. controls (140.03,235.3) and (138.91,234.16) .. (138.92,232.76) -- cycle ;
\draw  [dash pattern={on 4.5pt off 4.5pt}]  (141.45,232.79) -- (162.9,232.86) -- (190.4,232.95) ;
\draw  [color={rgb, 255:red, 208; green, 2; blue, 27 }  ,draw opacity=1 ][fill={rgb, 255:red, 208; green, 2; blue, 27 }  ,fill opacity=1 ] (198.27,172.89) .. controls (198.29,171.5) and (199.43,170.38) .. (200.82,170.39) .. controls (202.22,170.4) and (203.34,171.54) .. (203.33,172.94) .. controls (203.31,174.33) and (202.17,175.45) .. (200.78,175.44) .. controls (199.38,175.43) and (198.26,174.29) .. (198.27,172.89) -- cycle ;
\draw  [dash pattern={on 4.5pt off 4.5pt}]  (53.49,127.31) -- (90.5,169.79) ;

\draw (41.2,186) node [anchor=north west][inner sep=0.75pt]  [font=\footnotesize] [align=left] {Desired \\Configuration};
\draw (189.2,192.4) node [anchor=north west][inner sep=0.75pt]  [font=\footnotesize] [align=left] {Undesirable \\Configuration};
\draw (86.8,142) node [anchor=north west][inner sep=0.75pt]    {$\theta $};
\draw (192.4,138.8) node [anchor=north west][inner sep=0.75pt]    {$-\theta $};

\end{tikzpicture}
\caption{2R robot finger showing desired and undesirable configurations.}
\label{fig:finger config}
\end{figure}
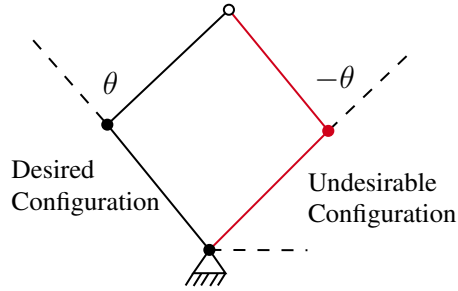

This controller works for most scenarios, however, there is one special case that must be considered to create a more robust controller. Consider the scenario of setting the desired position of the fingertip to a specified $(X,Y)$ coordinate. For much of the finger workspace, there are two possible configurations for the fingertip to reach these coordinates. Usually there is one configuration where $\theta_2 > 0$, and another configuration where $\theta_2 < 0$. However, for performing most tasks, we only want the finger to consider the configuration where $\theta_2 > 0$. This is represented in Fig.~\ref{fig:finger config}.


In order to solve this problem, under the condition $\theta_2 < 0$ (or some small positive angle), we modify equation \eqref{impedance_finger}, the joint torque equation, by simply setting the torque at joint 2 equal to a constant positive value $\tau^*$ as
\begin{equation}
\tau_{\theta,2} = \tau^*
\label{tau mod}
\end{equation}
Essentially, adding a set torque to joint 2 to prevent $\theta_2$ from going negative. The constant, $\tau^*$ defines how much the torque at joint 2 increases when the value of $\theta_2$ becomes negative. This is a very simple way of doing this which can perhaps be improved in the future, but it does get the job done for our purposes.

\subsubsection{Joint Space Impedance Controller}
The joint space impedance controller is actually even simpler to implement than the Cartesian space impedance controller. We simply modify equation \eqref{impedance_finger} as
\begin{equation}
\boldsymbol{\tau}_{\theta} = \boldsymbol{K}_{\theta}(\boldsymbol{\theta}_d-\boldsymbol{\theta})-\boldsymbol{B}_{\theta}\boldsymbol{\dot{\theta}}
\label{tau_joint}
\end{equation}
where $\boldsymbol{K}_{\theta}$ and $\boldsymbol{B}_{\theta}$ are the desired stiffness and damping behavior of each of the finger joints as
\begin{equation}
\boldsymbol{K}_{\theta}=\begin{bmatrix}
k_{\theta_1} & 0 \\ 
0 & k_{\theta_2}
\end{bmatrix}, \quad
\boldsymbol{B}_{\theta}=\begin{bmatrix}
b_{\theta_1} & 0 \\ 
0 & b_{\theta_2}
\end{bmatrix}.
\label{theta_k_b}
\end{equation}

\section{Block Diagram of Impedance Controller Implementation}
\begin{figure}[!htbp]
\centering
\subfloat[]{\includegraphics[scale=0.9]{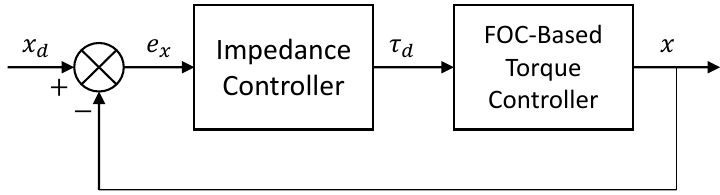}}\\ \vspace{3mm}
\subfloat[]{\includegraphics[scale=0.47]{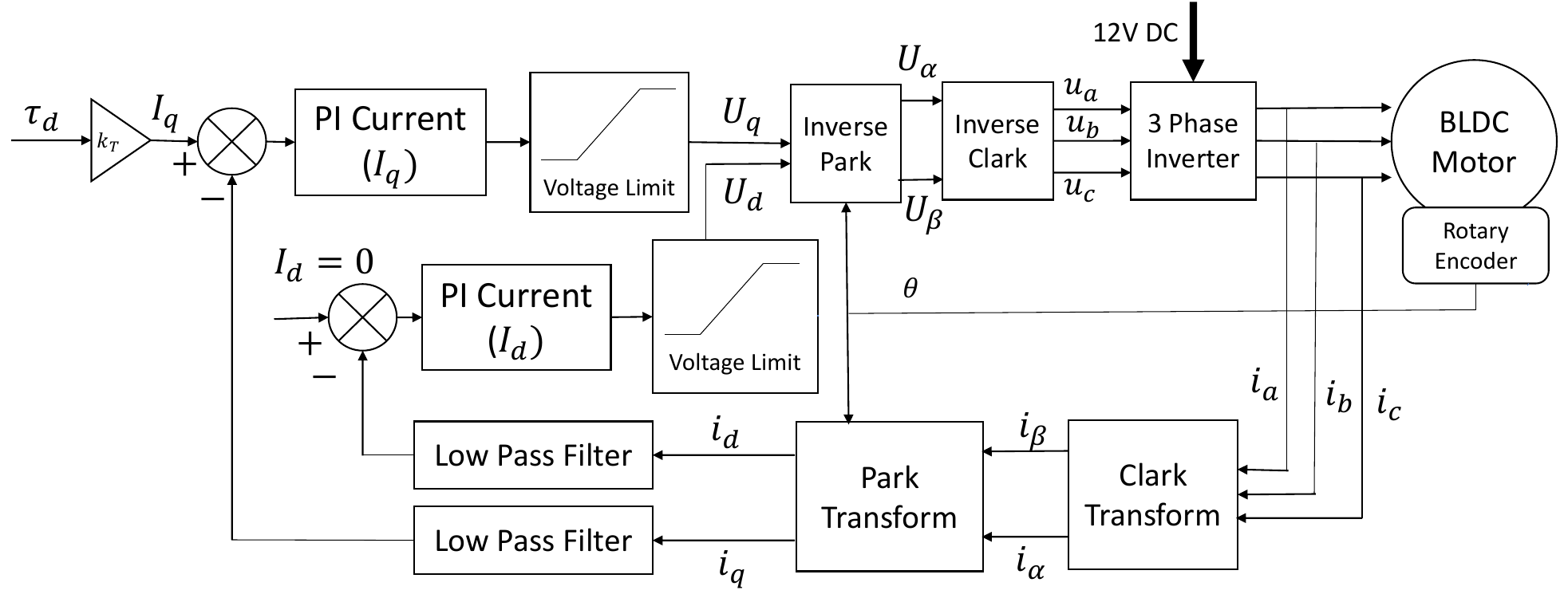}}
\caption{(a) Block diagram of impedance controller implementation for each motor (b) Expanded detailed view of FOC-Based Torque Controller. The parameters are defined in Table~\ref{table:Block Diagram Parameters}.}
\label{fig:ControllerBlockDiagram}
\end{figure}

The control system for each motor of the fingers of the robot hand consists of an outer impedance controller for adjusting the stiffness and damping coefficients and an inner FOC-based torque controller to control the torque provided by the impedance controller as shown in Fig.~\ref{fig:ControllerBlockDiagram}. The FOC torque control (Fig.~\ref{fig:ControllerBlockDiagram}-B, Table~\ref{table:Block Diagram Parameters}) is implemented by leveraging the SimpleFOC open-source Arduino library \cite{simplefoc2022}. This method enables accurate and efficient torque control at high bandwidth which is essential for dynamic manipulation tasks \cite{act8040071}. The inner torque controller runs 5 times faster than the outer loop impedance controller. See Sec.~\ref{sec:FOC} for a more in-depth look at FOC control.

\begin{table}[htbp]
\caption{Block Diagram Parameters.}
\label{Block-Diagram-table}
\centering
\begin{tabular}{ |c|c| } 
\hline
 \textbf{Parameter} & \textbf{Decription} \\
\hline
 $X_d$        &  Desired Position     \\
 $e_x$       & Error in Position     \\
 $\tau_d$    &  Desired Motor Torque      \\
 $k_t$        & Motor torque Constant   \\
 $I_q$      & Motor Desired Quadrature Current       \\
 $I_d$      & Motor Desired Direct Current       \\
 $U_q$        & Motor Quadrature Voltage      \\
 $U_d$       & Motor Direct Voltage       \\
 $u_a, u_b, u_c$ & Motor Phase Voltages \\
 $i_a, i_b, i_c$ & Motor Phase Currents \\
 $i_q$ & Measured Quadrature Current \\
 $i_d$ & Measured Direct Current \\
 $i_\alpha, i_\beta$ & 2D Stationary Axis Current\\
 $U_\alpha, U_\beta$ & 2D Stationary Axis Voltage\\
 \hline
\end{tabular}
\label{table:Block Diagram Parameters}
\end{table}

\section{Field-Oriented Control}
\label{sec:FOC}
Field-Oriented Control (FOC), also known as vector control, is the method by which we control the torque of the BLDC motors which is required to implement impedance control. FOC-based torque control, while being more complicated to implement, allows for more accurate torque control, and greater motor efficiency across the whole range of its operating speed \cite{Kalouche-2016-5569}. For our application, this is imperative, because we are not relying on an extra sensors to regulate the motor torque output. DD and QDD actuators already suffer from reduced torque density compared to higher gear ratio motors, so using a more efficient means of torque control helps to mitigate this. Of course, this comes at the cost of increased computation, but as computer hardware becomes cheaper, this is likely to improve in the future.

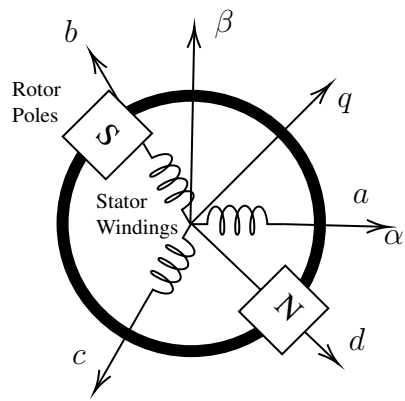
\begin{figure}[!htbp]
\centering
\tikzset{every picture/.style={line width=0.75pt}} 

\begin{tikzpicture}[x=0.75pt,y=0.75pt,yscale=-1,xscale=1]

\draw    (148.91,147.34) -- (221.02,216.68) ;
\draw [shift={(222.47,218.07)}, rotate = 223.88] [color={rgb, 255:red, 0; green, 0; blue, 0 }  ][line width=0.75]    (10.93,-3.29) .. controls (6.95,-1.4) and (3.31,-0.3) .. (0,0) .. controls (3.31,0.3) and (6.95,1.4) .. (10.93,3.29)   ;
\draw  [color={rgb, 255:red, 0; green, 0; blue, 0 }  ,draw opacity=1 ][fill={rgb, 255:red, 255; green, 255; blue, 255 }  ,fill opacity=1 ][line width=0.75]  (125.07,107.14) -- (129.28,114.43) .. controls (132.73,112.58) and (136.21,111.93) .. (138.05,112.78) .. controls (139.89,113.63) and (139.71,115.81) .. (137.61,118.28) .. controls (135.96,120.19) and (133.46,122.03) .. (130.75,123.32) .. controls (129.74,123.9) and (128.71,124.01) .. (128.45,123.56) .. controls (128.19,123.12) and (128.8,122.28) .. (129.81,121.7) .. controls (132.28,119.99) and (135.12,118.74) .. (137.61,118.28) .. controls (140.29,117.85) and (142.22,118.3) .. (142.93,119.53) .. controls (143.63,120.76) and (143.06,122.65) .. (141.35,124.76) .. controls (139.7,126.68) and (137.21,128.52) .. (134.49,129.8) .. controls (133.48,130.39) and (132.45,130.5) .. (132.19,130.05) .. controls (131.94,129.6) and (132.55,128.76) .. (133.56,128.18) .. controls (136.03,126.48) and (138.87,125.23) .. (141.35,124.76) .. controls (144.04,124.34) and (145.96,124.79) .. (146.67,126.02) .. controls (147.38,127.24) and (146.81,129.14) .. (145.1,131.25) .. controls (143.45,133.17) and (140.95,135) .. (138.24,136.29) .. controls (137.23,136.87) and (136.2,136.98) .. (135.94,136.54) .. controls (135.68,136.09) and (136.29,135.25) .. (137.3,134.67) .. controls (139.77,132.96) and (142.61,131.72) .. (145.1,131.25) .. controls (148.29,130.66) and (150.26,131.6) .. (150.08,133.62) .. controls (149.89,135.64) and (147.59,138.32) .. (144.26,140.38) -- (148.48,147.68) ;
\draw  [line width=4.5]  (84.67,147.34) .. controls (84.67,111.86) and (113.43,83.1) .. (148.91,83.1) .. controls (184.38,83.1) and (213.14,111.86) .. (213.14,147.34) .. controls (213.14,182.81) and (184.38,211.57) .. (148.91,211.57) .. controls (113.43,211.57) and (84.67,182.81) .. (84.67,147.34) -- cycle ;
\draw  [color={rgb, 255:red, 0; green, 0; blue, 0 }  ,draw opacity=1 ] (148.48,147.68) -- (156.9,147.68) .. controls (157.02,143.77) and (158.19,140.43) .. (159.85,139.26) .. controls (161.5,138.09) and (163.31,139.33) .. (164.39,142.39) .. controls (165.23,144.77) and (165.57,147.85) .. (165.33,150.85) .. controls (165.33,152.01) and (164.91,152.96) .. (164.39,152.96) .. controls (163.87,152.96) and (163.45,152.01) .. (163.45,150.85) .. controls (163.21,147.85) and (163.55,144.77) .. (164.39,142.39) .. controls (165.36,139.85) and (166.72,138.41) .. (168.14,138.41) .. controls (169.55,138.41) and (170.91,139.85) .. (171.88,142.39) .. controls (172.72,144.77) and (173.06,147.85) .. (172.82,150.85) .. controls (172.82,152.01) and (172.4,152.96) .. (171.88,152.96) .. controls (171.36,152.96) and (170.94,152.01) .. (170.94,150.85) .. controls (170.7,147.85) and (171.04,144.77) .. (171.88,142.39) .. controls (172.85,139.85) and (174.21,138.41) .. (175.62,138.41) .. controls (177.04,138.41) and (178.4,139.85) .. (179.37,142.39) .. controls (180.21,144.77) and (180.55,147.85) .. (180.31,150.85) .. controls (180.31,152.01) and (179.89,152.96) .. (179.37,152.96) .. controls (178.85,152.96) and (178.43,152.01) .. (178.43,150.85) .. controls (178.19,147.85) and (178.53,144.77) .. (179.37,142.39) .. controls (180.45,139.33) and (182.26,138.09) .. (183.91,139.26) .. controls (185.57,140.43) and (186.74,143.77) .. (186.86,147.68) -- (195.28,147.68) ;
\draw  [color={rgb, 255:red, 0; green, 0; blue, 0 }  ,draw opacity=1 ][fill={rgb, 255:red, 255; green, 255; blue, 255 }  ,fill opacity=1 ] (148.91,147.34) -- (144.7,154.63) .. controls (148.02,156.69) and (150.33,159.38) .. (150.51,161.39) .. controls (150.7,163.41) and (148.72,164.35) .. (145.53,163.76) .. controls (143.05,163.3) and (140.21,162.05) .. (137.74,160.34) .. controls (136.72,159.76) and (136.11,158.92) .. (136.37,158.48) .. controls (136.63,158.03) and (137.66,158.14) .. (138.67,158.72) .. controls (141.38,160.01) and (143.88,161.85) .. (145.53,163.76) .. controls (147.24,165.88) and (147.81,167.77) .. (147.1,169) .. controls (146.39,170.22) and (144.47,170.68) .. (141.78,170.25) .. controls (139.3,169.78) and (136.46,168.54) .. (133.99,166.83) .. controls (132.98,166.25) and (132.37,165.41) .. (132.63,164.96) .. controls (132.89,164.52) and (133.92,164.63) .. (134.93,165.21) .. controls (137.64,166.5) and (140.14,168.33) .. (141.78,170.25) .. controls (143.5,172.36) and (144.07,174.25) .. (143.36,175.48) .. controls (142.65,176.71) and (140.73,177.16) .. (138.04,176.74) .. controls (135.56,176.27) and (132.72,175.02) .. (130.25,173.32) .. controls (129.23,172.73) and (128.62,171.9) .. (128.88,171.45) .. controls (129.14,171) and (130.17,171.11) .. (131.18,171.7) .. controls (133.89,172.98) and (136.39,174.82) .. (138.04,176.74) .. controls (140.14,179.2) and (140.32,181.39) .. (138.48,182.24) .. controls (136.64,183.08) and (133.16,182.43) .. (129.72,180.58) -- (125.5,187.88) ;
\draw    (195.28,147.68) -- (245.47,149.01) ;
\draw [shift={(247.47,149.07)}, rotate = 181.53] [color={rgb, 255:red, 0; green, 0; blue, 0 }  ][line width=0.75]    (10.93,-3.29) .. controls (6.95,-1.4) and (3.31,-0.3) .. (0,0) .. controls (3.31,0.3) and (6.95,1.4) .. (10.93,3.29)   ;
\draw [color={rgb, 255:red, 0; green, 0; blue, 0 }  ,draw opacity=1 ]   (125.07,107.14) -- (99.47,62.8) ;
\draw [shift={(98.47,61.07)}, rotate = 60] [color={rgb, 255:red, 0; green, 0; blue, 0 }  ,draw opacity=1 ][line width=0.75]    (10.93,-3.29) .. controls (6.95,-1.4) and (3.31,-0.3) .. (0,0) .. controls (3.31,0.3) and (6.95,1.4) .. (10.93,3.29)   ;
\draw [color={rgb, 255:red, 0; green, 0; blue, 0 }  ,draw opacity=1 ]   (125.5,187.88) -- (101.45,230.33) ;
\draw [shift={(100.47,232.07)}, rotate = 299.53] [color={rgb, 255:red, 0; green, 0; blue, 0 }  ,draw opacity=1 ][line width=0.75]    (10.93,-3.29) .. controls (6.95,-1.4) and (3.31,-0.3) .. (0,0) .. controls (3.31,0.3) and (6.95,1.4) .. (10.93,3.29)   ;
\draw    (148.48,147.68) -- (150.43,50.07) ;
\draw [shift={(150.47,48.07)}, rotate = 91.15] [color={rgb, 255:red, 0; green, 0; blue, 0 }  ][line width=0.75]    (10.93,-3.29) .. controls (6.95,-1.4) and (3.31,-0.3) .. (0,0) .. controls (3.31,0.3) and (6.95,1.4) .. (10.93,3.29)   ;
\draw  [fill={rgb, 255:red, 255; green, 255; blue, 255 }  ,fill opacity=1 ] (195.84,168.51) -- (217.34,190.02) -- (194.26,213.1) -- (172.76,191.6) -- cycle ;
\draw  [fill={rgb, 255:red, 255; green, 255; blue, 255 }  ,fill opacity=1 ] (107.84,80.51) -- (129.34,102.02) -- (106.26,125.1) -- (84.76,103.6) -- cycle ;
\draw    (148.48,147.68) -- (217.06,78.49) ;
\draw [shift={(218.47,77.07)}, rotate = 134.75] [color={rgb, 255:red, 0; green, 0; blue, 0 }  ][line width=0.75]    (10.93,-3.29) .. controls (6.95,-1.4) and (3.31,-0.3) .. (0,0) .. controls (3.31,0.3) and (6.95,1.4) .. (10.93,3.29)   ;

\draw (197.41,180.1) node [anchor=north west][inner sep=0.75pt]  [rotate=-45] [align=left] {N};
\draw (108.23,93.45) node [anchor=north west][inner sep=0.75pt]  [rotate=-44.8] [align=left] {S};
\draw (228,128.4) node [anchor=north west][inner sep=0.75pt]    {$a$};
\draw (84,43.4) node [anchor=north west][inner sep=0.75pt]    {$b$};
\draw (88,209.4) node [anchor=north west][inner sep=0.75pt]    {$c$};
\draw (243.47,149.47) node [anchor=north west][inner sep=0.75pt]    {$\alpha $};
\draw (159,38.4) node [anchor=north west][inner sep=0.75pt]    {$\beta $};
\draw (58,75) node [anchor=north west][inner sep=0.75pt]  [font=\scriptsize] [align=left] {Rotor\\Poles};
\draw (100,131) node [anchor=north west][inner sep=0.75pt]  [font=\scriptsize] [align=left] {Stator\\Windings};
\draw (226,198.4) node [anchor=north west][inner sep=0.75pt]    {$d$};
\draw (220.47,80.47) node [anchor=north west][inner sep=0.75pt]    {$q$};

\end{tikzpicture}
\caption{Motor reference frames used in FOC calculations.}
\label{fig:motor frames}
\end{figure}

FOC involves transforming the motor voltages and currents between three reference frames; (1) a stationary 3D  frame $a-b-c$ which is aligned with the motor stator windings, (2) a stationary 2D frame $\alpha-\beta$, (3) the rotating $d-q$ frame which rotates with the motor's rotor. These are represented visually in Fig.~\ref{fig:motor frames}.

FOC begins with a measurement of the motor's three-phase currents, ($i_a$, $i_b$, $i_c$), in our case, using the current sensors on our motor driver. After this, the phase currents are transformed from the stationary 3D frame $a-b-c$ representing the motor windings to the stationary 2D frame $\alpha-\beta$ using the Clark transform \cite{hein} as 
\begin{equation}
i_\alpha=i_a-\frac{1}{2}i_b-\frac{1}{2}i_c
\end{equation}
\begin{equation}
i_\beta=\frac{\sqrt{3}}{2}(i_b-i_c)
\end{equation}
Next, the Park transform is used to transform the current vectors from the stationary frame $\alpha-\beta$ to the rotating 2D frame $d-q$ aligned with the motor rotor poles as
\begin{equation}
i_d = i_\alpha\cos(\theta)+i_\beta\sin(\theta)
\end{equation}
\begin{equation}
i_q = i_\beta\cos(\theta)-i_\alpha\sin(\theta)
\end{equation}
 The magnetic encoder estimates the motor rotor position $\theta$, which is used by the Park transform. In the rotating frame $d-q$, the quadrature current is what contributes to the motor torque, while the direct current does not. Two independent PI controllers are used to keep the quadrature current, $I_q$ equal to the desired current value, and the direct current, $I_d$, equal to zero, as shown in Fig.~\ref{fig:ControllerBlockDiagram}-B. The PI controllers generate desired voltages, $U_q$ and $U_d$, in the frame $d-q$. These are transformed back to the stationary frame $\alpha-\beta$ with the inverse Park transform as
\begin{equation}
U_\alpha = U_d\cos(\theta)-U_q\sin(\theta)
\end{equation}
\begin{equation}
U_\beta = U_q\cos(\theta)+U_d\sin(\theta)
\end{equation}
The inverse Clark transform, \eqref{eq:inverse Clark 1}-\eqref{eq:inverse Clark 3}, is then used to generate the desired phase voltages in the motor stator frame $a-b-c$ to be sent to the motor driver via PWM signal from the ESP32 microcontroller.
\begin{equation}
u_a=U_\alpha
\label{eq:inverse Clark 1}
\end{equation}
\begin{equation}
u_b=\frac{1}{2}(-U_\alpha+\sqrt{3}U_\beta)
\label{eq:inverse Clark 2}
\end{equation}
\begin{equation}
u_c=\frac{1}{2}(-U_\alpha-\sqrt{3}U_\beta)
\label{eq:inverse Clark 3}
\end{equation}

\begin{figure}[!htbp]
\centering
\includegraphics[width = 14 cm]{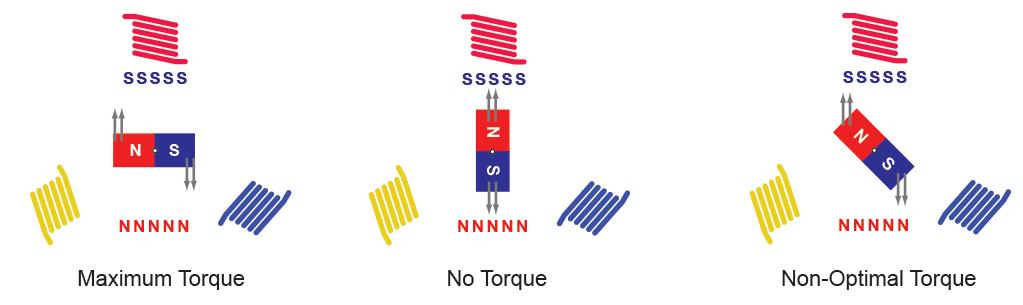}
\caption{Optimal motor torque generation for BLDC motor \cite{centiva}.}
\label{fig:foc torque}
\end{figure}

As a result of these operations, the electromagnetic field generated by the stator windings always stays perpendicular to the magnetic field of the permanent magnets of the rotor, ensuring optimal torque production. Optimal and non-optimal torque generation for a BLDC motor is presented in Fig. \ref{fig:foc torque}.

\begin{figure}[!htbp]
\centering
\subfloat[]{\includegraphics[width=13cm]{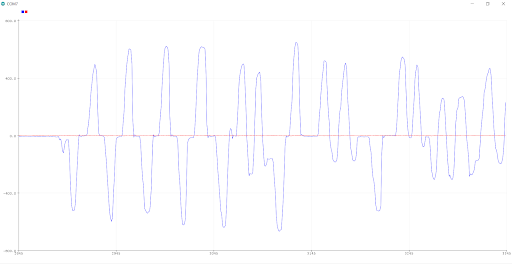}}\\
\subfloat[]{\includegraphics[width=13cm]{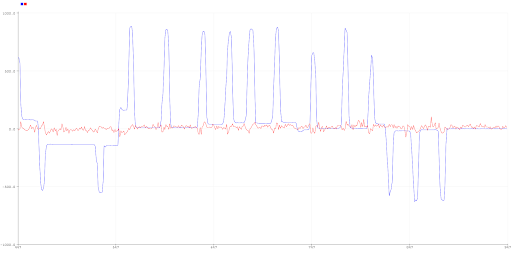}}
\caption{Quadrature current $i_q$ (blue) and direct current $i_d$ (red) with respect to time, (a) with direct current ($i_d$) PI controller, (b) without direct current ($i_d$) PI controller.}
\label{fig:foctest}
\end{figure}

The effect of FOC control is clearly seen with some simple motor tests. An early test we did was running the impedance controller with a single motor and specified values of stiffness, damping, and angular position. When the actual position is equal to the desired position, the current vectors $i_q$ and $i_d$ are both equal to 0. However, when the motor is disturbed and displaced from its desired position, the quadrature current, $i_q$, spikes up as the motor tries to correct its position. The direct current does not contribute to torque, so the PI controller tries to keep this value at 0 by setting the value of the motor direct voltage $U_d$. If we get rid of the direct current ($i_d$) PI controller and instead simply set $U_d$ to zero, what we find is that the direct current $i_d$ is very noisy and oscillates around 0. These graphs are shown in Fig. \ref{fig:foctest}. In these figures, $i_q$ is shown in blue while $i_d$ is shown in red. These graphs verify our controller worked as intended.

\pagebreak

\section{Force-Closure and Form-Closure Grasps}
The design for our robot hand is unique in comparison with other QDD hands in its ability to grasp with both force-closure and form-closure. Both methods can be useful depending on the task at hand.

\begin{figure}[!htbp]
\centering
\tikzset{every picture/.style={line width=0.75pt}} 

\begin{tikzpicture}[x=0.75pt,y=0.75pt,yscale=-0.8,xscale=0.8]

\draw  [fill={rgb, 255:red, 0; green, 0; blue, 0 }  ,fill opacity=0.24 ] (137.23,281.4) -- (137.23,241.9) .. controls (137.23,220.08) and (142.94,202.4) .. (150,202.4) .. controls (157.06,202.4) and (162.77,220.08) .. (162.77,241.9) -- (162.77,281.4) -- cycle ;
\draw    (150,202.4) -- (43.61,280.22) ;
\draw [shift={(42,281.4)}, rotate = 323.82] [color={rgb, 255:red, 0; green, 0; blue, 0 }  ][line width=0.75]    (10.93,-3.29) .. controls (6.95,-1.4) and (3.31,-0.3) .. (0,0) .. controls (3.31,0.3) and (6.95,1.4) .. (10.93,3.29)   ;
\draw    (150,202.4) -- (272.23,267.46) ;
\draw [shift={(274,268.4)}, rotate = 208.02] [color={rgb, 255:red, 0; green, 0; blue, 0 }  ][line width=0.75]    (10.93,-3.29) .. controls (6.95,-1.4) and (3.31,-0.3) .. (0,0) .. controls (3.31,0.3) and (6.95,1.4) .. (10.93,3.29)   ;
\draw  [fill={rgb, 255:red, 255; green, 255; blue, 255 }  ,fill opacity=0.46 ] (107.11,205.17) -- (148.92,179.62) -- (192.89,199.63) -- (151.08,225.18) -- cycle ;
\draw  [fill={rgb, 255:red, 255; green, 255; blue, 255 }  ,fill opacity=1 ] (205.03,106.72) -- (150,202.4) -- (91.8,108.31) -- cycle ;
\draw  [fill={rgb, 255:red, 155; green, 155; blue, 155 }  ,fill opacity=1 ] (92.5,107.9) .. controls (92.5,98.79) and (117.69,91.4) .. (148.77,91.4) .. controls (179.84,91.4) and (205.03,98.79) .. (205.03,107.9) .. controls (205.03,117.01) and (179.84,124.4) .. (148.77,124.4) .. controls (117.69,124.4) and (92.5,117.01) .. (92.5,107.9) -- cycle ;
\draw    (150,107.9) -- (149.02,9.3) ;
\draw [shift={(149,7.3)}, rotate = 89.43] [color={rgb, 255:red, 0; green, 0; blue, 0 }  ][line width=0.75]    (10.93,-3.29) .. controls (6.95,-1.4) and (3.31,-0.3) .. (0,0) .. controls (3.31,0.3) and (6.95,1.4) .. (10.93,3.29)   ;
\draw  [fill={rgb, 255:red, 0; green, 0; blue, 0 }  ,fill opacity=0.24 ] (325.23,273) -- (325.23,233.5) .. controls (325.23,211.68) and (330.94,194) .. (338,194) .. controls (345.06,194) and (350.77,211.68) .. (350.77,233.5) -- (350.77,273) -- cycle ;
\draw    (338,194) -- (338,79.4) ;
\draw [shift={(338,77.4)}, rotate = 90] [color={rgb, 255:red, 0; green, 0; blue, 0 }  ][line width=0.75]    (10.93,-3.29) .. controls (6.95,-1.4) and (3.31,-0.3) .. (0,0) .. controls (3.31,0.3) and (6.95,1.4) .. (10.93,3.29)   ;
\draw    (290.5,193.3) -- (385.5,194.7) ;
\draw  [color={rgb, 255:red, 155; green, 155; blue, 155 }  ,draw opacity=1 ] (338,194) -- (286.03,77.4) -- (387.08,77.4) -- cycle ;
\draw    (338,77.4) -- (385.08,77.4) ;
\draw [shift={(387.08,77.4)}, rotate = 180] [color={rgb, 255:red, 0; green, 0; blue, 0 }  ][line width=0.75]    (10.93,-3.29) .. controls (6.95,-1.4) and (3.31,-0.3) .. (0,0) .. controls (3.31,0.3) and (6.95,1.4) .. (10.93,3.29)   ;

\draw (157,39.4) node [anchor=north west][inner sep=0.75pt]    {$z$};
\draw (50,236.4) node [anchor=north west][inner sep=0.75pt]    {$x$};
\draw (242,225.4) node [anchor=north west][inner sep=0.75pt]    {$y$};
\draw (315,86.4) node [anchor=north west][inner sep=0.75pt]    {$f_{n}$};
\draw (353,48.4) node [anchor=north west][inner sep=0.75pt]    {$\mu f_{n}$};
\draw (345,129.4) node [anchor=north west][inner sep=0.75pt]    {$\alpha $};

\end{tikzpicture}
\caption{Friction cone for a spacial (left) and planar (right) point contact.}
\label{fig:friction cone}
\end{figure}
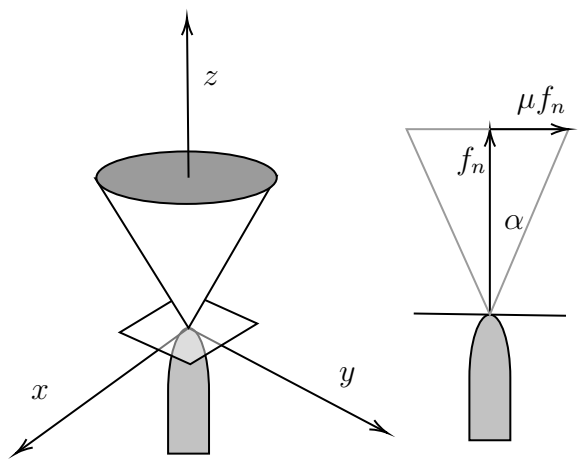

In simple terms, force closure refers to grasping an object with friction. Force equilibrium must be maintained such that the friction between the object and the finger compensates for the weight of the object and external wrenches. Friction is most commonly modeled using Coulomb friction \cite{b3c2715a826847b9b8b5282c93c1f325}, where the frictional force, $f_f$, and the normal force, $f_n$, are related by the friction coefficient, $\mu$, such that $f_f\le\mu f_n$. From the Coulomb friction model, the set of forces which can be transmitted through a frictional point contact can be represented by a friction cone (Fig. \ref{fig:friction cone}).

When an object is held with multiple points of frictional contact, force-closure occurs if the contact forces (normal force and friction), can balance out any arbitrary external wrench on the object. In the 2D planer case, at least 2 frictional contacts are required for force closure. Whether or not force-closure occurs is dependent on the contact position and friction coefficient.

\begin{figure}[!htbp]
\centering
\includegraphics[height = 5.25cm]{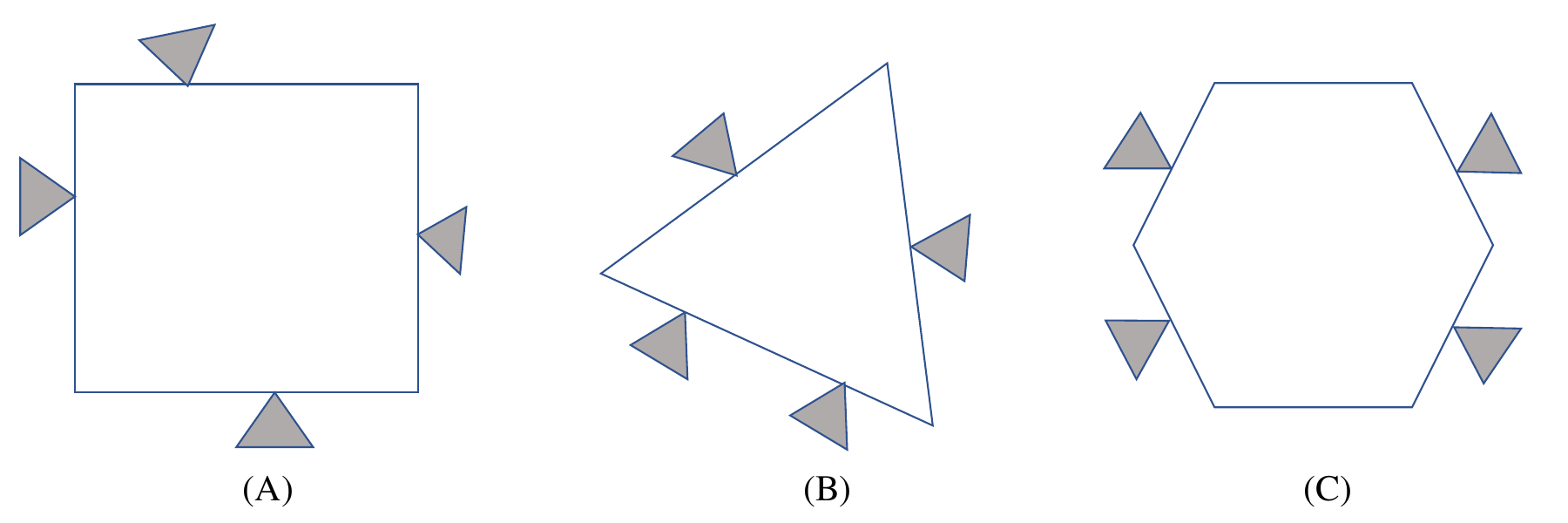}
\caption{Examples of form-closure grasps where the triangles represent contact points with the object.}
\label{fig:form closure examples}
\end{figure}

By contrast, form closure requires the hand to geometrically constrain the object with the contact points between the object and hand. Form closure in theory does not depend on friction to maintain stable grasps around an object. A form closure grasp is achieved when the constraints on the object set by the robot fingers prevent motion of the object \cite{b3c2715a826847b9b8b5282c93c1f325}. For the purposes of our hand, we will only be considering a planar motion for form closure. For 2D planar bodies, we need at least 4 points of contact to obtain form-closure, and at least 7 for an object in 3D space. Fig.~\ref{fig:form closure examples} depicts examples of form closure of a rigid body in planar space. The points of contact represented as triangles are stationary, and thus, the objects are fully constrained.

\chapter{Design of Quasi-Direct-Drive Robot Hand}

\section{Mechanical Design}
\label{sec:Mechanical Design}
Our approach presents a variety of design challenges. In order to minimize inertia of the finger, all motors should be mounted within the base of the robot hand, and not within the fingers. Therefore, there must be some way to transfer motion from the motors to the first and second joints of the finger. One way of doing this is to use a tendon/cable mechanism that pulls on the finger joints in order to drive them. However, a disadvantage of this approach is that it is difficult to reduce friction on the cable as it passes through the finger. Since the only torque/force feedback used is current measurement, friction should be reduced as much as possible, so that the motor torque has a linear relationship with the force output of the finger. Additionally, the gear ratio of the motor should be minimized to preserve the relationship between motor torque and current. Gearing also needs to be backdrivable and compliant to reduce the inherent stiffness of the overall mechanism. Therefore, we chose to use a simple 2.57:1 belt reduction connected to a differential gearbox at the base of the first finger joint (Fig.~\ref{fig:gears}). The belt reduction is highly backdrivable, and the bevel gears used in the differential gearbox have relatively low friction.

One problem with a quasi-direct drive, or proprioceptive actuation scheme is that the overall torque density of the actuators tends to be lower when compared to using motors with larger reduction gearboxes. Using a differential drive gearbox as we have implemented helps to compensate for this. The torque at each joint is shared between the two motors for each finger. This allows for higher possible torque at each joint, minimizing the disadvantage which comes with using a smaller gear ratio.

For position feedback, the motor shaft each contains a polarized magnet. The magnetic field is measured by magnetic encoders which send the angular position of each motor to a microcontroller to be used in the impedance control loop, and FOC control loop.

\subsection{Mechanical Design Iteration}

\begin{figure}[!htbp]
\centering
\includegraphics[scale=0.5]{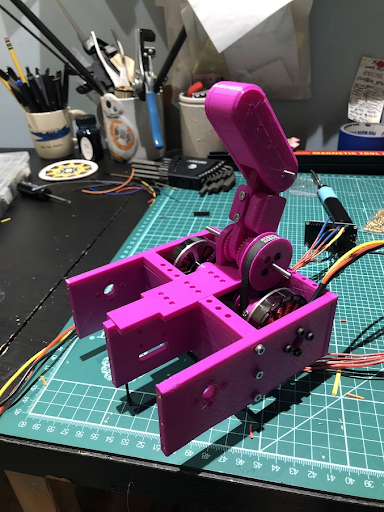}
\caption{Initial finger design prototype.}
\label{fig:prototype1}
\end{figure}

The first finger mechanism design iteration is shown in Fig. \ref{fig:prototype1}. In this design, the feasibility of the mechanical concept of using differential gears and belt drives to drive the joints was validated.

\begin{figure}[htp]
\centering
\captionsetup{width=6cm}
\subfloat[Second Finger Design Prototype]{\includegraphics[height=8cm]{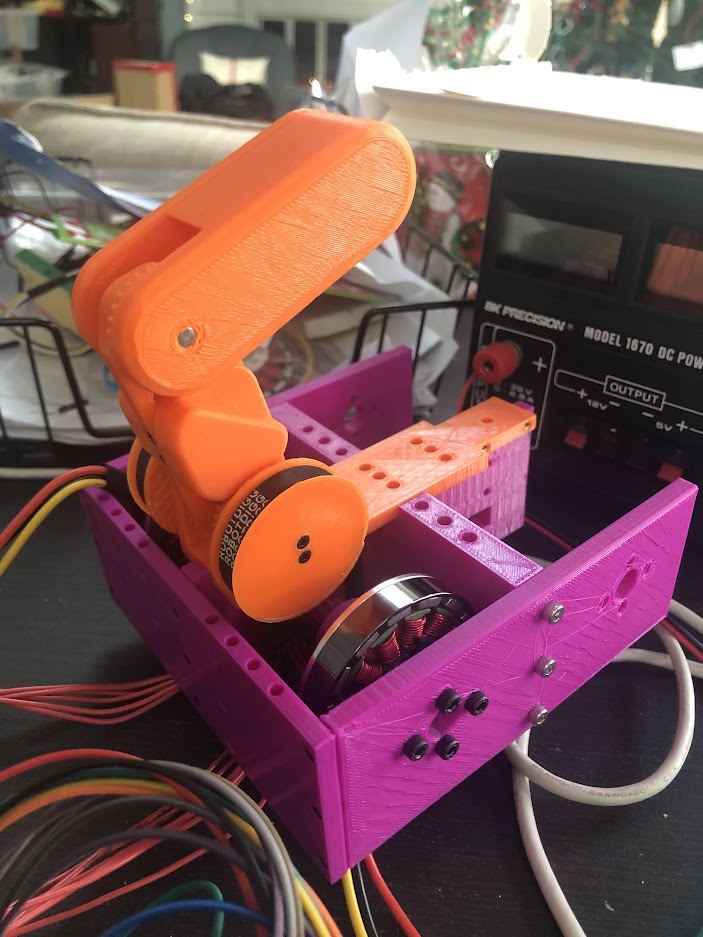}}%
\hspace{1cm}
\captionsetup{width=4cm}
\subfloat[Second Finger Design Section View]{\includegraphics[height=8cm]{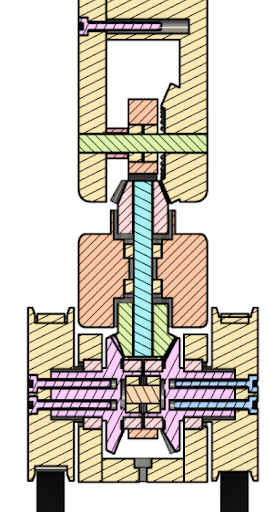}}%
\captionsetup{width=10cm}
\caption{Second iteration of robot finger design.}
\label{fig:prototype2}
\end{figure}

The second finger design iteration (Fig.~\ref{fig:prototype2}) increased the range of motion of the upper joint by approximately 45 degrees. Additionally, by using additional larger bearings in the differential gearing box, lateral play in the first joint was eliminated increasing the robustness of the mechanism. This is important for when the finger is under high load, and for when the hand must perform precise movements.

\begin{figure}[!htbp]
\centering
\includegraphics[height = 10cm]{Figures/Finger_Design_3.jpg}
\includegraphics[height = 10cm]{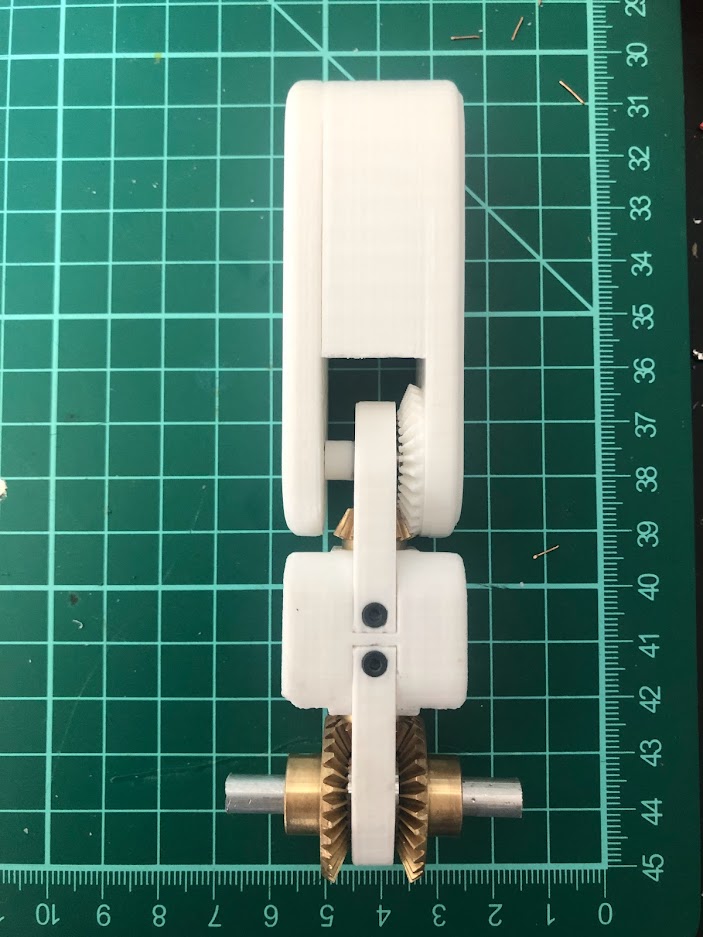}
\caption{Third and final finger design iteration, incorporating brass bevel gears.}
\label{fig:Finger3}
\end{figure}

One problem with this design was that during initial tests the 3D-printed gears in the differential gearbox at the bottom finger joint would skip teeth causing the finger to lose its position in relation to the motors. This is a problem because when the gears slip it reduces the gripping capacity of the hand, and the controller no longer knows the position of the finger. In order to prevent this the 3D-printed bevel gears were replaced with brass ones. This improved the structural rigidity of the mechanism by preventing gear slip and also reduced the overall friction in the joints because the dimensional accuracy of the brass gears is more precise than that of the 3D printed ones. Incorporating brass gears into the design required new bearings to fit onto the gears and a slightly larger housing for the bearings at the bottom finger joint (Fig. \ref{fig:Finger3}). The top bevel gear remained 3D printed, as the top joint did not experience any slippage in the bevel gears. This allowed the overall inertia of the finger to remain relatively low with the slightly heavier brass gears.

\subsection{Considerations for Encoder Mounting}
\begin{figure}[!htbp]
\centering
\includegraphics[scale=0.45]{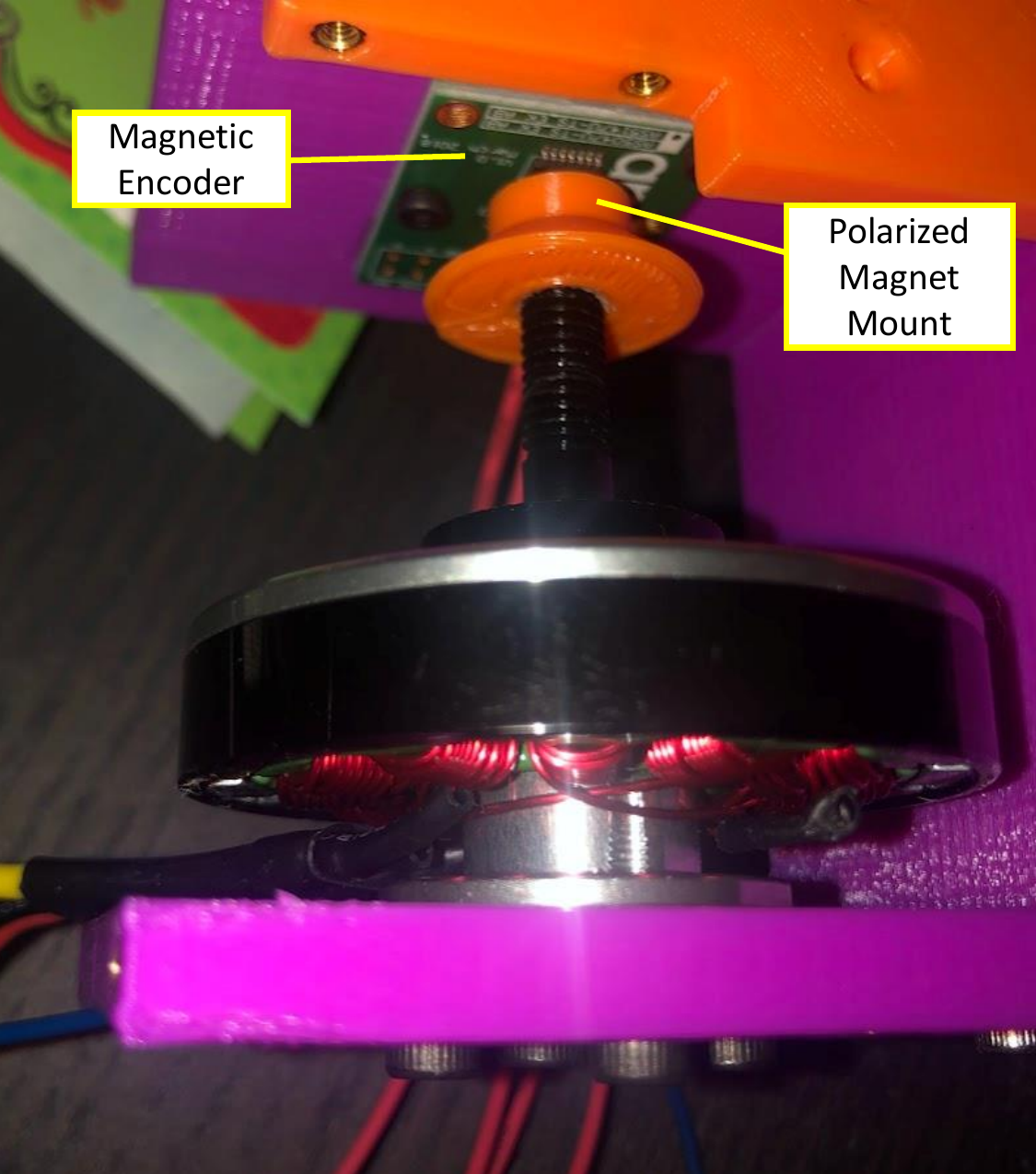}
\caption{Initial design magnetic encoder placement.}
\label{fig:encoder_v1}
\end{figure}

Precise motor position feedback is essential for both accurate FOC torque control and impedance control. Placement of the magnetic encoder can affect the accuracy of motor position feedback. At first, the polarized magnet for the encoder was mounted at the end of the motor shaft (Fig.~\ref{fig:encoder_v1}). This works, however, it results in some reduction in sensor accuracy. This is because small differences in the motor and encoder placements cause the magnet to rotate off-center in relation to the encoder. This eccentricity in the magnet rotation is not desirable. In addition, mounting the encoders in the center of the base of the hand makes wiring, installation, and sensor repair/replacement difficult.

\begin{figure}[!htbp]
\centering
\includegraphics[height = 6cm]{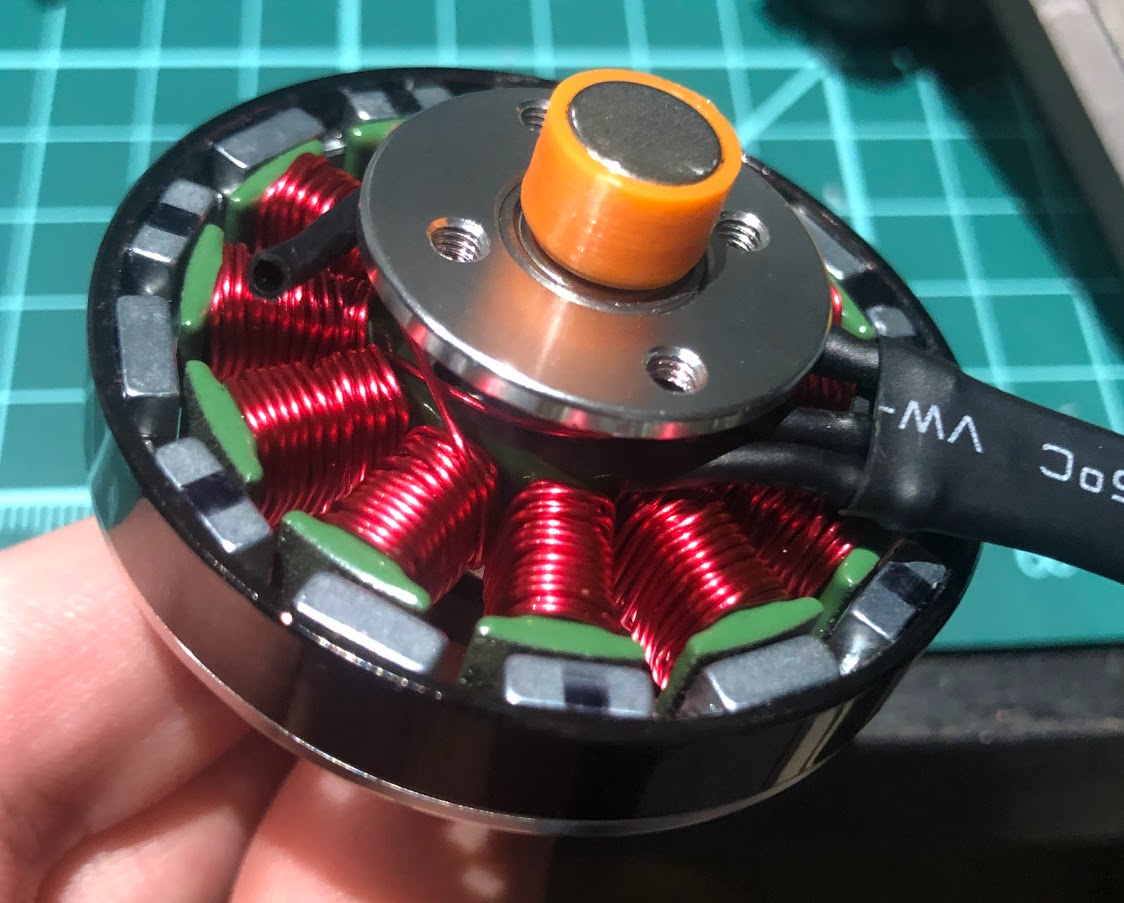}
\includegraphics[height = 6cm]{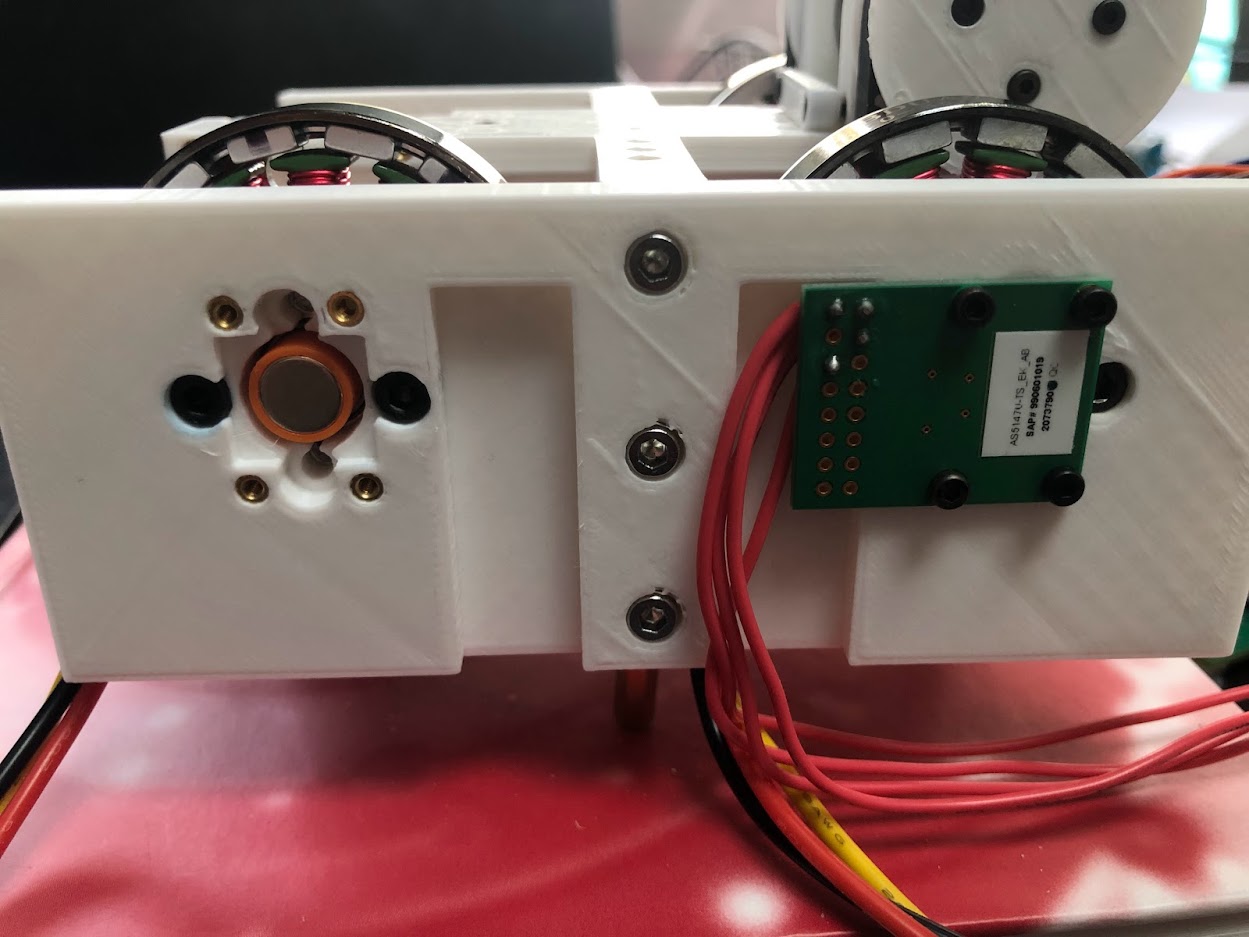}
\caption{Magnet mount for magnetic encoder - final design.}
\label{fig:mag}
\end{figure}

The solution to this problem is mounting the encoder magnet as close to the motor as possible in order to minimize the effect of errors in the magnet and sensor placements. Encoder magnets are mounted to the backside of the motor shaft with encoders mounted to the outside of the base of the hand (Fig.~\ref{fig:mag}).

\subsection{Finger Workspace}

\begin{figure}[!htbp]
\centering
\includegraphics[height = 12cm]{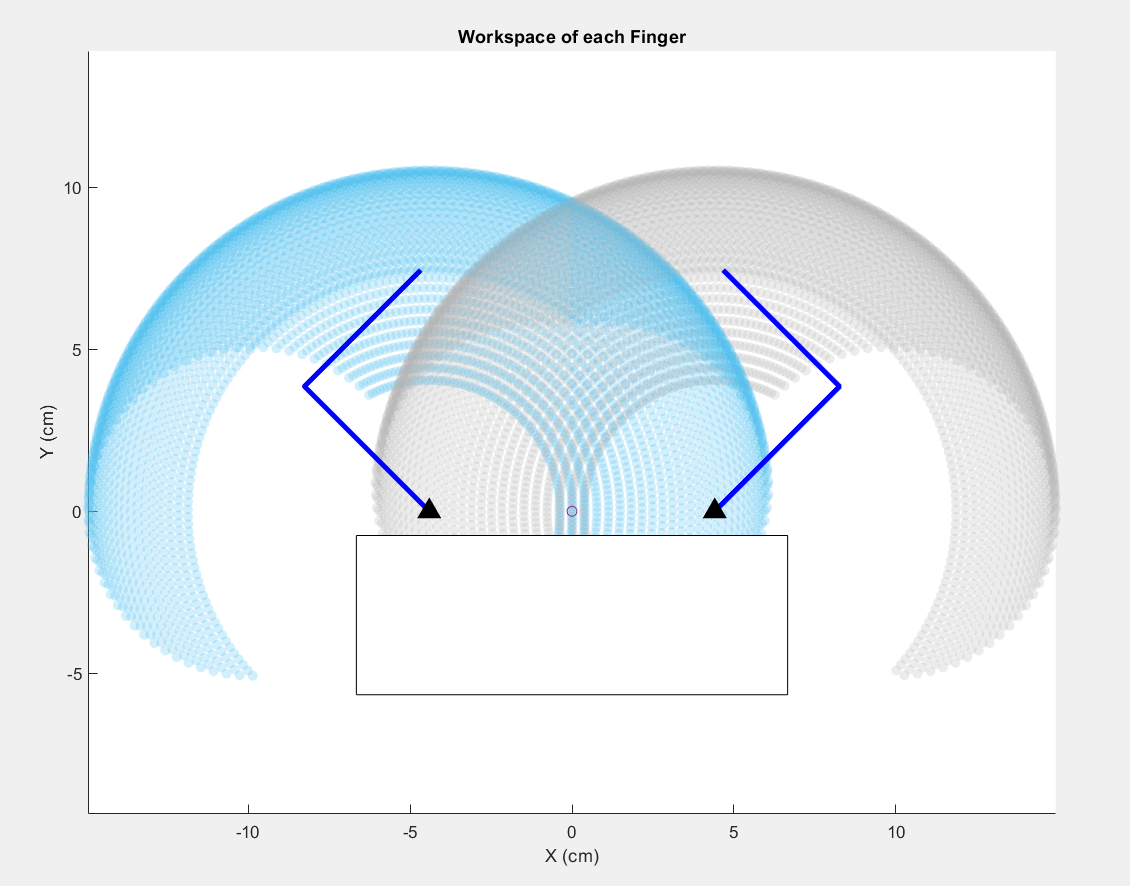}
\caption{Plotted workspace of each fingertip.}
\label{fig:work}
\end{figure}

An important aspect of the finger design is to attempt to maximize the reachable workspace of each finger. The workspace of each finger should also intersect so that the fingers can work together to grasp objects and perform tasks. Using the joint limits of the first and second joints of each finger, we can plot the reachable workspace of the fingertips using the forward kinematics model for a 2R serial manipulator. In radians, $\theta_1$ has joint limits $(0,3)$, while $\theta_2$ has joint limits $(-\pi/2,3\pi/4)$. The results are shown in Fig. \ref{fig:work} with the left and right fingertip workspace plotted in blue and grey, respectively, along with a simplified diagram of the hand.

\section{Electrical Hardware and Control Architecture}

\begin{figure}[!htbp]
\centering
\includegraphics[scale=0.8]{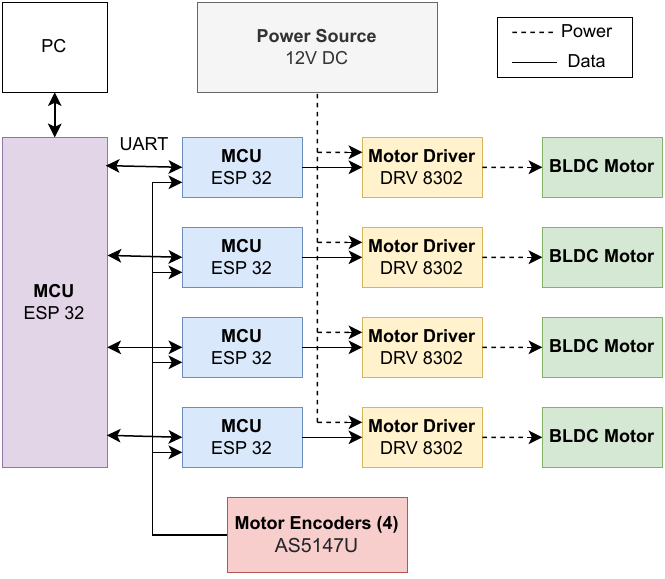}
\caption{Original plan for control architecture.}
\label{fig:Control architecture}
\end{figure}

\begin{figure}[!htbp]
\centering
\includegraphics[scale=0.2]{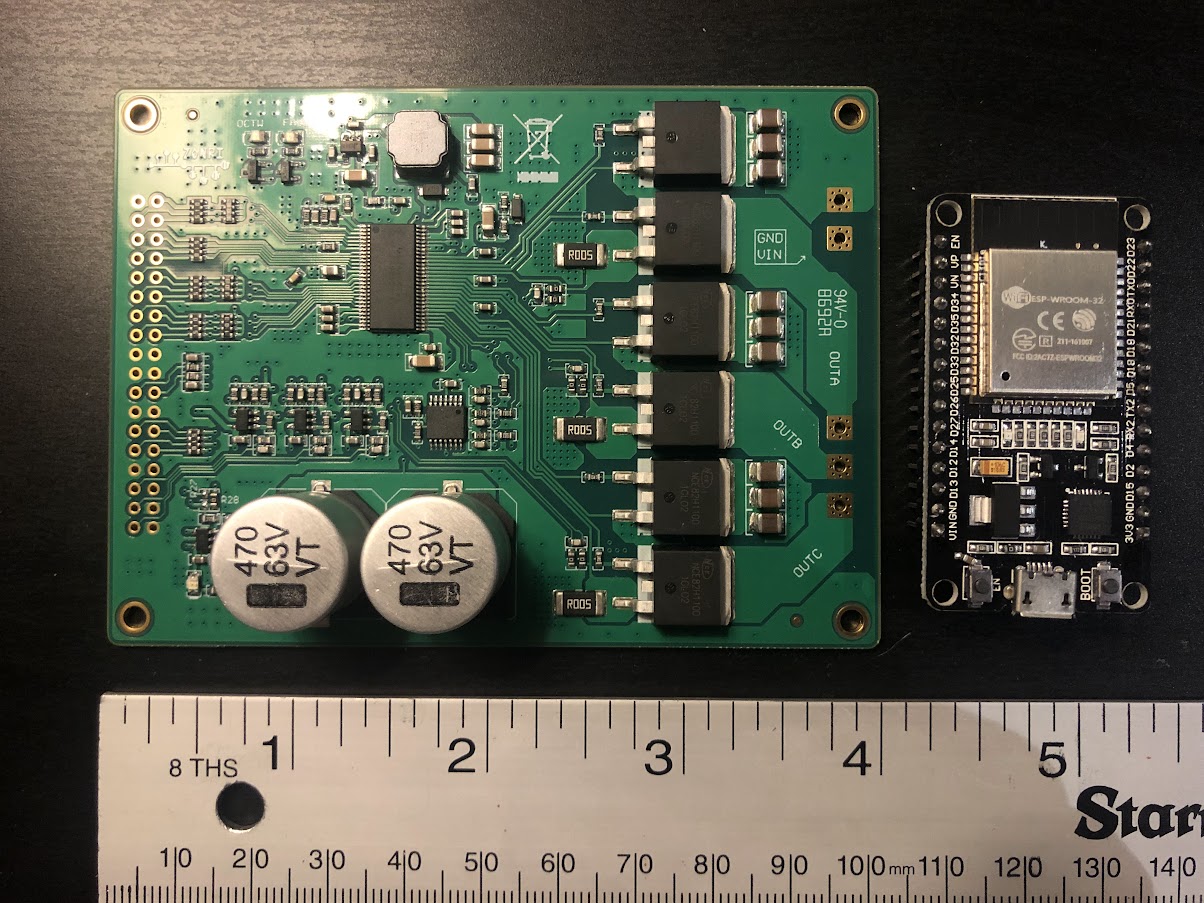}
\caption{DRV 8302 Motor Driver (left) and ESP32 microcontroller (right).}
\label{fig:Electronics}
\end{figure}

The original plan for the system control architecture was to have each motor controlled by its own ESP32 microcontroller (MCU) as shown in Fig.~\ref{fig:Control architecture}. The microcontroller would be connected to a DRV8302 motor driver module (Fig.~\ref{fig:Electronics}), each of which is connected to a 12V DC power source and each respective motor. It would be impossible to have the motors all driven by the same microcontroller, as each motor driver requires too many input/output pins to operate. Magnetic motor encoders send motor position data to each motor MCU which communicates with an additional central controller via UART communication. Unfortunately, the serial communication of motor angle and torque values introduced too much lag in the program execution, resulting in the system becoming unstable causing the finger to oscillate uncontrollably. This method has the benefit of being scalable to any amount of motors, however, our original program implementation was far too slow for this application. In the future, should we wish to add additional degrees of freedom to each finger, a control architecture similar to this would be necessary to fully implement.

\begin{figure}[!htbp]
\centering
\includegraphics[scale=0.8]{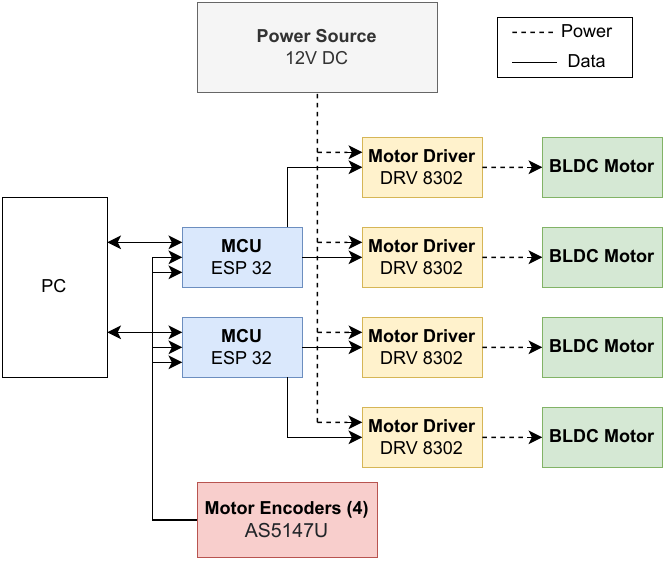}
\caption{Final Diagram of the control architecture.}
\label{fig:Control architecture2}
\end{figure}

Instead, in the end, we went with a much simpler approach for the control of the robot hand. The two motors for each finger are controlled by a single ESP 32 microcontroller. This reduces the total amount of MCUs needed for the hand while eliminating the need for complex communication protocols between devices. Fig. \ref{fig:Control architecture2} shows the final control architecture used in the hand. Trajectory planning for each of the two fingers is programmed separately. The PC is used for uploading code and data acquisition.

\section{Motor Selection}
\begin{figure}[!htbp]
\centering
\includegraphics[scale=0.95]{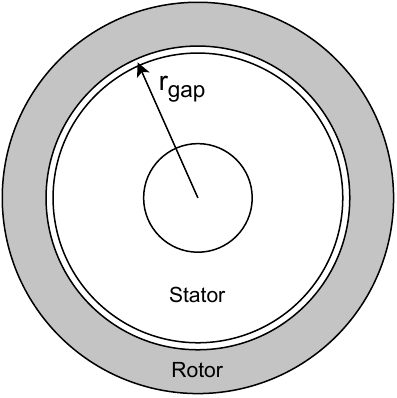}
\caption{Simplified diagram of an outrunner BLDC motor.}
\label{fig:mot_diagram}
\end{figure}

There were several factors that went into the selection of a suitable motor for our application. Previous designs incorporating DD and QDD actuators use BLDC motors due to their increased torque density compared to brushed DC motors while also running with greater efficiency. This allows the hands to remain functional with much lower gear reductions. BLDC motors can be broadly classified into two groups: inrunner and outrunner motors. Inrunner motors place spinning permanent magnets on the interior of the motor surrounded by the stator, containing the motor electromagnets. For outrunner motors the reverse is true: the rotor with permanent magnets spins on the outside of the motor stator. While inrunner motors have lower inertia than outrunner motors, outrunner motors have higher stall torque density, motor constant, and efficiency \cite{5979940}. This makes them the better candidate for the QDD actuators in our application where we are more concerned with torque output than the reflected inertia of the motors. Fig. \ref{fig:mot_diagram} shows a diagram of an outrunner BLDC motor, where $r_\mathrm{gap}$ is the gap radius. The gap radius is defined as the distance from the center of the motor to the center of the gap in between the stator and the rotor.

\begin{figure}[!htbp]
\centering
\includegraphics[scale=0.25]{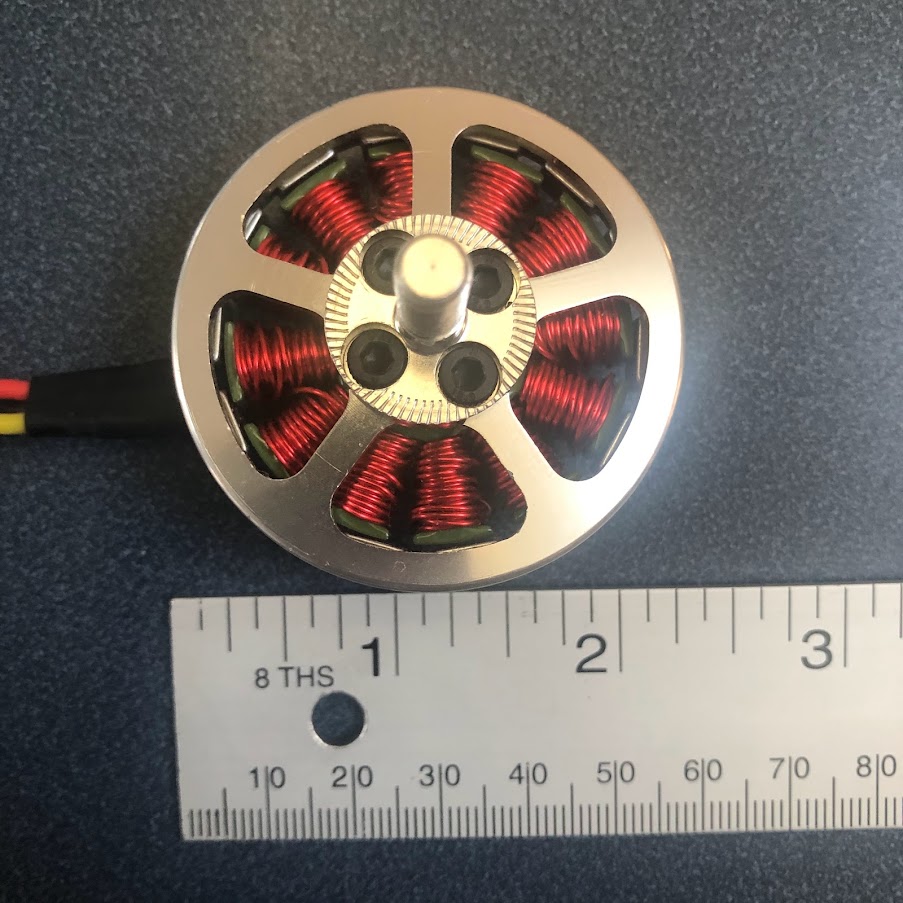}
\caption{Motor used in QDD hand with a ruler for scale.}
\label{fig:mot_pic}
\end{figure}

Using the parameter gap radius we can make some generalizations about the motor properties. In general,

\begin{center}
$m\sim{r_\mathrm{gap}}$\\
$\tau\sim{r_\mathrm{gap}}^2$\\
$I\sim{r_\mathrm{gap}}^3$  
\end{center}
where $m$ is mass, $\tau$ is torque, and $I$ is rotor inertia \cite{biomimetics.mit.edu}. From these equations, we can also say that torque density, or torque divided by mass, is similar to the gap radius of the motor. Therefore, in our motor selection, we attempt to maximize the radius of the motor. Of course, we cannot pick too large of a motor as we do not want the rotor inertia to become too large, and we still need to pick motors which can fit within the structure of our hand. Keeping these factors in mind, we selected the motor pictured in Fig. \ref{fig:mot_pic}. The motor is a BDUAV 5010 360KV rated motor which can be purchased on Amazon for less than 20 dollars. The KV rating is the motor RPM per volt.

One point to note is that the market for affordable outrunner BLDC motors is dominated by motors designed for applications in drones, remote-controlled airplanes, and other small unmanned aerial vehicles. These motors are designed for use at very high angular velocities, much higher than would be experienced in our hand. In aerial vehicles, they also have the advantage of having air constantly being forced over them by a propeller, providing lots of active cooling which prevents the motors from overheating even at high currents. In our hand, the motors do not have this luxury which is part of the reason why we use current sensing as a means to ensure that the motors do not overheat in addition to controlling the motor torque.

\section{Final Design Summary}
\begin{figure}[!htbp]
\centering
\includegraphics[scale=0.52,angle=90,origin=c]{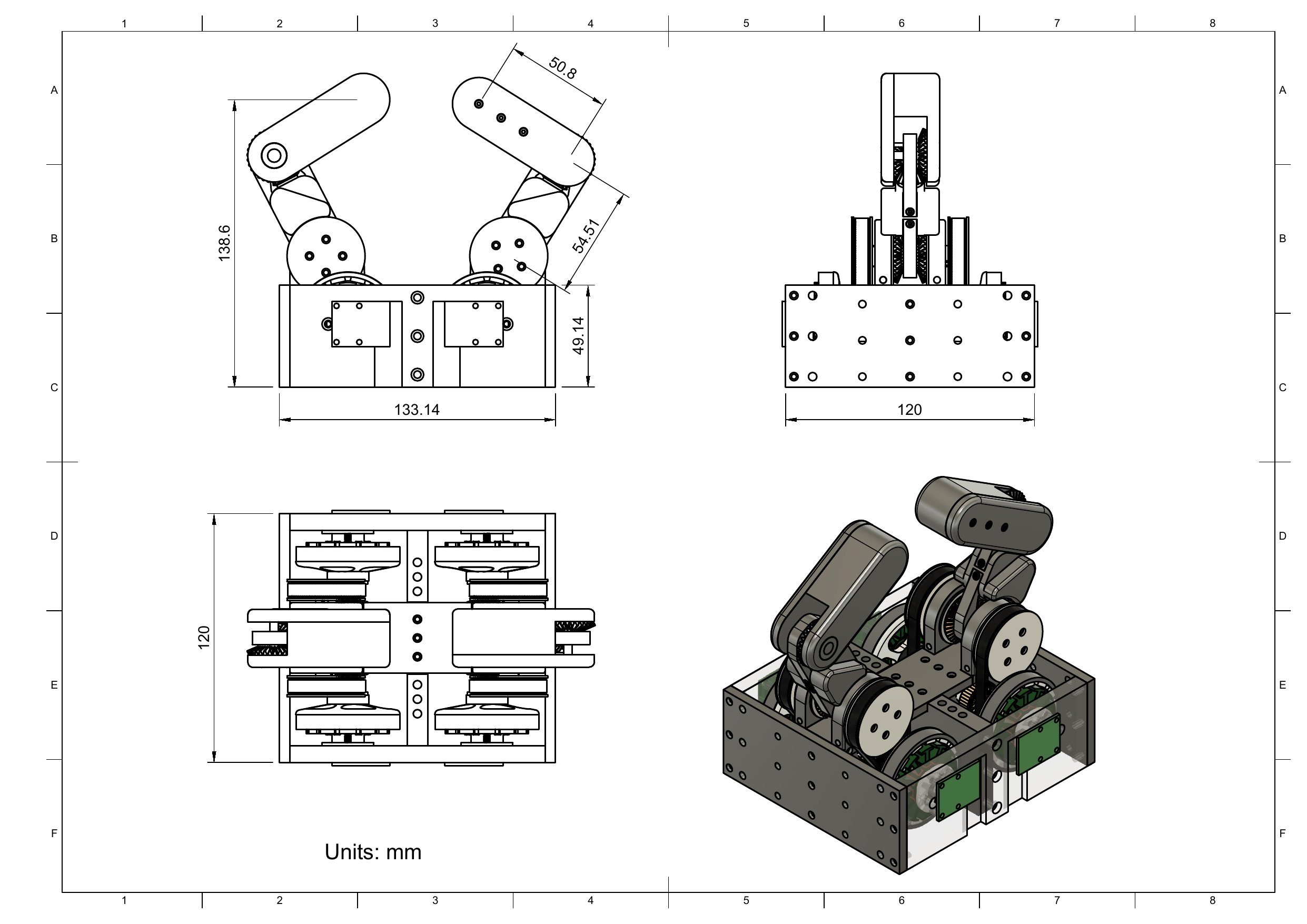}
\caption{Final top-level assembly drawing.}
\label{fig:drawing}
\end{figure}

\begin{figure}[!htbp]
\centering
\includegraphics[height=20cm]{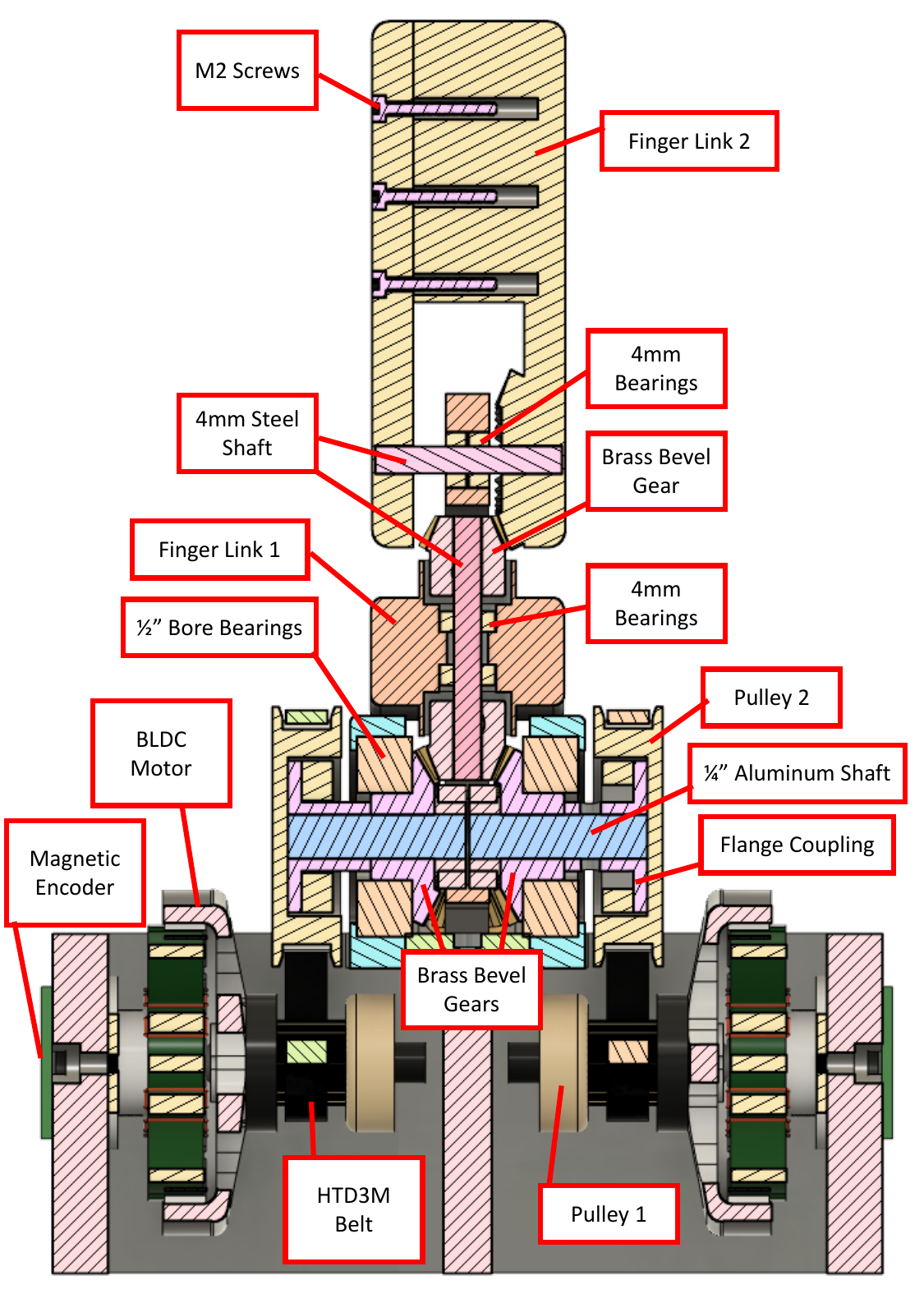}
\caption{Labeled section view of the finger.}
\label{fig:section view}
\end{figure}

This section gives an overview of the completed QDD robot hand design. Table \ref{table:design specs} presents some of the pertinent design specifications of the hand. Fig. \ref{fig:drawing} shows a drawing with dimensions of the completed hand CAD assembly. Fig. \ref{fig:section view} depicts a labeled section view of the finger. In this figure, all of the internal components of the mechanism are shown such as the bearings, metal shafts, belts, and pulleys.

\begin{table}[htbp]
\caption{Design Specifications.}
\label{demo-table}
\centering
\begin{tabular}{ |c|c| } 
\hline
 \textbf{Design Parameters} & \textbf{Value} \\
\hline
 Force Output at Fingertip &  8.2N      \\
 Weight                    &  1250 g      \\
 Motor Weight              &  92 g      \\
 Motor KV Rating           & 360KV   \\
 Degrees of Freedom        & 4       \\
 Operating Voltage         & 12V       \\
 Encoder Resolution        & 16384CPR      \\
 Total Cost                &  $\sim$\$400       \\
 \hline
\end{tabular}
\label{table:design specs}
\end{table}

\begin{figure}[!htbp]
\centering
\includegraphics[scale=0.35]{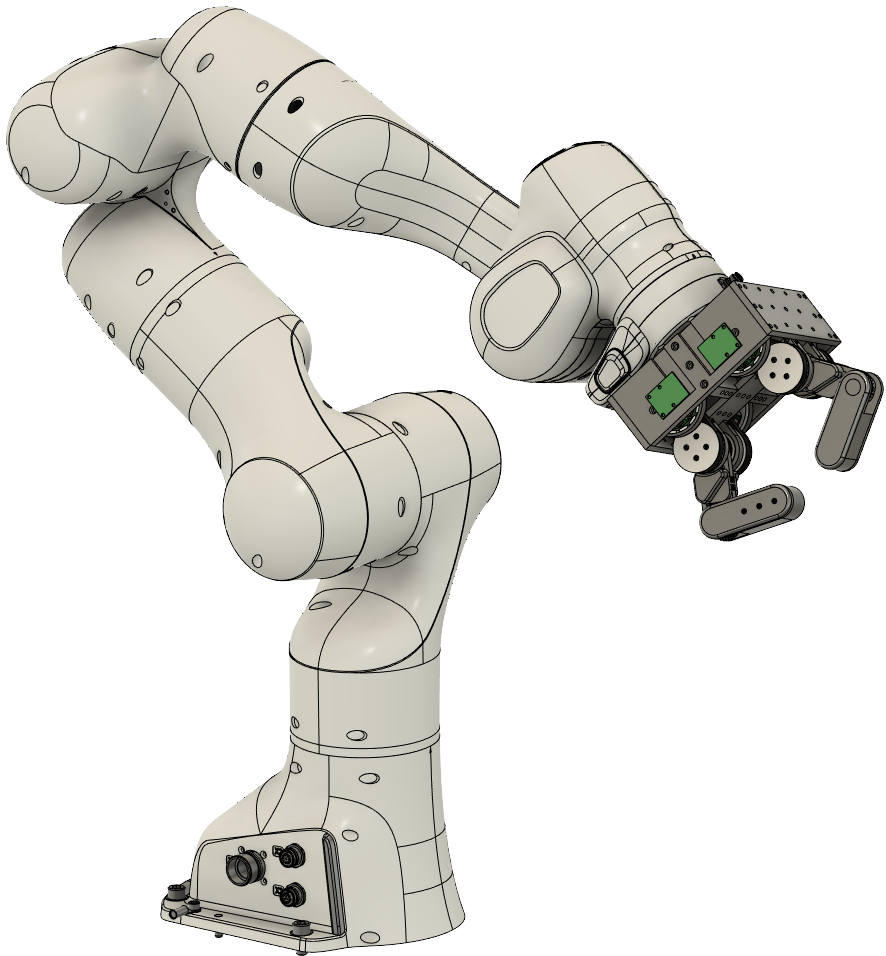}
\caption{CAD rendering of the QDD robot hand mounted to the Franka Emika Panda robot arm.}
\label{fig:panda}
\end{figure}

In the future, we plan on mounting our hand to the Franka Emika Panda robot arm \cite{franka} in order to further test the capabilities of the hand. A CAD rendering of what this might look like can be seen in Fig. \ref{fig:panda}.

\chapter{Experiments and Results}

\section{Motor Testing}

\begin{figure}[!htbp]
\centering
\subfloat[]{\includegraphics[width=9cm]{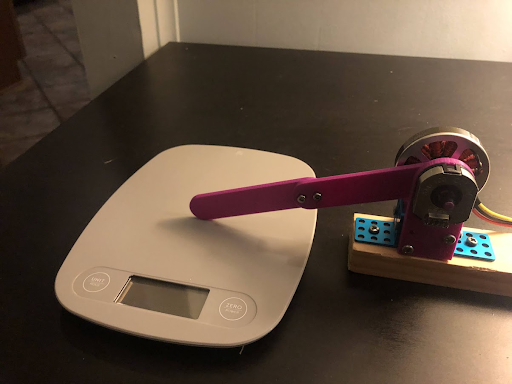}}\\
\subfloat[]{\includegraphics[width=12cm]{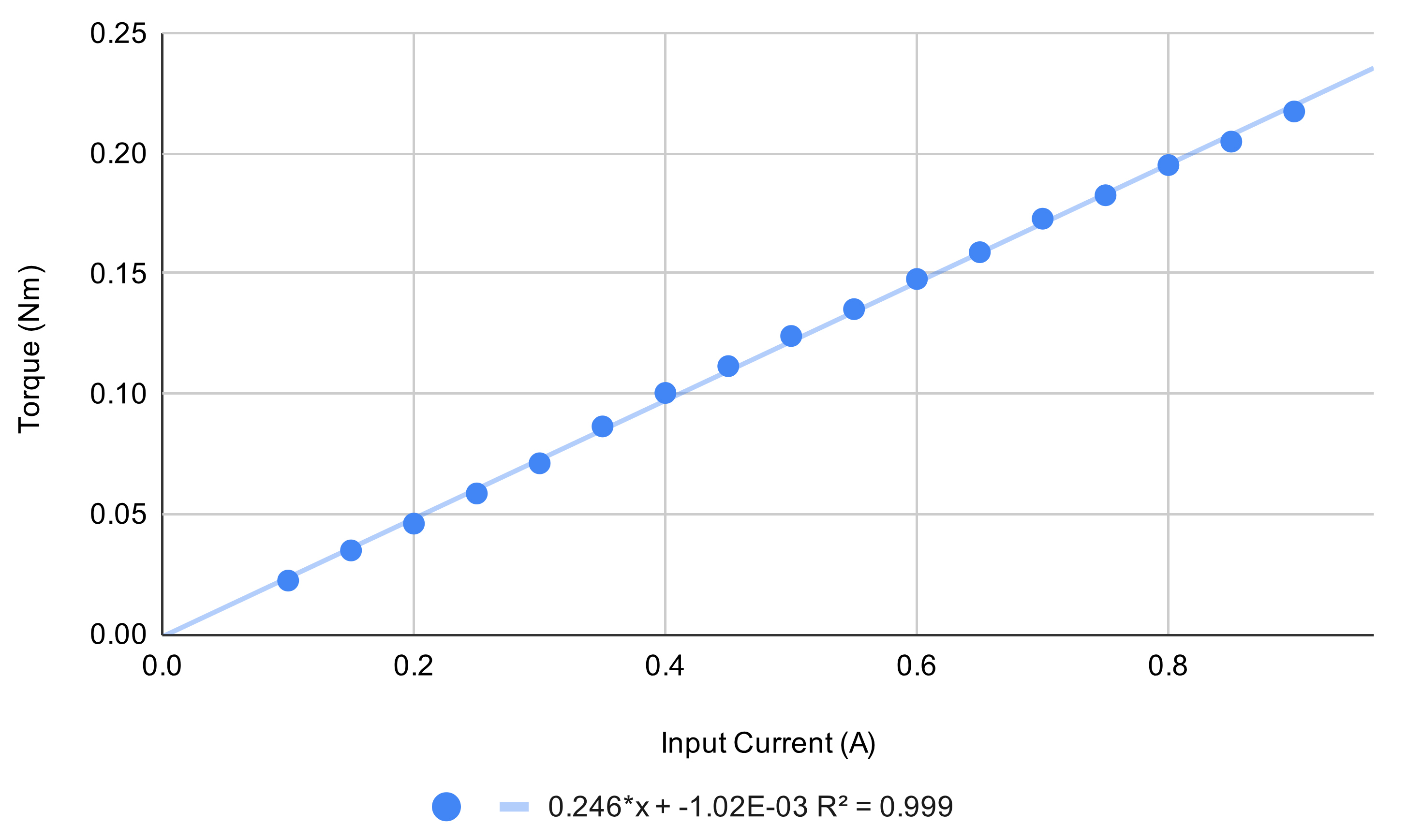}}
\caption{Motor testing, (a) Test setup, (b) Plot of motor force with respect to the Input current.}
\label{fig:Motor Test/torque}
\end{figure}

Initially, we performed several experiments to verify the viability of impedance control for grasping in a simple application, as well as test different electronic hardware, motors, and control schemes. Most affordable BLDC motors on the market are designed for applications in hobby remote-controlled planes and drones, where the motors run at extremely high RPM. Therefore, these motors frequently do not come with specifications applicable to much lower-speed robot applications. One of the first things we tested was the torque output of the motor we selected and plotted with respect to the measured motor winding current. The slope of this plot is equal to the motor torque constant. Fig. \ref{fig:Motor Test/torque}-A shows a testing rig designed to measure the output torque of the motor using an arm of known length pressing on a digital scale. Using this setup, we can test the accuracy of different methods of torque control. As expected, using the FOC-based torque controller yields a strong linear relationship between motor current and torque (Fig. \ref{fig:Motor Test/torque}-B). At the maximum torque we tested, the motor ran very hot, so we stopped the test here as we did not want the motor to overheat, and the torque we knew the torque would be sufficient for our application. However, should we want to increase the torque with this motor, in the future we could incorporate active cooling methods such as a fan over the motor.

\section{A Simple Parallel Jaw Gripper}
Now that we have verified successful control of the BLDC torque by measuring current, we can apply it to a simple manipulation system: the parallel jaw gripper. Using a rack and pinion mechanism the rotation of the motor is turned into translation of the gripper. While not as versatile as a multi-fingered hand, the parallel jaw gripper allows us to test the impedance control scheme. The actuator is quasi-direct-drive, with no gearbox aside from the rack and pinion. This means that the mechanism is easily back-driven and has low friction. With low frictional losses, motor current feedback can be used to control the force output of the gripper. A rotary encoder provides motor position feedback to the controller. The gripper also uses the same motor as the QDD hand, which lets us test and verify that the motor torque would be sufficient for grasping applications.

\begin{figure}[htp]
\centering
\subfloat[]{\includegraphics[scale=0.4]{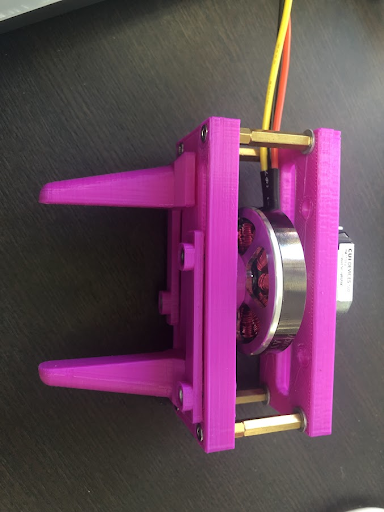}}%
\hspace{0.1cm}
\subfloat[]{\includegraphics[scale=0.4]{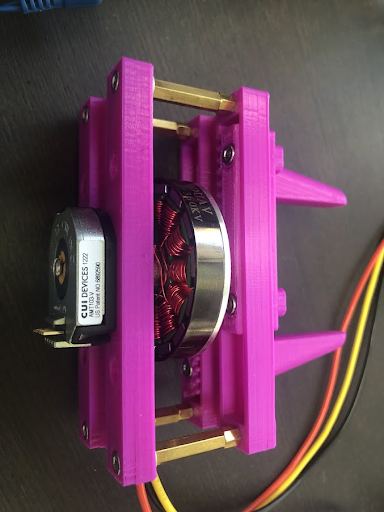}}%
\caption{Parallel jaw gripper prototype.}
\end{figure}

\section{Fingertip Force Output}
\begin{figure}[!htbp]
\centering
\subfloat[]{\includegraphics[height = 6cm]{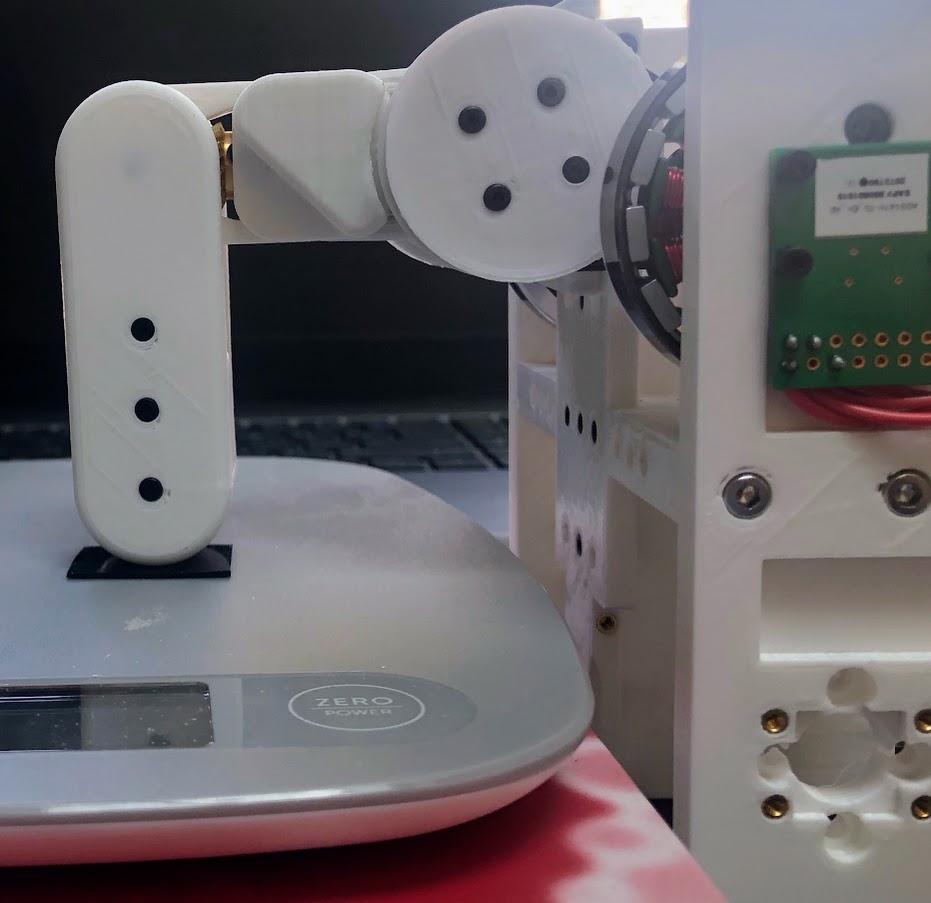}}
\subfloat[]{\includegraphics[height = 6cm]{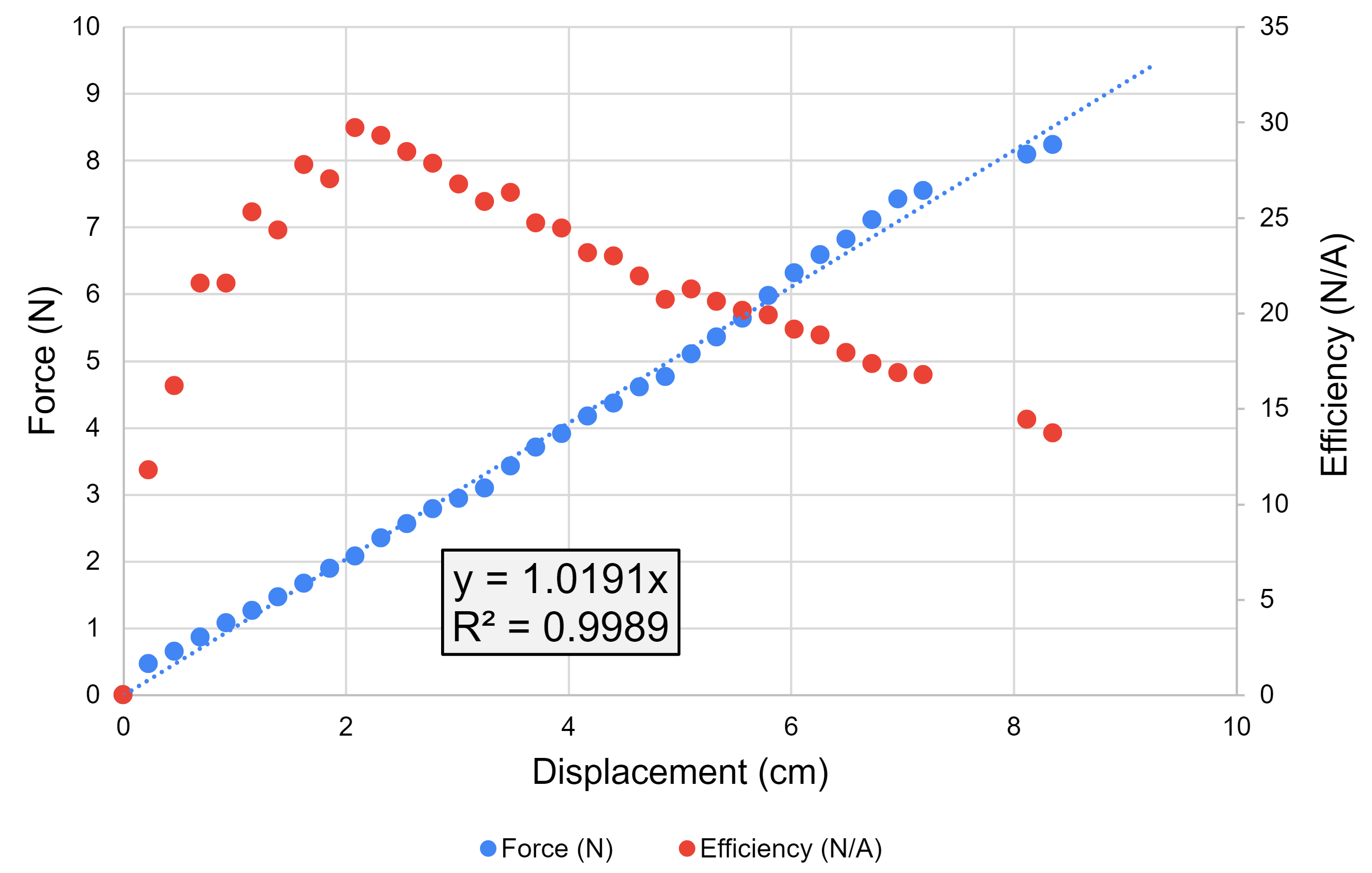}}\\
\caption{Fingertip force output, (a) Test setup, (b) Plot of fingertip force with respect to displacement.}
\label{fig:Plot1}
\end{figure}

In order to successfully implement the impedance controller to grasp objects, the force output at the fingertip of the hand must be robustly controlled. In our case, this is done through FOC-based torque control to control motor torque and thus also fingertip force. A test setup and a plot of force with respect to displacement are shown in Fig.~\ref{fig:Plot1}. The fingertip presses vertically on a scale which measures the force output. Displacement is measured in centimeters as the difference between the actual and desired positions in the direction orthogonal to the scale. The slope of this graph is expected to be equal to the stiffness coefficient, $K$, of the impedance controller. For this test, we used a stiffness coefficient of 1 in the vertical direction which was very close to the experimental slope of 1.02. This indicates our system works as intended, and that the friction in the motor gearing and drive-train is minimal. The maximum force output we tested was 8.2N, which is sufficient for the tasks we plan on completing with the hand.

The efficiency of the force output is also plotted in Fig.~\ref{fig:Plot1}. Here, efficiency is taken to be the force output divided by the total current draw of the system as measured by the 12V DC power supply. The motors run most efficiently where the force output is just over 2N. While not possible for all tasks, it would be beneficial to run the motors within this region of force output. This decrease in efficiency at higher force is the result of increased losses in the motor driver, not decreased efficiency in the actual motor. The relationship between motor torque and motor current draw stays constant even at high current/torque values. 

\section{Trajectory Tracking}

\begin{figure}[!htbp]
\centering
\subfloat[]{\includegraphics[scale = 0.75]{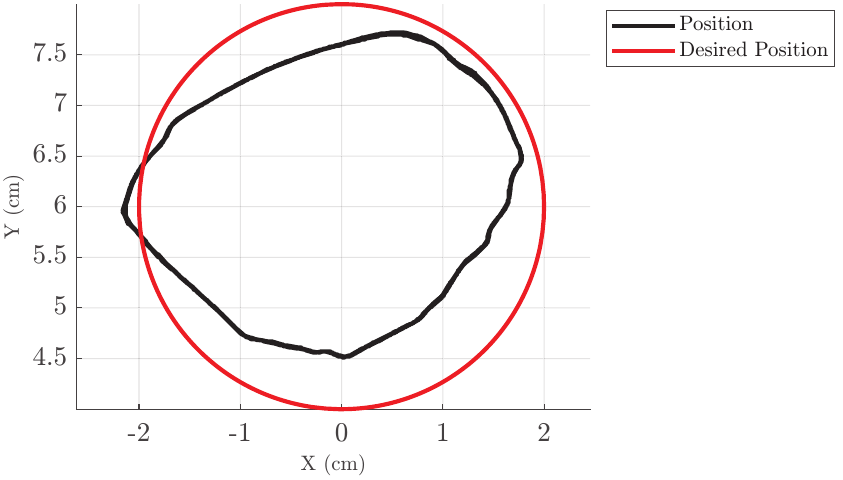}}\\
\subfloat[]{\includegraphics[scale = 0.6]{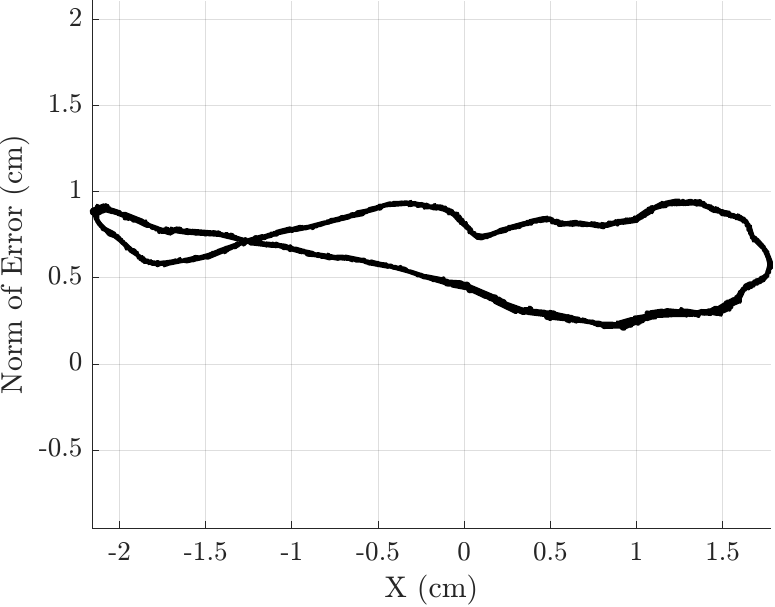}}
\caption{(a) Plot of fingertip circular trajectory (b) Plot of fingertip position error.}
\label{fig:circle}
\end{figure}

To perform functional tasks, the fingers must be able to follow along a desired trajectory reasonably well. This ability gives our hand maximum flexibility making it more useful as a general-purpose hand, compared to a parallel jaw gripper for example, which can only open and close. Trajectory tracking in the fingers is especially important for in-hand dexterous manipulation tasks, where the motion must be relatively precise in order to obtain a desired outcome. Fig.~\ref{fig:circle} shows the finger tracking a circular trajectory with Cartesian impedance control. Desired values of $X$ and $Y$ are generated simply by setting the desired $X$ equal to $2\cos(t)$ and desired $Y$ equal to $2\sin(t)+6$ where $t$ is equal to time and fingertip position is measured in centimeters. The total error of the finger as measured by the magnetic encoders stays under 1 cm for all trajectory positions. While this can be improved with modifications to the controller, hardware, and finger design, this positional error will be acceptable for most of the tasks we plan on performing. 

\begin{figure}[!htbp]
\centering
\includegraphics[scale = 0.85]{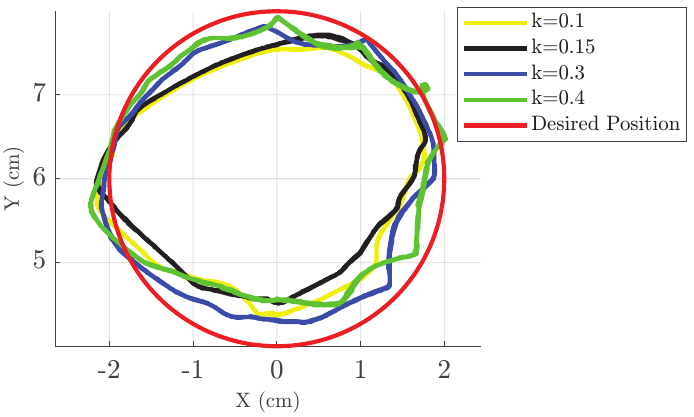}
\caption{Circular trajectory with varying stiffness and constant damping.}
\label{fig:circle2}
\end{figure}

One thing we can do to improve the accuracy in the trajectory of the fingertip is to increase the stiffness coefficients $k_x$ and $k_y$. This can be seen in Fig.~\ref{fig:circle2}. In general, the error decreased with increasing stiffness coefficients as expected. At the highest value of stiffness tested, the finger experiences some oscillations during portions of the trajectory as a result of the finger overshooting its desired trajectory. This can be improved by increasing the damping coefficient at higher stiffness values.

\begin{figure}[!htbp]
\centering
\includegraphics[scale=0.8]{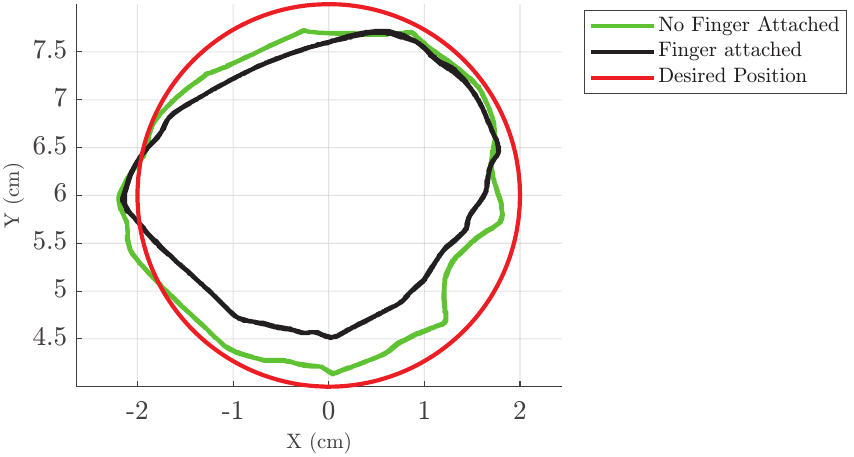}
\caption{Circular trajectory finger mechanism not attached.}
\label{fig:circle no finger}
\end{figure}

While the fingers are designed to have very low friction, it is impossible to eliminate friction entirely. Fig. \ref{fig:circle no finger} helps determine the effect of friction and finger mass on the accuracy of the finger trajectory. The trajectory of the finger from Fig.~\ref{fig:circle}-A is plotted along with the plot of the controller running the same code with the finger mechanism not attached and the motors running freely. As shown in the figure, without anything attached, the trajectory is closer to the desired trajectory, but there is still some error. This is due to the relatively high cogging torque of the motors used causing them to run un-smoothly at times as a result of torque ripple. This effect is most noticeable when running the motors at lower speeds. Again, the resultant error, as a result, is relatively small and yields results good enough for our applications. However, the effect of cogging torque ripple can be improved by selecting more expensive motors, or by improved software control through calibration \cite{inproceedings}. Improving torque ripple in software would be a useful improvement for future work.

\begin{figure}[!htbp]
\centering
\subfloat[]{\includegraphics[scale = 0.8]{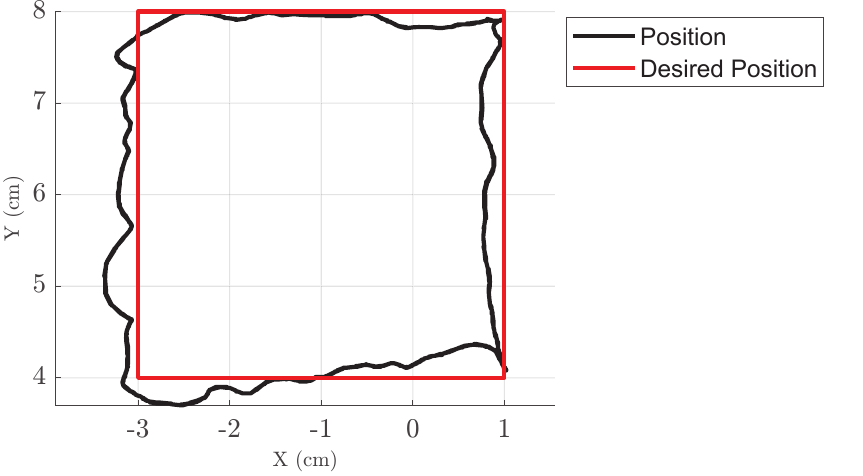}}\\
\subfloat[]{\includegraphics[scale = 0.6]{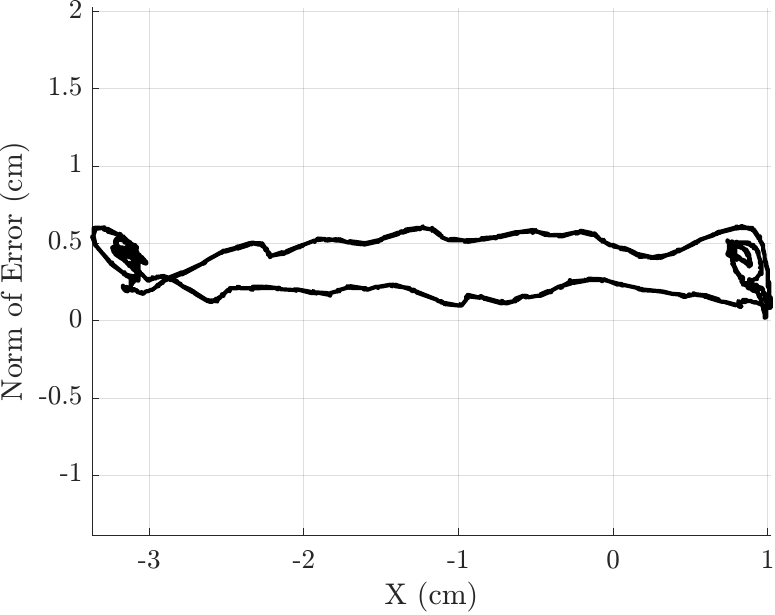}}\\
\caption{(a) Plot of fingertip rectangular trajectory (b) Plot of fingertip position error.}
\label{fig:rectangle}
\end{figure}

In addition to circles, the fingertip can track any arbitrary path in its workspace. Fig. \ref{fig:rectangle} shows the finger tracking a rectangular trajectory with minimal error.

\section{Evaluation of Impedance Controller Special Case}
In Sec.~\ref{sec:Application to the Robot Finger}, we discussed the special condition for the Cartesian Impedance controller where $\theta_2<0$. In this section, we will evaluate the effectiveness of modifying the impedance controller using equation \eqref{tau mod} to prevent undesired configurations of the finger. For comparison, a video of the controller without this modification can be found here \url{https://youtu.be/HuPaaSSRv9Q}. When the finger is manipulated so that $\theta_2$ is negative, the finger gets ``stuck" in this undesired position. With the modified controller, the finger is prevented from moving to this position, but otherwise behaves the same as the standard Cartesian impedance controller. A video of this controller in action can be found here: \url{https://youtu.be/MrbVwZfQA1o}.

\section{Grasping Objects}
\subsection{Force-Closure Grasp}
\label{sec:force closure}
\begin{figure}[!htbp]
\centering
\subfloat[]{\includegraphics[scale=0.05]{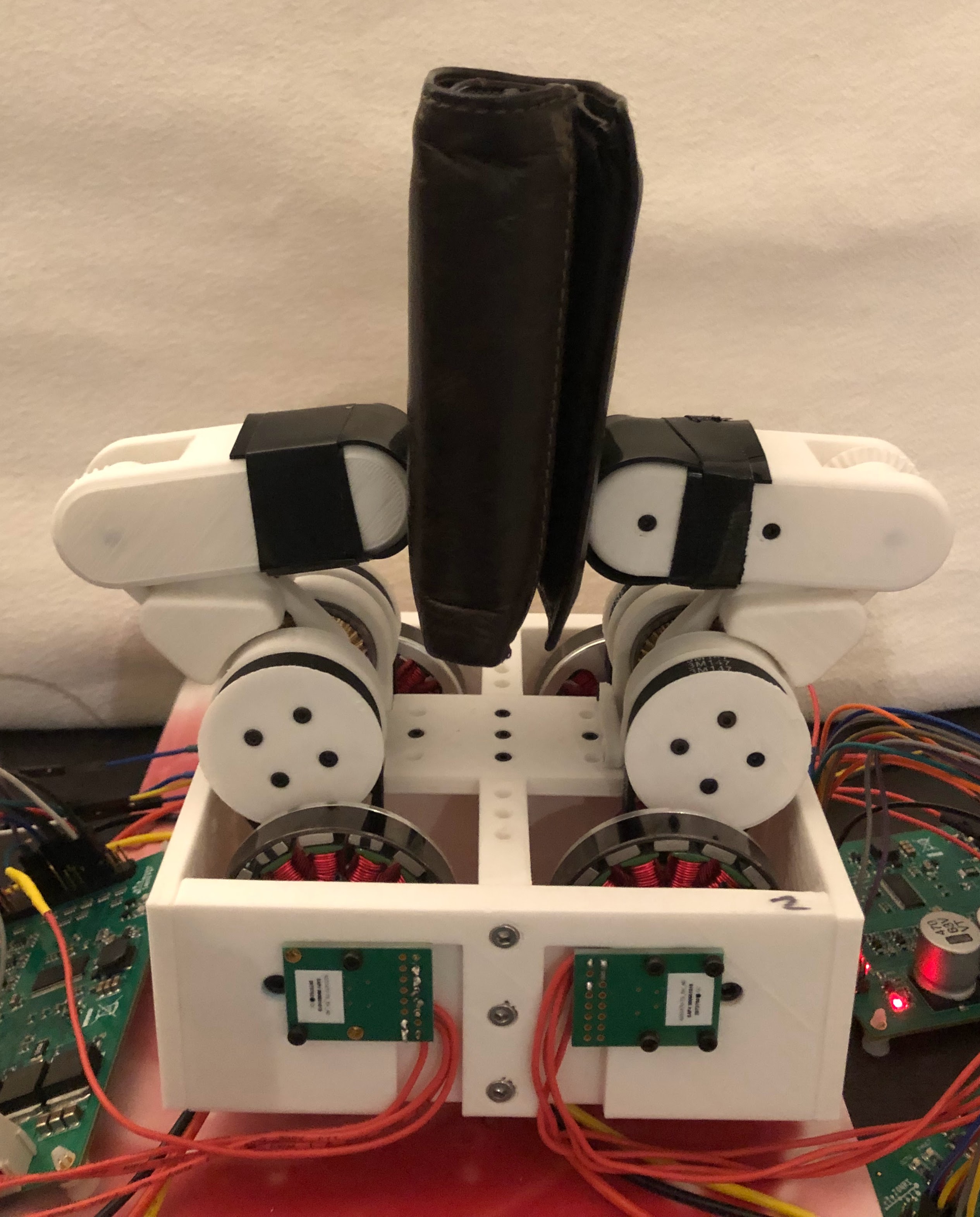}}
\hspace{0.5cm}
\subfloat[]{\includegraphics[scale=0.15]{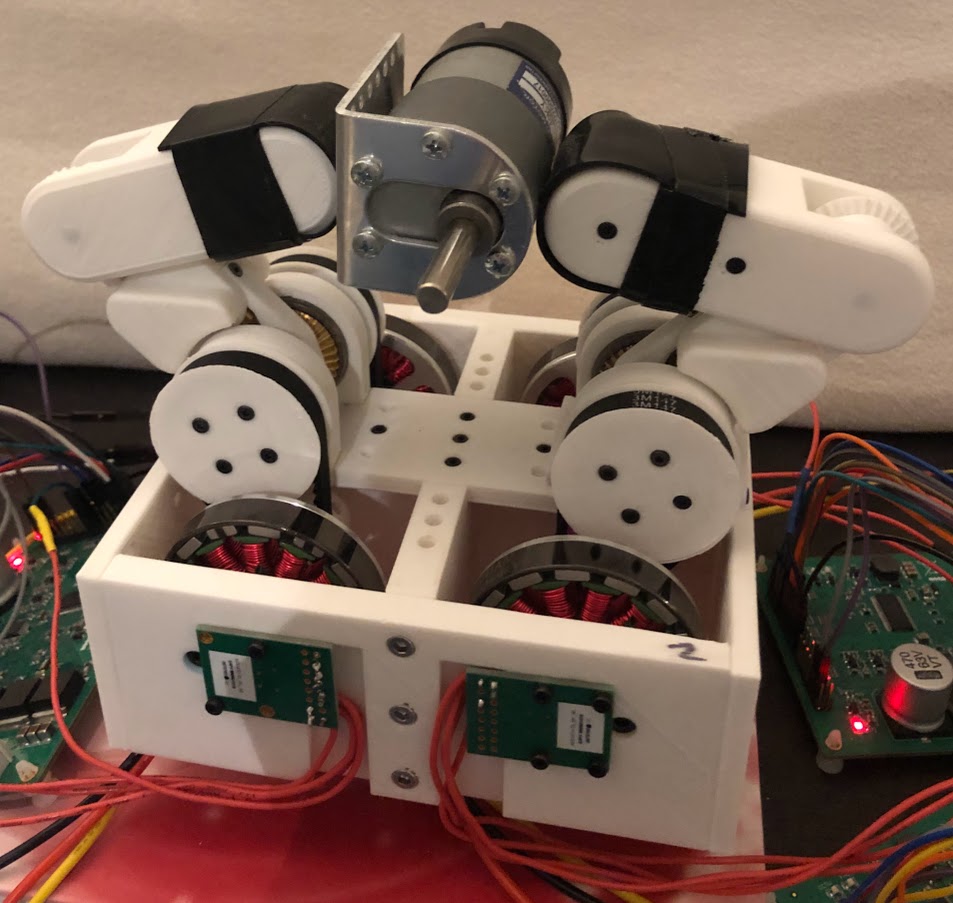}}
\hspace{0.5cm}
\subfloat[]{\includegraphics[scale=0.2]{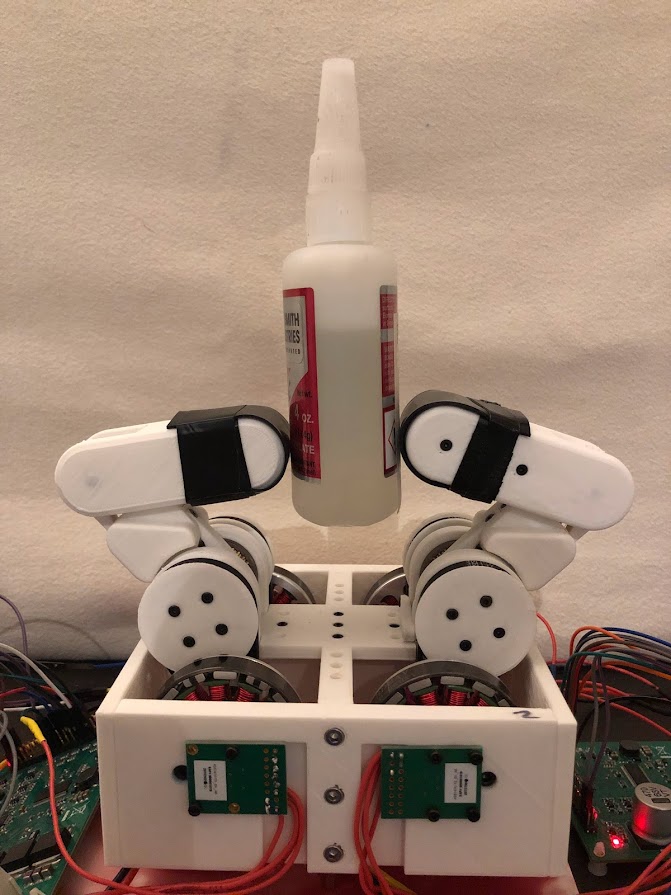}}\\
\subfloat[]{\includegraphics[scale=0.2]{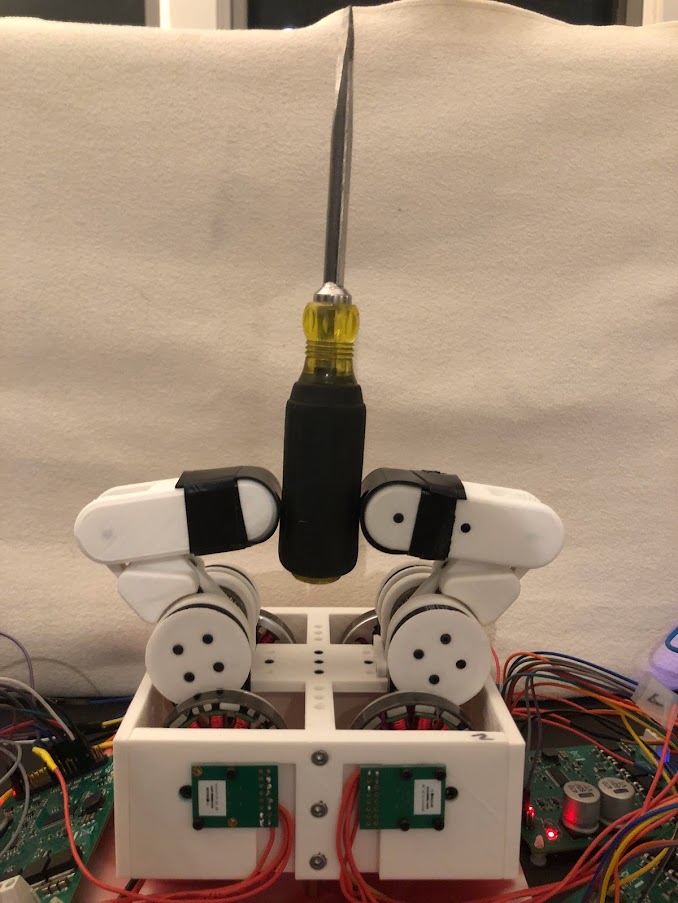}}
\hspace{0.5cm}
\subfloat[]{\includegraphics[scale=0.2]{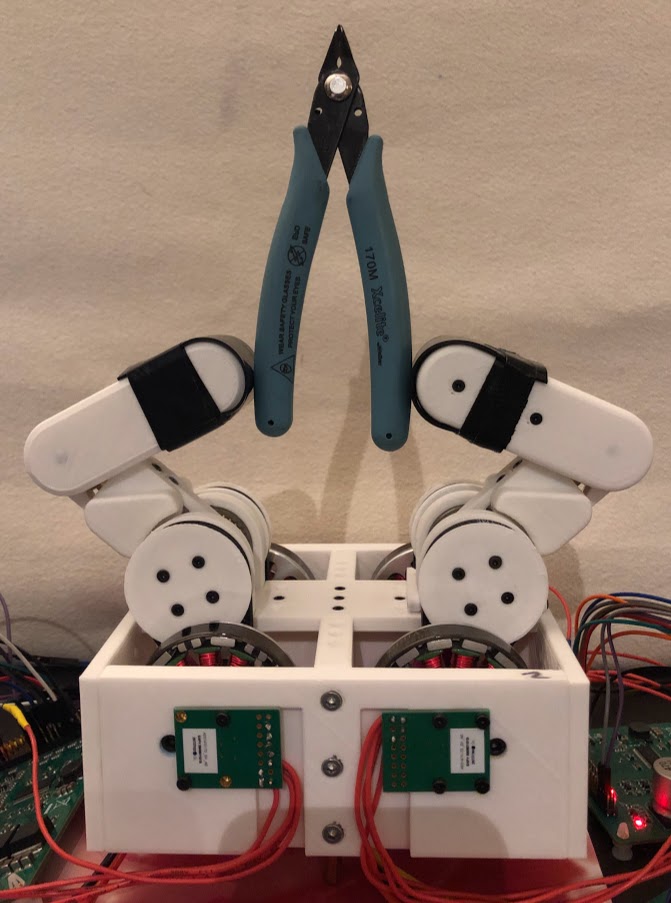}}
\hspace{0.5cm}
\subfloat[]{\includegraphics[scale=0.2]{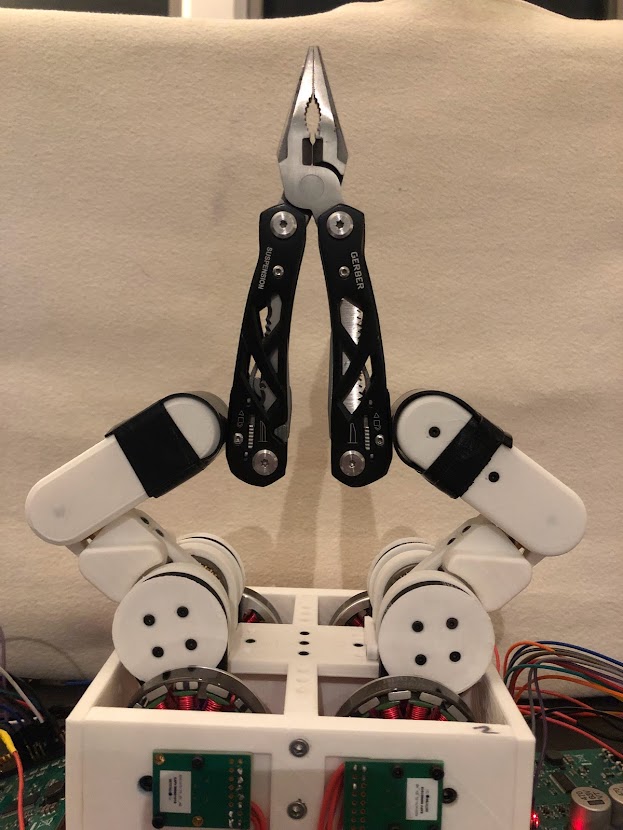}}\\
\subfloat[]{\includegraphics[scale=0.16]{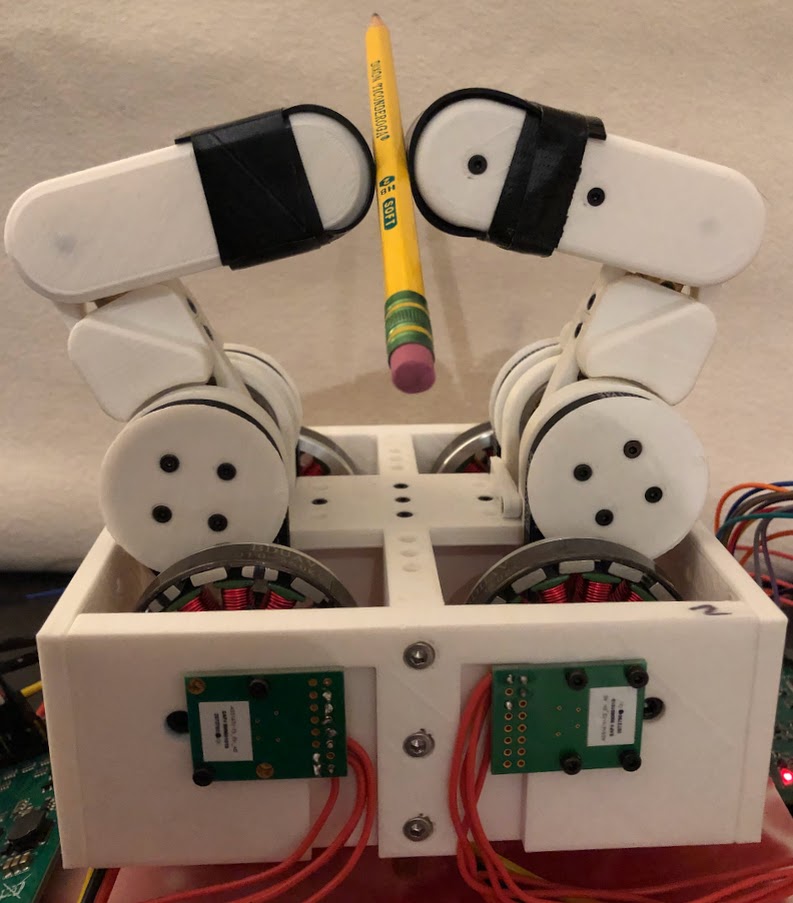}}
\hspace{0.5cm}
\subfloat[]{\includegraphics[scale=0.16]{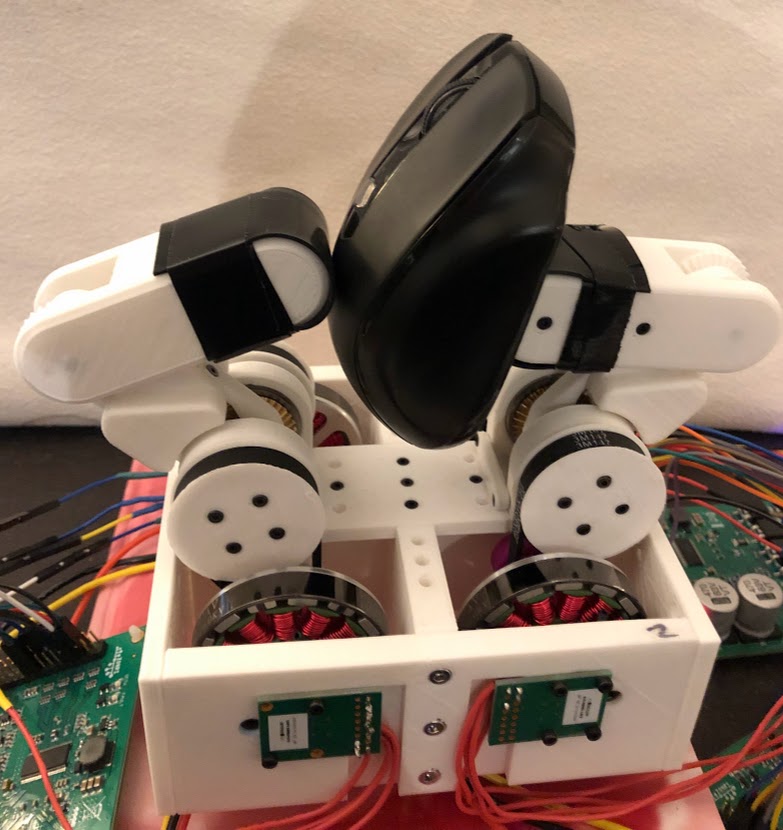}}
\hspace{0.5cm}
\subfloat[]{\includegraphics[scale=0.16]{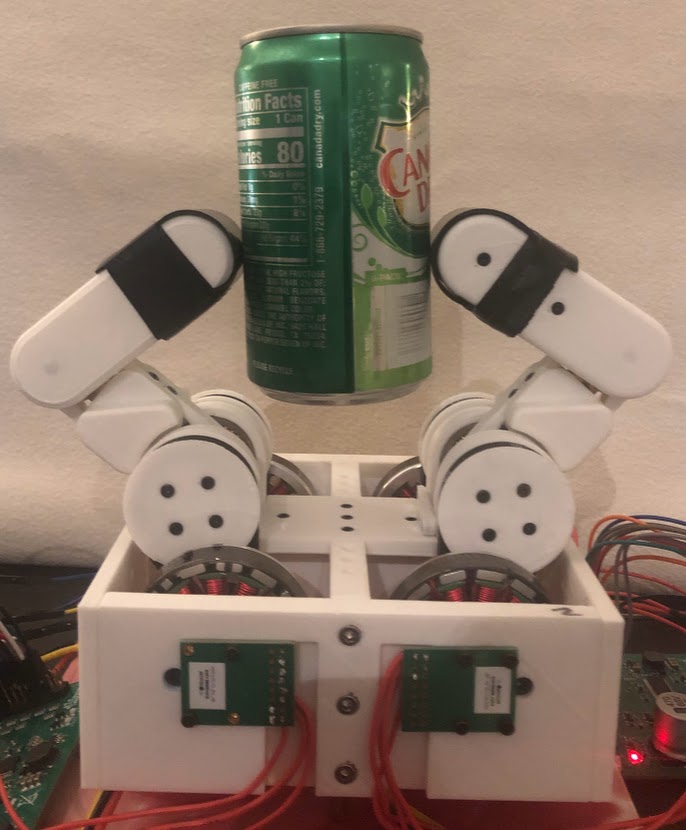}}
\caption{Demonstration of the force-closure grasp of various household objects.}
\label{fig:objects}
\end{figure}

This section will examine the effectiveness of the QDD hand in terms of its effectiveness in grasping a variety of objects with its fingertips. The objects are held in force closure, meaning that friction from contact forces between the object and the fingers is the only thing holding the objects in place. For these grasps, the Cartesian impedance control model is used as we are primarily concerned with applying force on the object in the task/planar Cartesian space. Fig.~\ref{fig:objects} shows the hand grasping objects in its fingertips of different shapes and weights. For each of the objects grasped, the desired position for each finger is located inside the object. The difference between the actual and desired positions of the fingers creates a spring-like force on the object holding it in place. The force applied in each direction is proportional to the difference between actual and desired positions by the spring constant used for each test in the $X$ and $Y$ directions.

\begin{figure}[!htbp]
\centering
\subfloat[]{\includegraphics[height = 4cm]{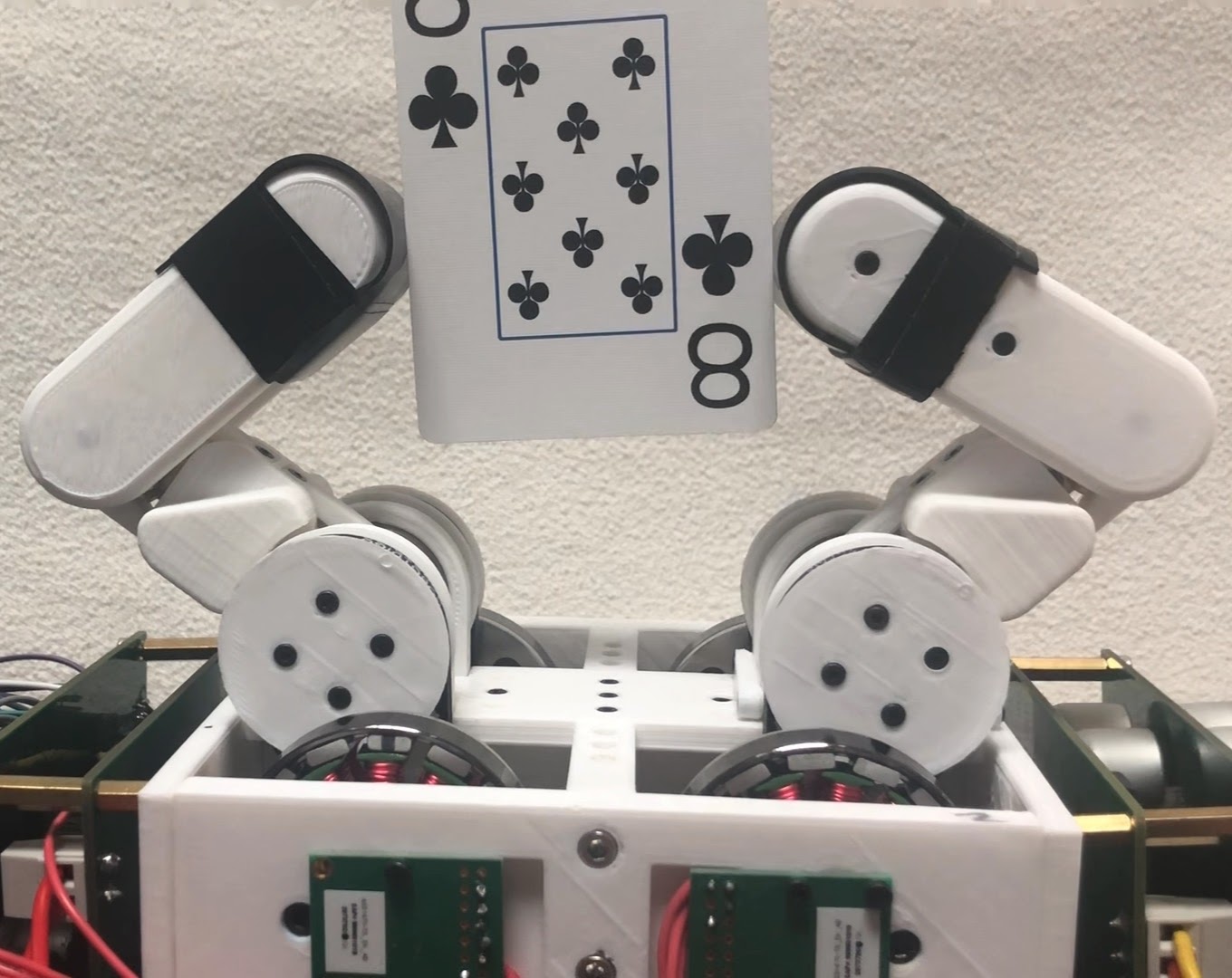}}
\hspace{0.5cm}
\subfloat[]{\includegraphics[height = 4cm]{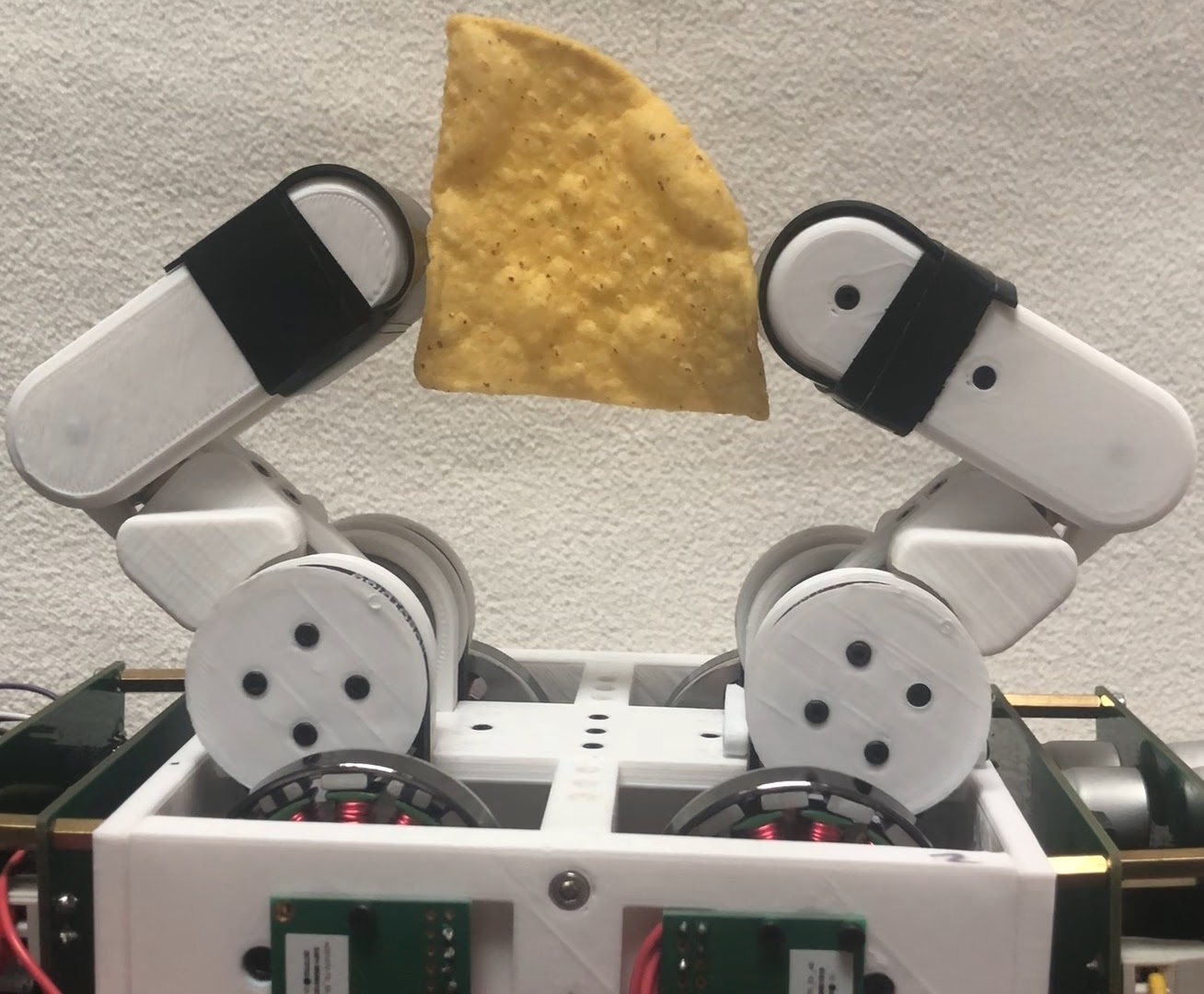}}\\
\subfloat[]{\includegraphics[height = 5cm]{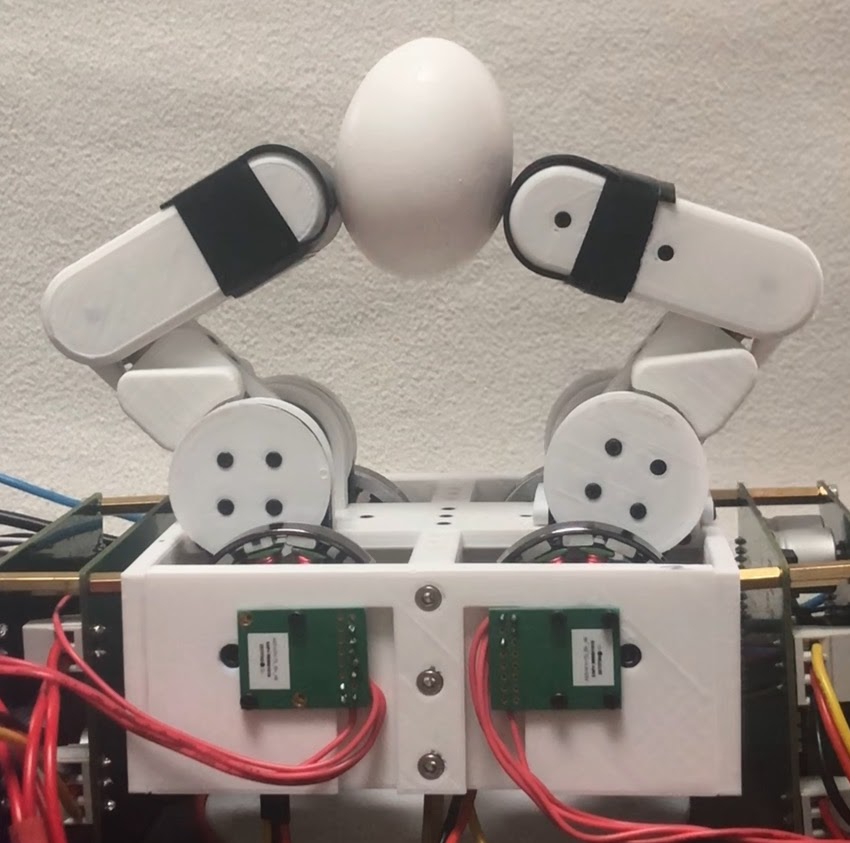}}
\hspace{0.5cm}
\subfloat[]{\includegraphics[height = 5cm]{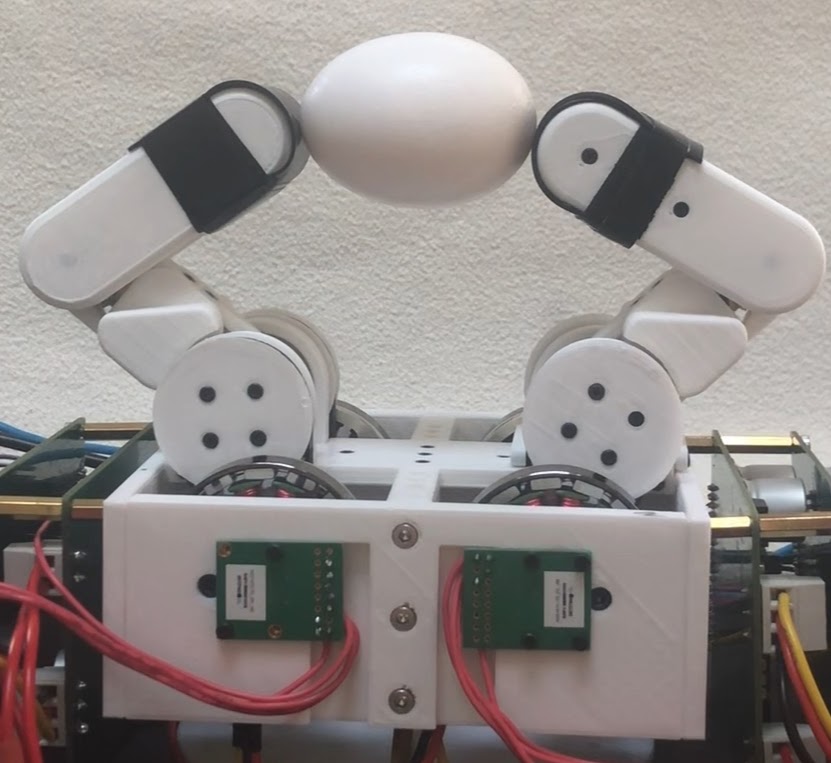}}
\caption{Demonstration of the force-closure grasp of fragile/delicate objects.}
\label{fig:delicate objects}
\end{figure}

In force-closure grasp, we can also demonstrate the effectiveness of our hand in grasping delicate and fragile objects of unknown size (Fig.~\ref{fig:delicate objects}). We do this by limiting the maximum closing force of the fingers as they move in the $X$-direction and close around the object. This can be seen in the video of the hand grasping a playing card without bending (\url{https://youtu.be/jcZhpDPOkac}), grasping a tortilla chip without breaking (\url{https://youtu.be/nhsiFj69y4k}), and grasping an egg (\url{https://youtu.be/4gP3ush-n-A}).

\subsection{Form-Closure Grasp}
\begin{figure}[!htbp]
\centering
\subfloat[]{\includegraphics[scale=0.06]{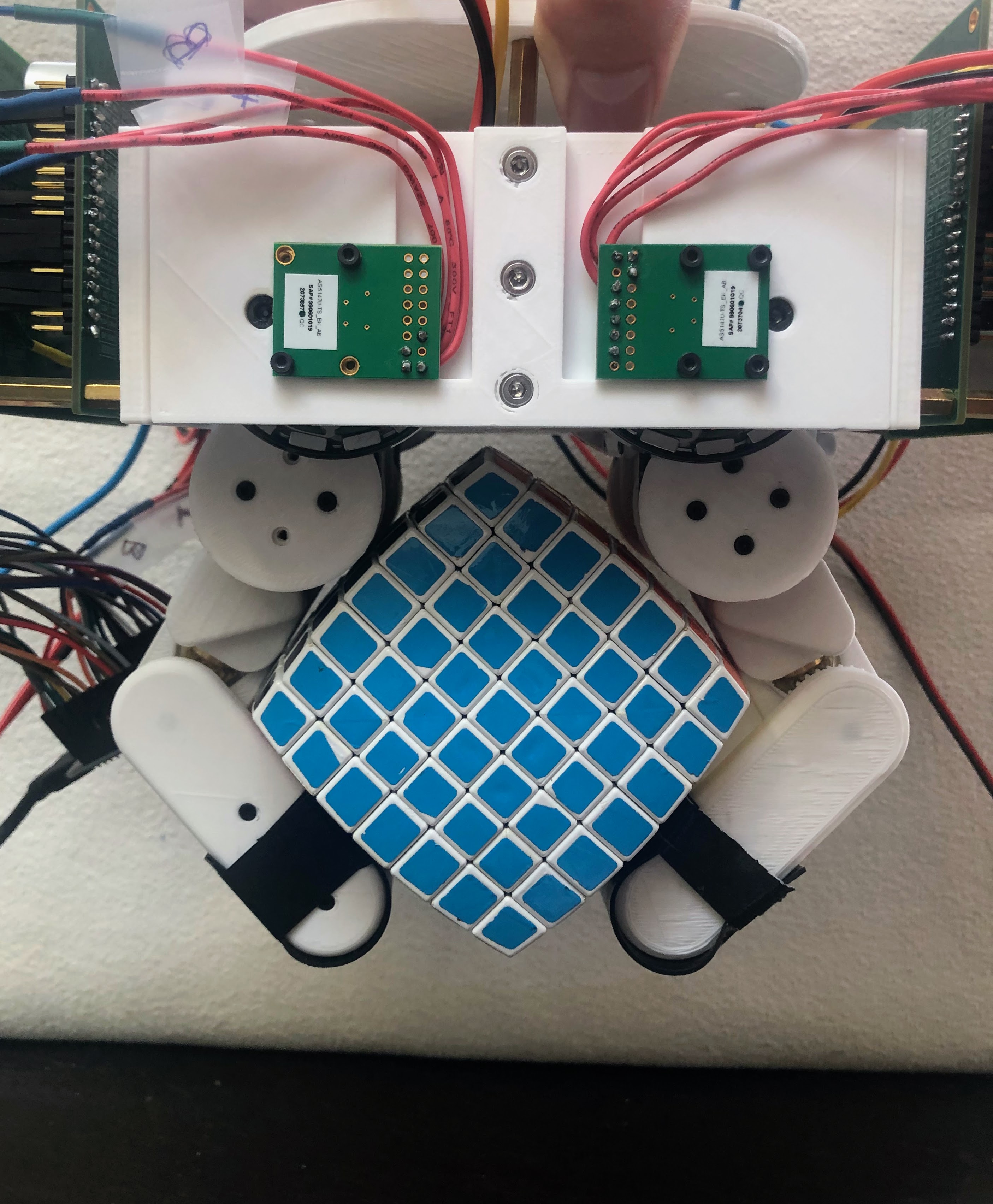}}
\hspace{0.5cm}
\subfloat[]{\includegraphics[scale=0.08]{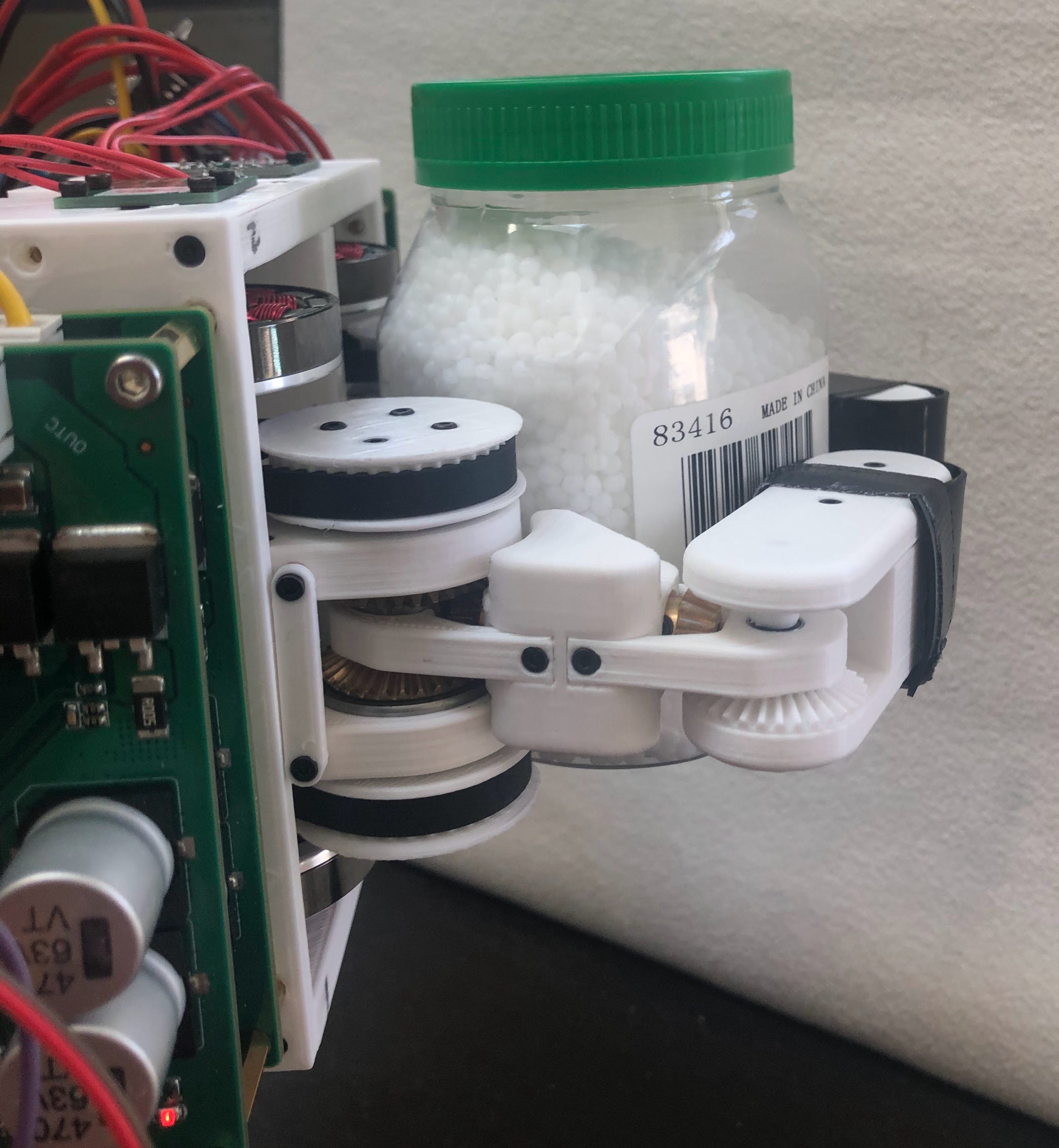}}\\
\subfloat[]{\includegraphics[scale=0.08]{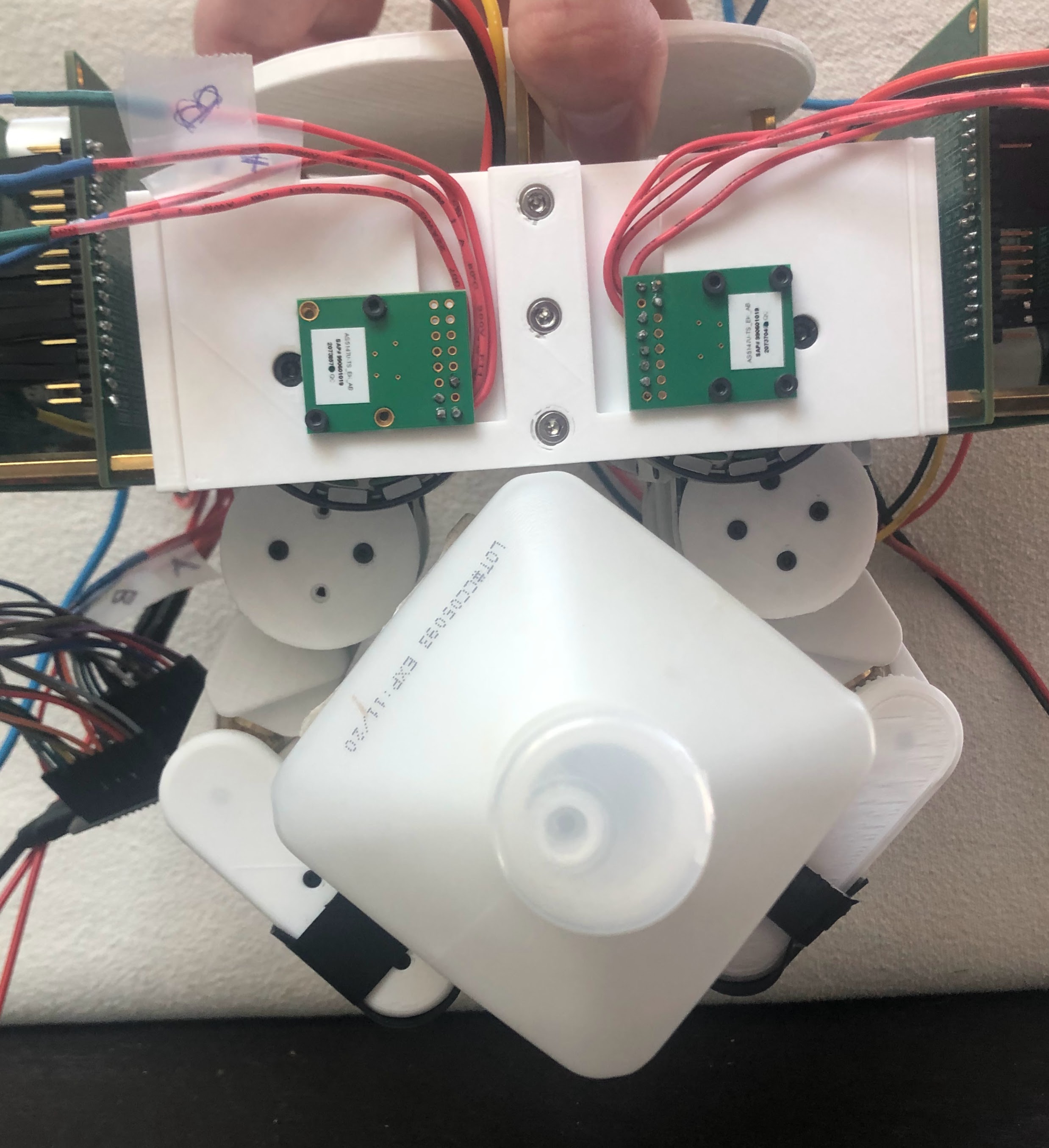}}
\hspace{0.5cm}
\subfloat[]{\includegraphics[scale=0.06]{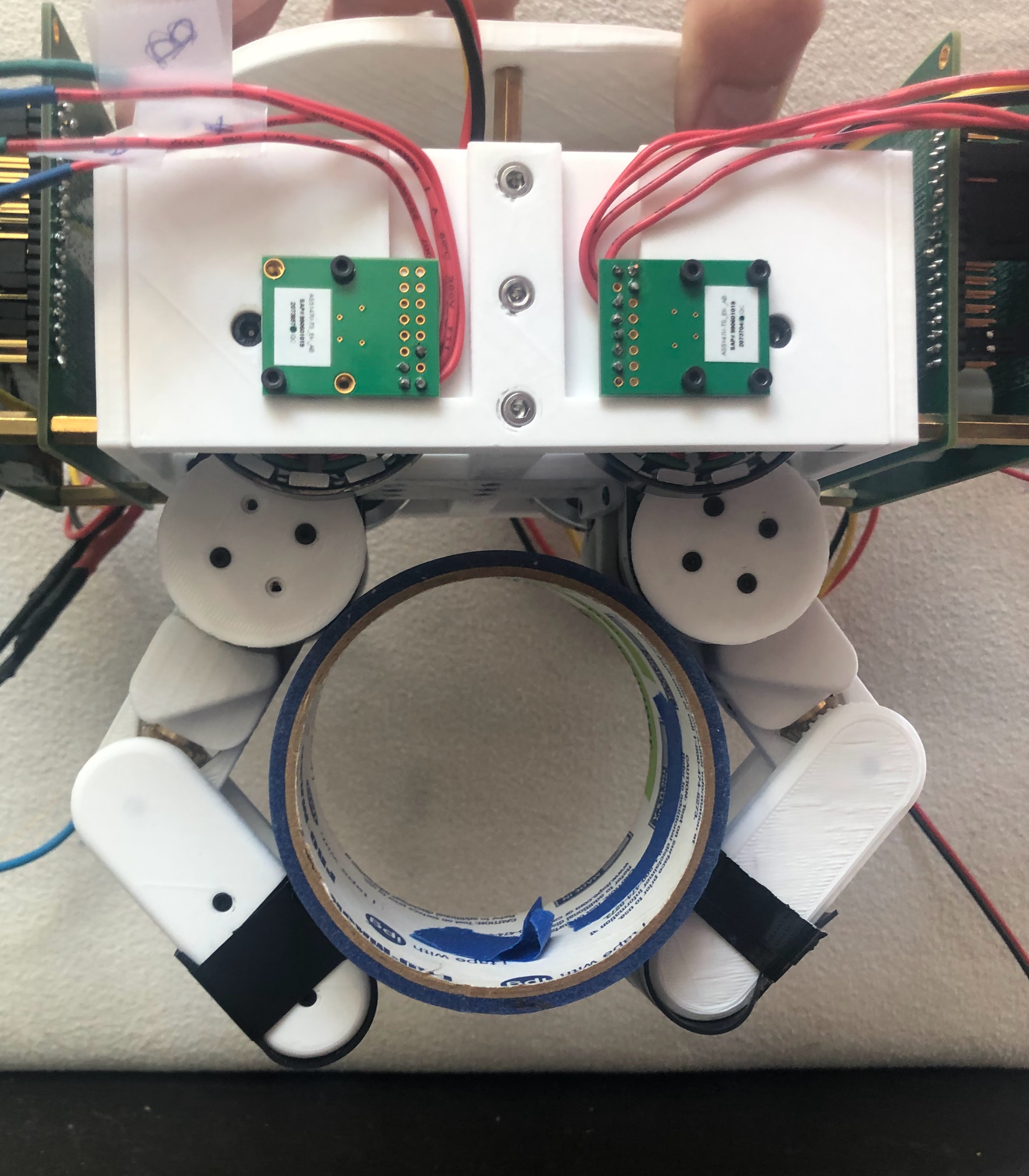}}
\caption{Demonstration of the form-closure grasp of various household objects.}
\label{fig:form}
\end{figure}

Our hand is unique compared to other prior DD and QDD hands in its ability to perform both force-closure and form-closure grasps. In contrast to force closure, form closure achieves a stable grasp through the shape of the gripper matching the shape of the object being grasped. 
We control the finger in the joint space rather than Cartesian space for form closure because the objects contact the finger along the faces of the two finger links and not just at the tips of the fingers. Using this technique, we can form the fingers around a variety of objects of different unknown sizes and shapes as shown in Fig.~\ref{fig:form}. A demonstration of form closure can be found here: \url{https://youtu.be/Sd9FdDblNxo}.

\section{Stable Grasps in Response to Disturbance}
\begin{figure}[!htbp]
\centering
\subfloat[]{\includegraphics[height = 4.6cm]{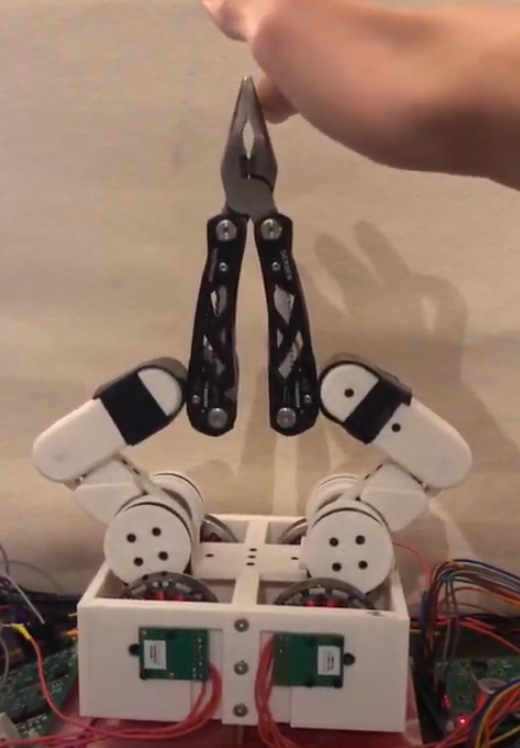}}
\subfloat[]{\includegraphics[height = 4.6cm]{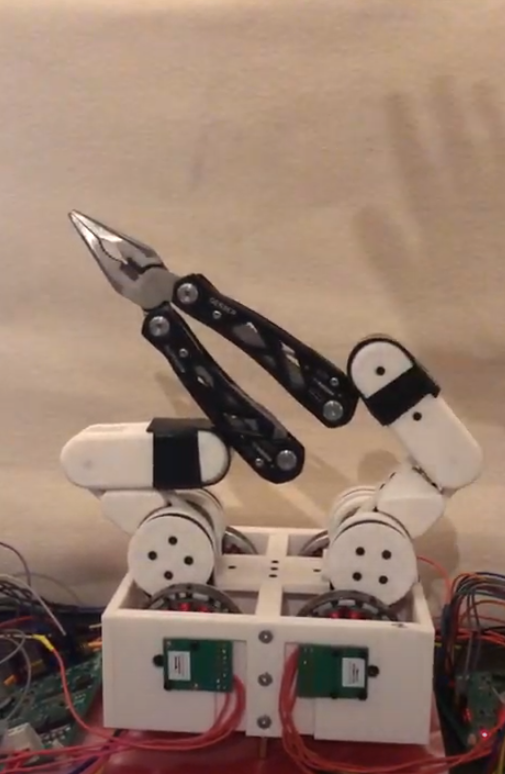}}
\subfloat[]{\includegraphics[height = 4.6cm]{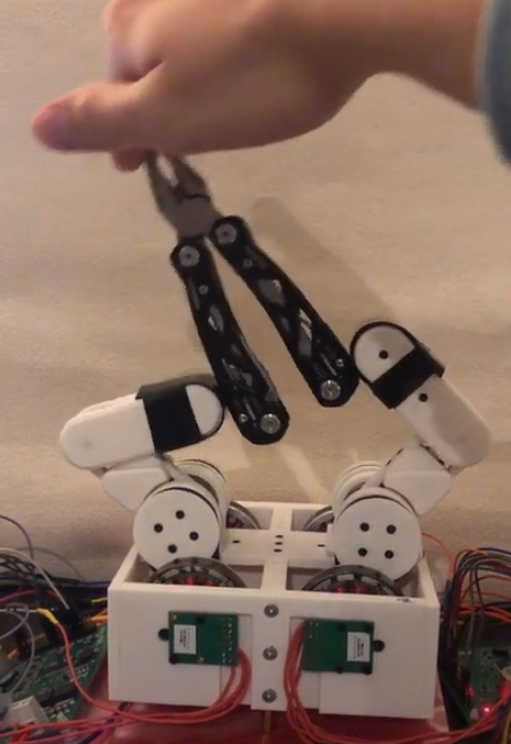}}
\subfloat[]{\includegraphics[height = 4.6cm]{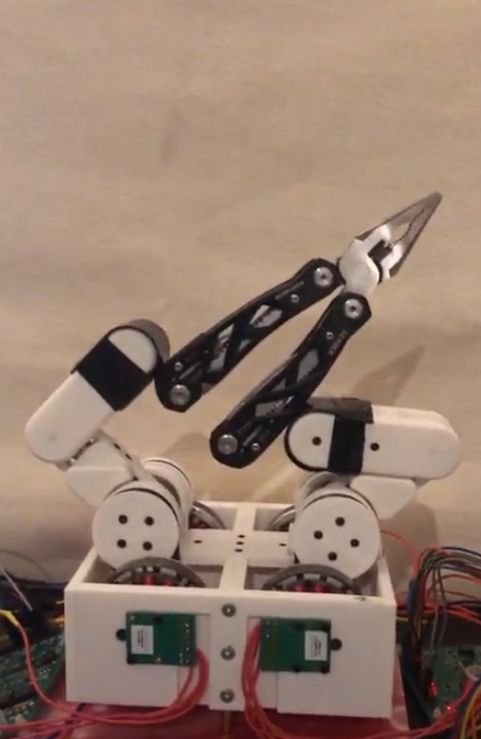}}
\subfloat[]{\includegraphics[height = 4.6cm]{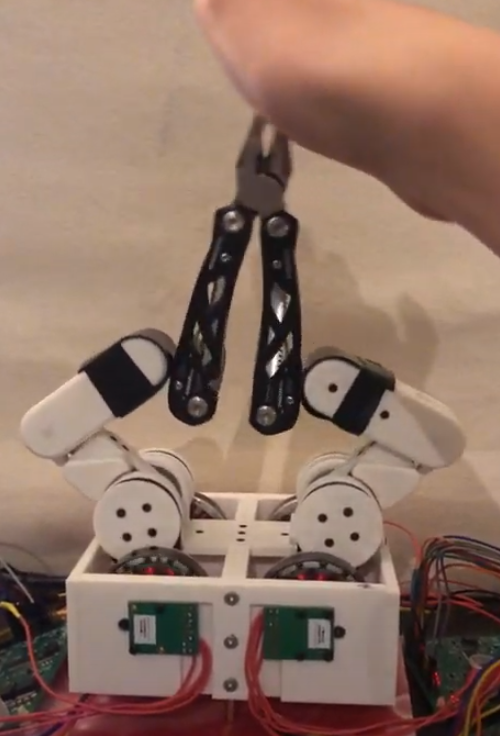}}
\caption{Fingertip grasping of pliers in response to an external wrench.}
\label{fig:pliers}
\end{figure}

When robotic manipulators are deployed in the real world, they are likely to undergo unplanned contact with the environment. As a result, robotic hands and grippers must be able to maintain a stable grasp in response to these disturbances. Impedance control helps make this possible. When performing force-closure grasp (Fig.~\ref{fig:objects}), the fingers attempt to maintain a desired position, but they are also compliant, allowing movement in response to external forces in accordance with the defined stiffness and damping matrices. As shown in Fig.~\ref{fig:pliers}, when an external wrench is applied to the pliers, the fingers retain rolling contact friction between the fingertips and the handles of the pliers. This allows the pliers to resist slipping from the grasp of the hand. After the external force is applied, the weight of the pliers prevents the fingers from returning to their original position, but a stable grasp is successfully maintained despite the hand having only two fingers. A video of this test being performed can be found here: \url{https://youtu.be/bYJElUFPpNE}.

\section{Smack-and-Snatch Manipulation}
\begin{figure}[!htbp]
\centering
\includegraphics[width = 16.5 cm]{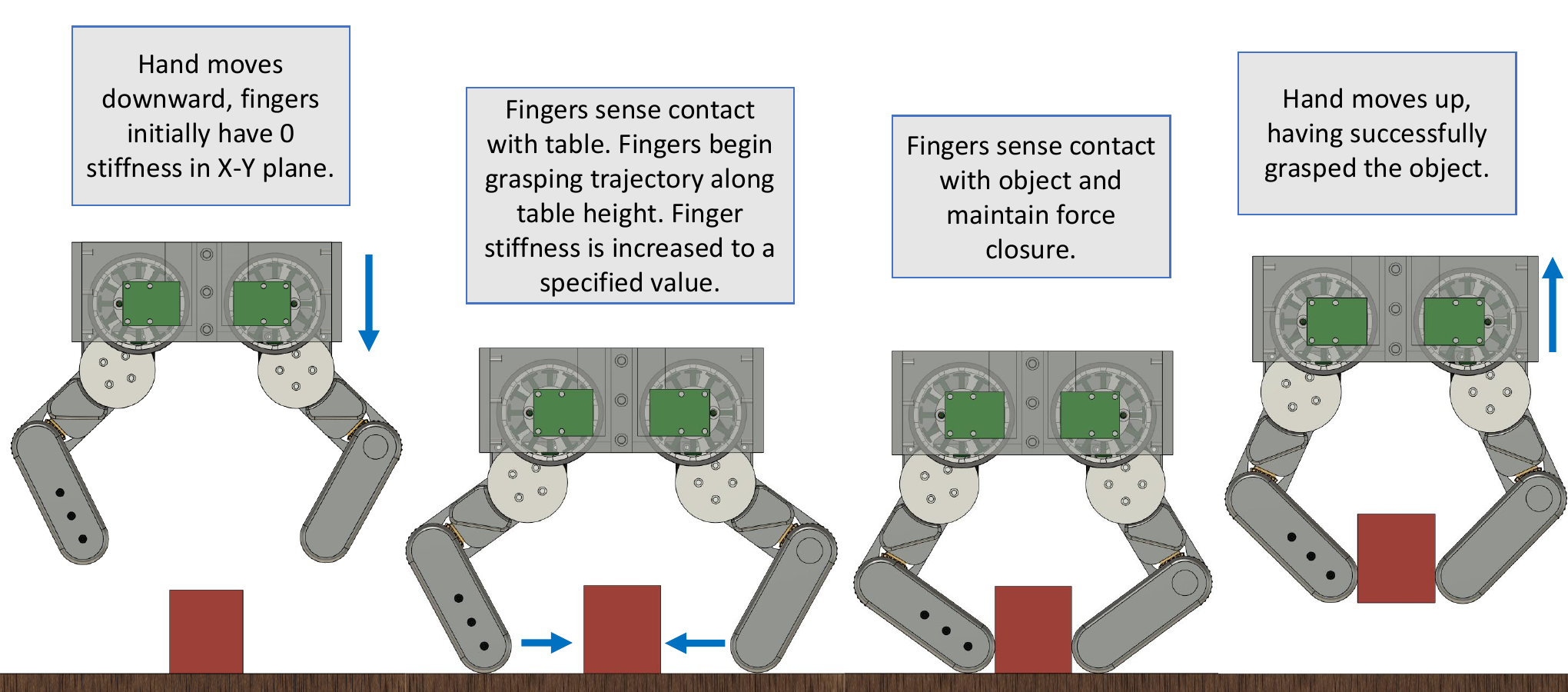}
\caption{Smack-and-snatch manipulation with our QDD robot hand.}
\label{fig:smack and snatch}
\end{figure}

\textit{Smack-and-snatch} refers to a dynamic manipulation task first coined and performed by the CMU DDHand \cite{Bhatia2019DirectDH}, and later recreated by Lin \textit{et al.} \cite{QuasiDirectDrive_Lin} with their linkage based QDD robotic hand. Essentially, the robotic hand or gripper must grasp an object from overhead at an unknown height. Since these hands have transparent, backdriveable drive-trains, the hand can safely sense contact with the table/environment using finger position sensors, and then, close the fingers of the hand after sensing contact with the table. Because of the low inertia and friction of the fingers, the impulse imparted on the hand by the environment is low, and the hand does not get damaged even at high velocities. This compliance in the hand allows the fingers to act as a contact sensor with the environment, and trigger the grasping trajectory of the hand. This method demonstrates how the fingers of our QDD hand can be used as a sensor to sense contact with the environment which can be exploited during a task. In addition, this method allows for potentially highly dynamic grasping at high speeds because the manipulator does not need to pause at the bottom of the trajectory for a successful grasp. The sequence of smack-and-snatch manipulation is visualized with our QDD hand in Fig.~\ref{fig:smack and snatch}.

\begin{figure}[!htbp]
\centering
\subfloat[]{\includegraphics[height = 5.2cm]{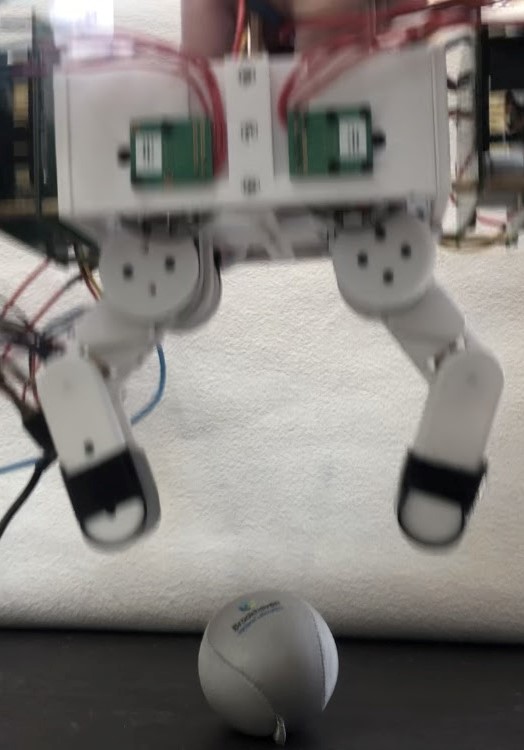}}
\subfloat[]{\includegraphics[height = 5.2cm]{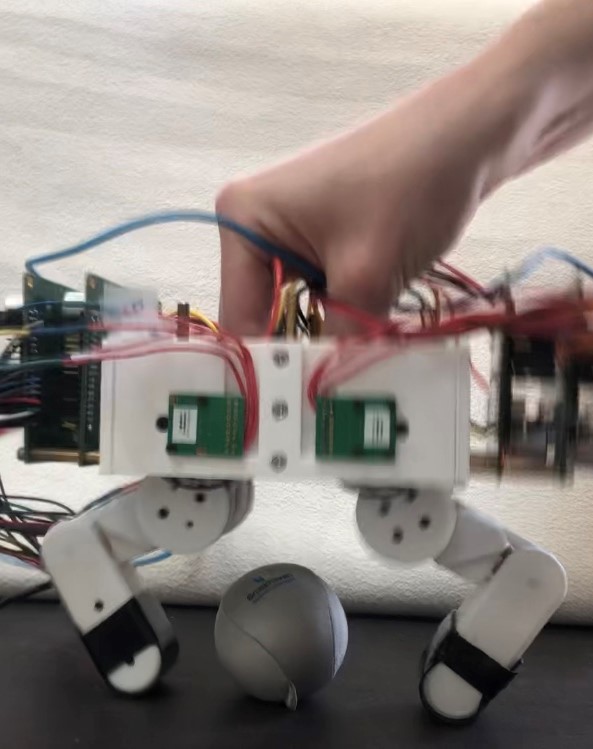}}
\subfloat[]{\includegraphics[height = 5.2cm]{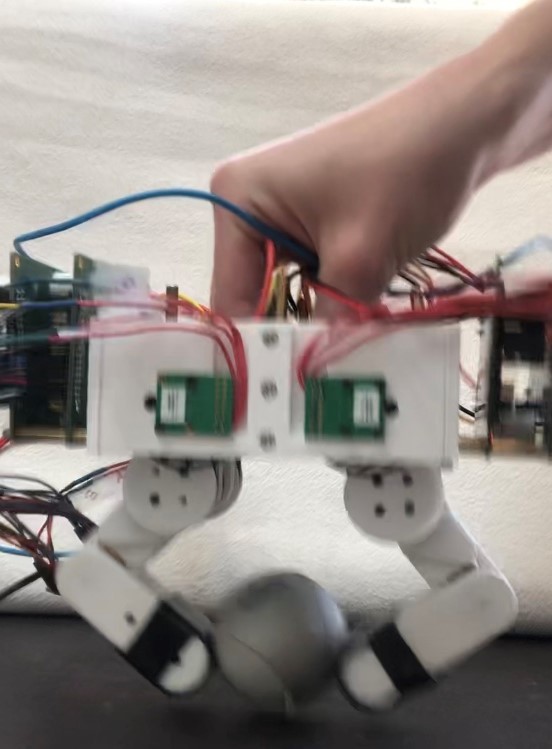}}
\subfloat[]{\includegraphics[height = 5.2cm]{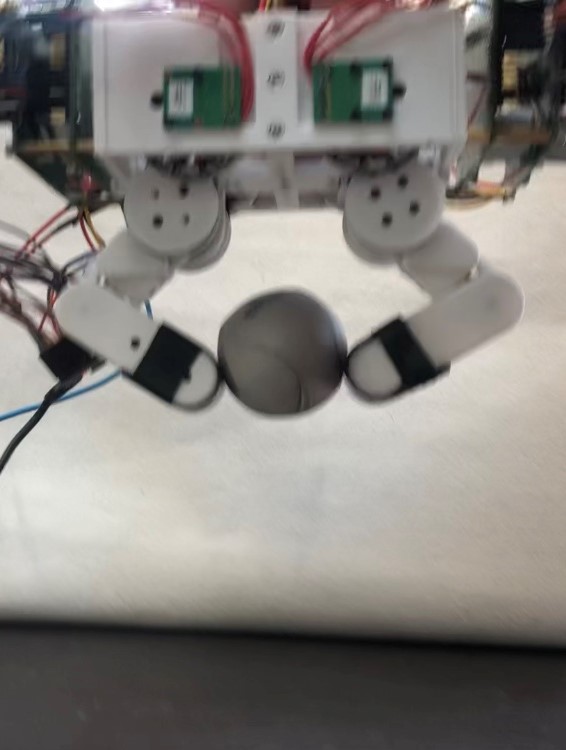}}
\caption{Smack-and-snatch manipulation of a ball.}
\label{fig:sas}
\end{figure}

We can test this concept by simply moving the robot hand manually up and down in a fast, vertical trajectory as shown in Fig.~\ref{fig:sas}. Initially, the fingers each have a stiffness matrix of zero. When the fingers are displaced by a specified value in the vertical direction, the stiffness matrix changes to some constant in the $X$ and $Y$ directions. At this point, the fingers each travel to their desired positions located in the center of the workspace. This enables the hand to grasp the ball in a quick, smooth motion without the need to pause at the bottom of the hand trajectory, and without the need for extra sensors to locate the table position. A video of this test can be found here: \url{https://youtu.be/0FjuOXXyo-8}.

\begin{figure}[!htbp]
\centering
\subfloat[]{\includegraphics[height = 5.2cm]{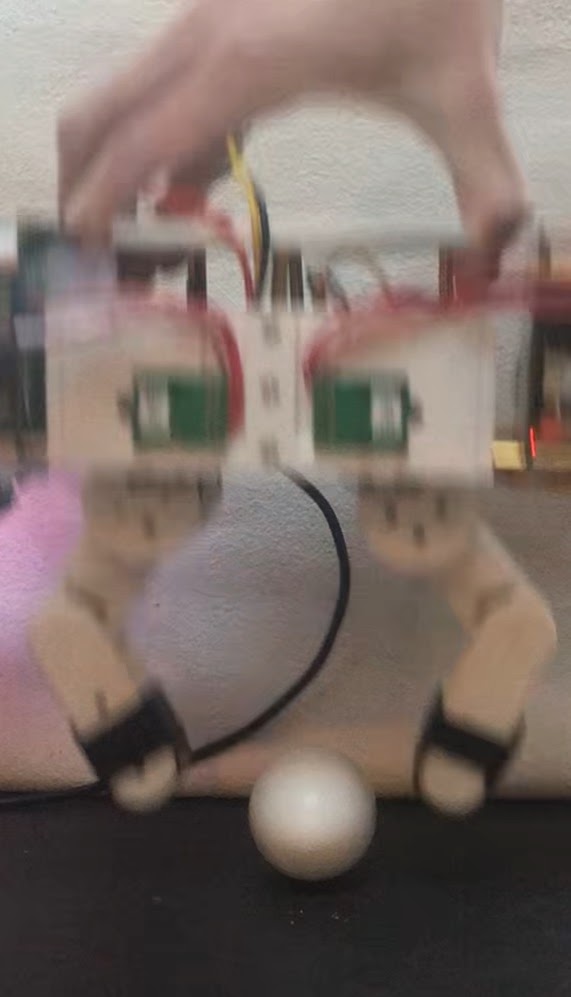}}
\subfloat[]{\includegraphics[height = 5.2cm]{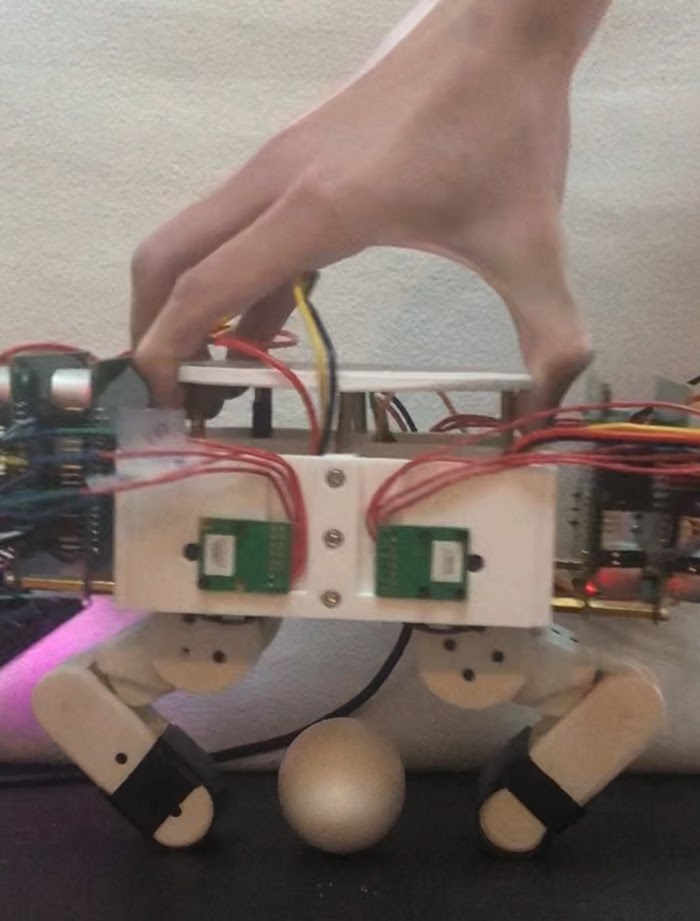}}
\subfloat[]{\includegraphics[height = 5.2cm]{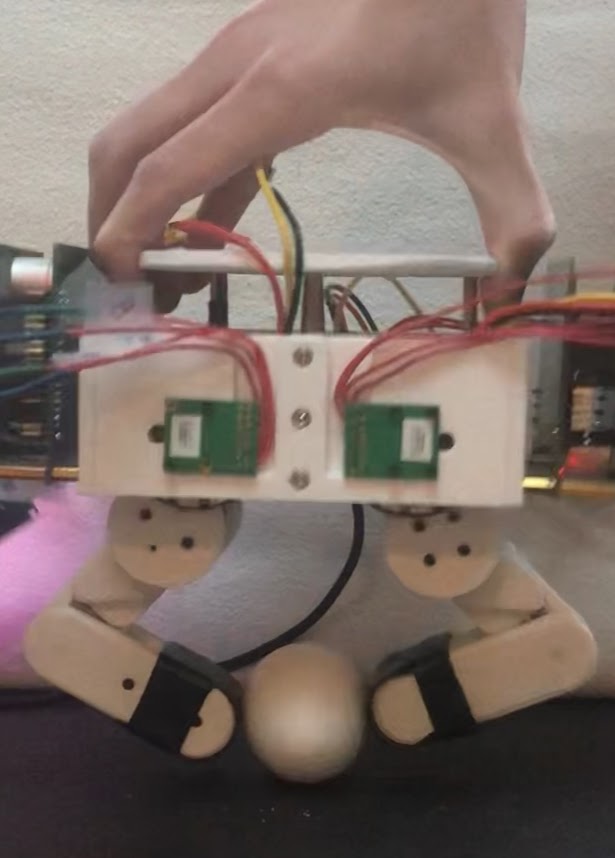}}
\subfloat[]{\includegraphics[height = 5.2cm]{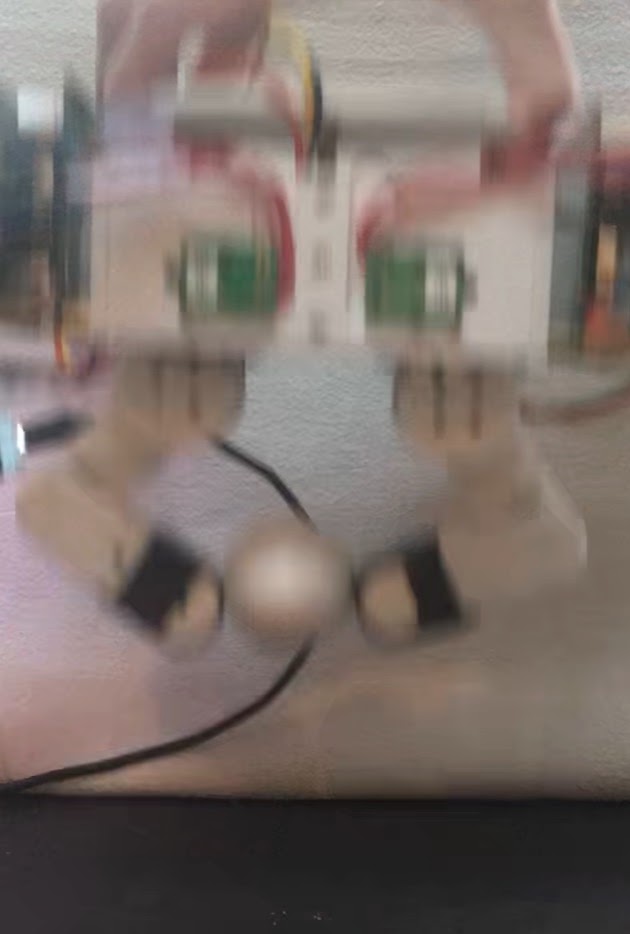}}
\caption{Smack-and-snatch manipulation of an egg.}
\label{fig:sas egg}
\end{figure}

This task can be applied to grasping objects which are fragile and of unknown size. By limiting the maximum force of the fingers in the $X$-direction we can perform the \textit{smack-and-snatch} maneuver with an egg, albeit at a lower speed to prevent the egg from cracking (Fig. \ref{fig:sas egg}). This is the same technique we used to grasp the egg in section \ref{sec:force closure}. A video of this can be found here: \url{https://youtu.be/0b7juuTGyyA}.

\section{In-Hand Manipulation}
\subsection{Rotation and Translation}
\begin{figure}[!htbp]
\centering
\subfloat[]{\includegraphics[height = 2.9cm]{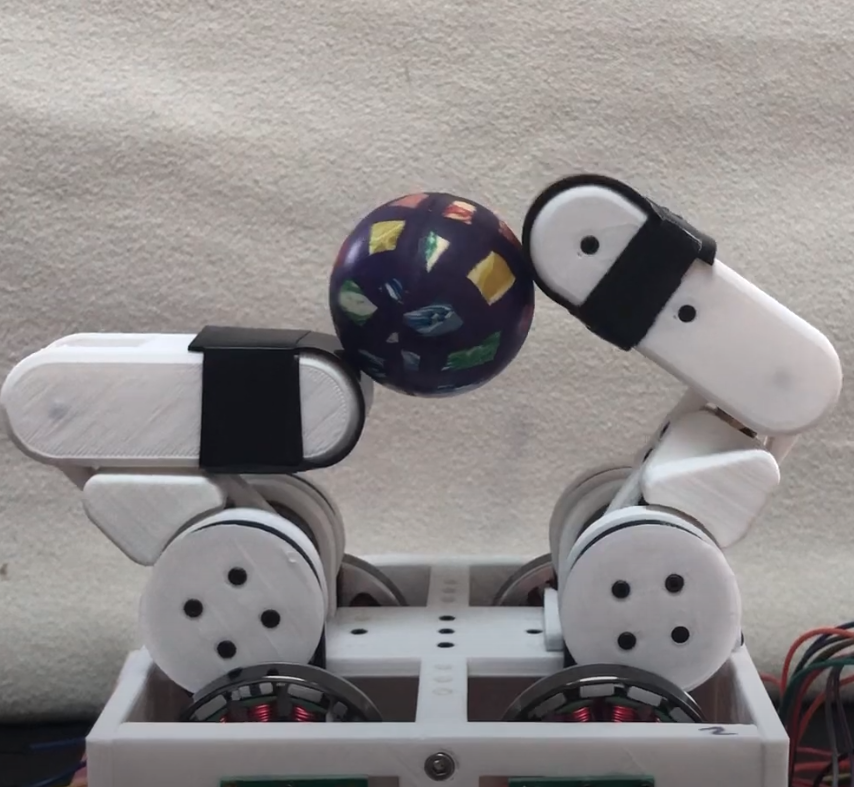}}
\subfloat[]{\includegraphics[height = 2.9cm]{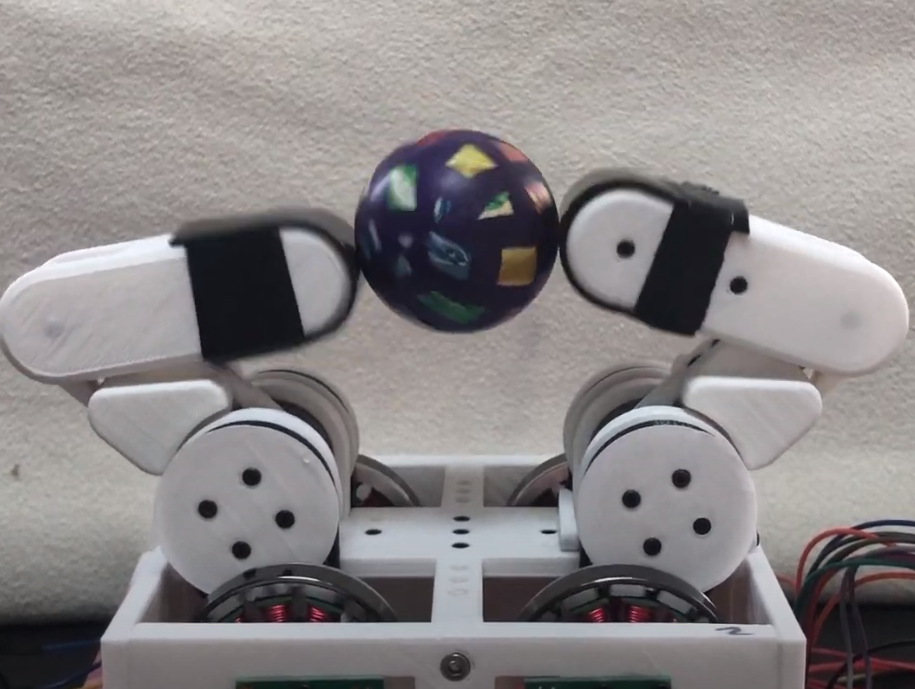}}
\subfloat[]{\includegraphics[height = 2.9cm]{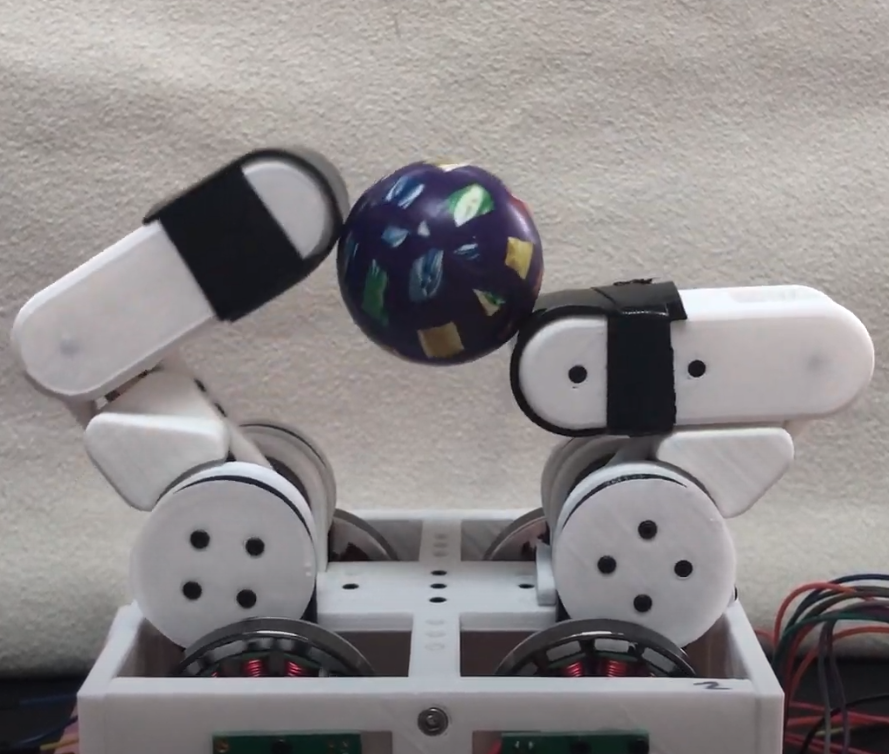}}
\subfloat[]{\includegraphics[height = 2.9cm]{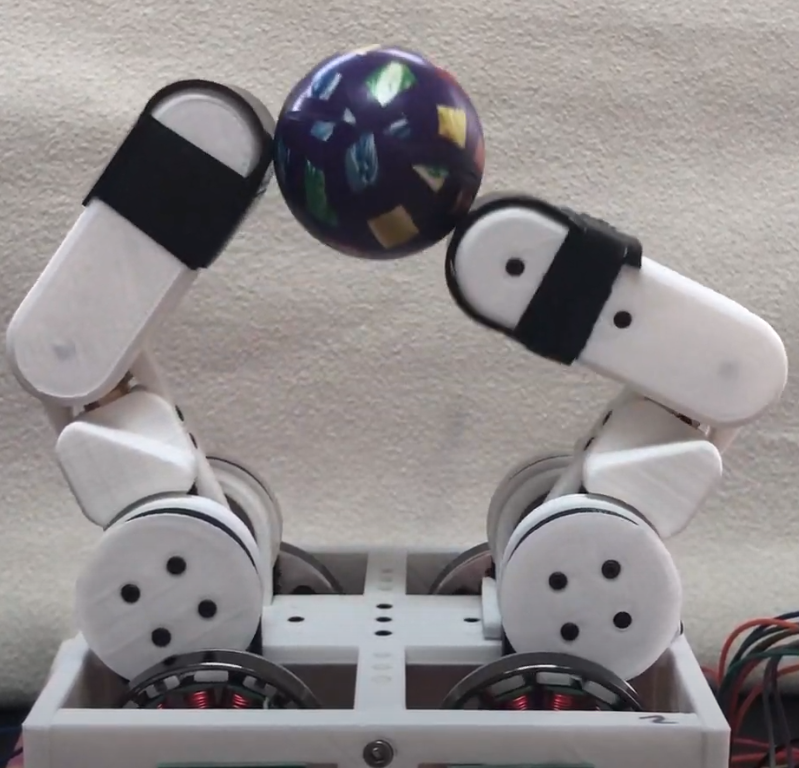}}
\subfloat[]{\includegraphics[height = 2.9cm]{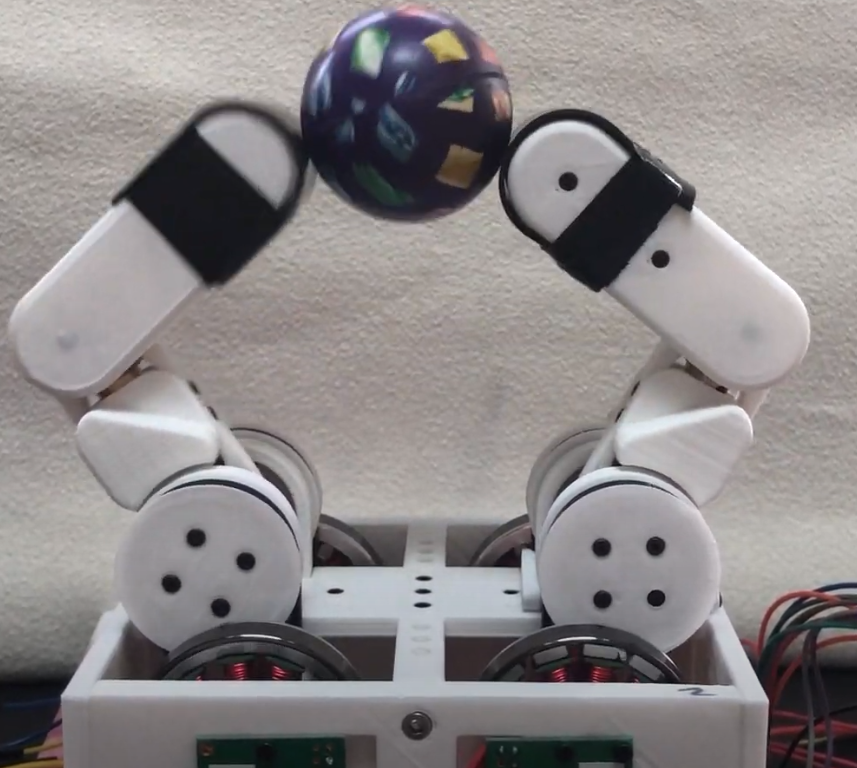}}
\caption{In-hand manipulation of a rubber ball.}
\label{fig:ball}
\end{figure}

In-hand manipulation is defined by the motion of the object in the robotic hand after it has been grasped. Due to the geometry of our hand, and the impedance controller used, the fingers can maintain contact force on the object while the fingers move through a specified trajectory. The fingers maintain rolling contact with the object as it is manipulated by the fingers. Rolling contact introduces nonholonomic constraints on the system which can be exploited to manipulate the object \cite{cui_sun_dai_2017}. This allows the hand to both rotate and translate the grasped object and greatly expands the capabilities of the hand compared to systems that can only pick and place objects. This task is demonstrated in Fig.~\ref{fig:ball}. The fingers of the hand rotate the ball while moving it up and down. The fingers have a linear desired trajectory in the $Y$-direction while trying to maintain a constant position in the $X$-direction less than that of the diameter of the rubber ball. The difference between the desired and actual finger positions in the $X$-direction is the source of the gripping force on the ball. Since Cartesian impedance control is being used here, the force applied is proportional to the stiffness coefficient in the $X$-direction. A video of this demonstration can be found here: \url{https://youtu.be/NOzzvAztOfY}.

In this example, we performed in-hand manipulation with both rotation and translation of the ball. We can also achieve pure rotation or translation of the object being grasped. A video of the ball being translated in a linear trajectory can be found here: \url{https://youtu.be/ytpdwM4pvCM}.

\subsection{Turning a Cylinder}
Another task which may require in-hand manipulation is turning a cylindrical object such as a screwdriver or bottle cap. In general, there are two ways humans go about accomplishing this task. We can either firmly grasp the object and use our hand or wrist to turn the object about its axis, or we can use our fingertips and turn the object with only our hands. Sometimes we may use a combination of both techniques. From a robotics perspective, turning with the fingertips requires much more dexterity, and would not be possible with a simple parallel jaw gripper. This is what we will attempt to replicate with our robotic hand.

\begin{figure}[!htbp]
\centering
\includegraphics[width = 12 cm]{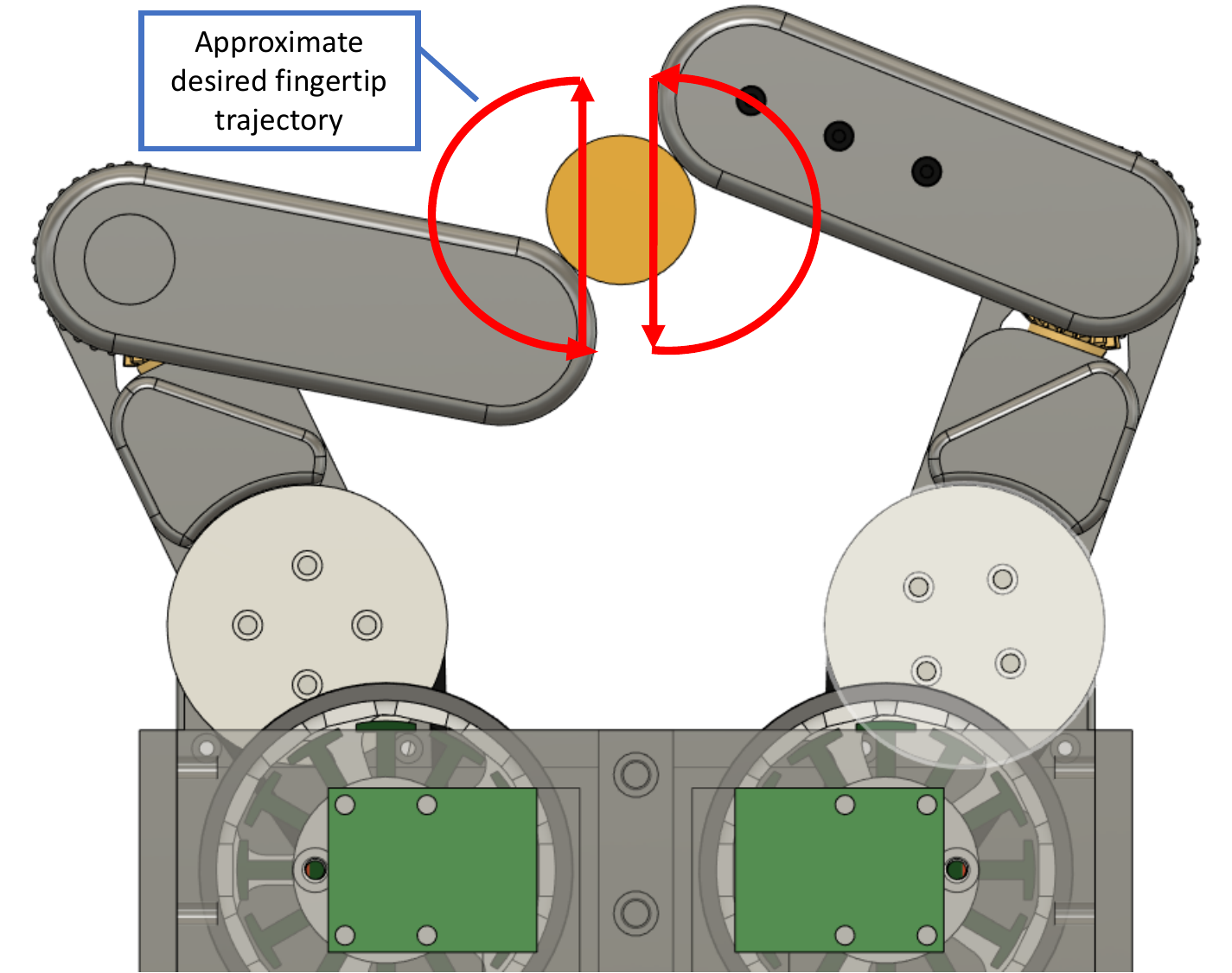}
\caption{Desired trajectory for task of turning a screwdriver (shown in yellow).}
\label{fig:screw}
\end{figure}

As shown in Fig.~\ref{fig:screw}, both the desired finger trajectories intersect with the cross-section of the screwdriver. In these regions, the actual path of the fingertip will be constrained with rolling frictional contact to the surface of the screwdriver. The difference between the actual and desired positions of the finger will cause force to be applied to the screwdriver proportional to the stiffness coefficient in the $X$ and $Y$ directions since we are using Cartesian impedance control. This greatly simplifies the motion planning for this particular task. Rather than trying to use force or position control to manipulate the object, we simply create a relation between position and force in the form of the impedance controller. By modifying the parameters of the impedance controller, we can create the desired outcome. In this case, this means applying torque to the cylinder such that it rotates.

\begin{figure}[!htbp]
\centering
\subfloat[]{\includegraphics[height = 3.1cm]{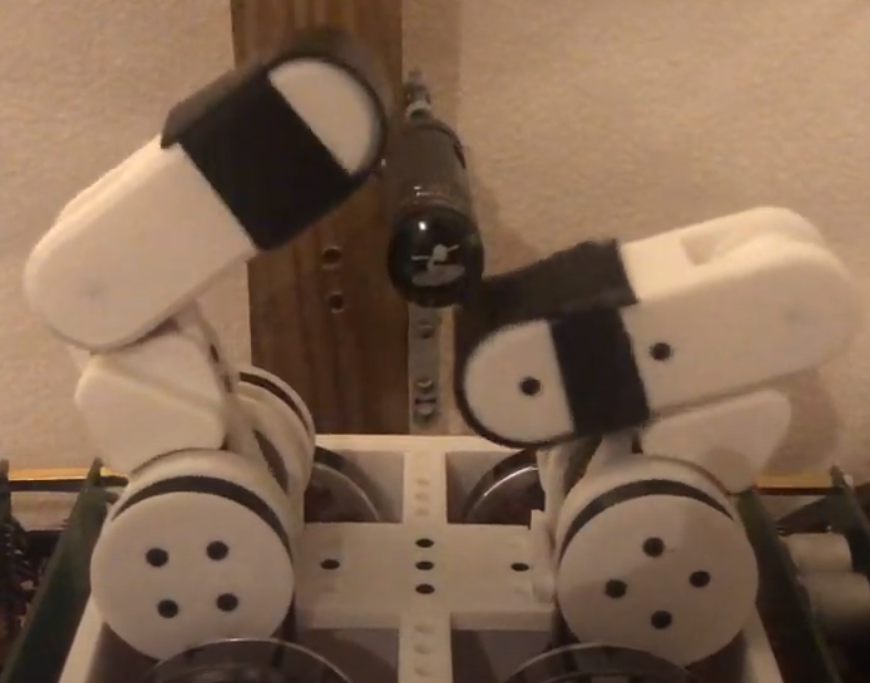}}
\subfloat[]{\includegraphics[height = 3.1cm]{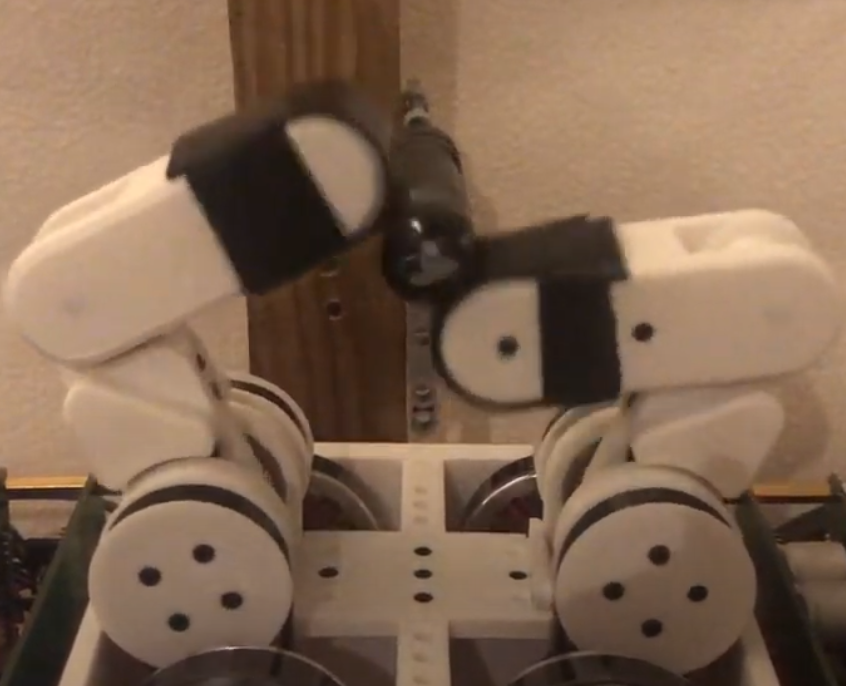}}
\subfloat[]{\includegraphics[height = 3.1cm]{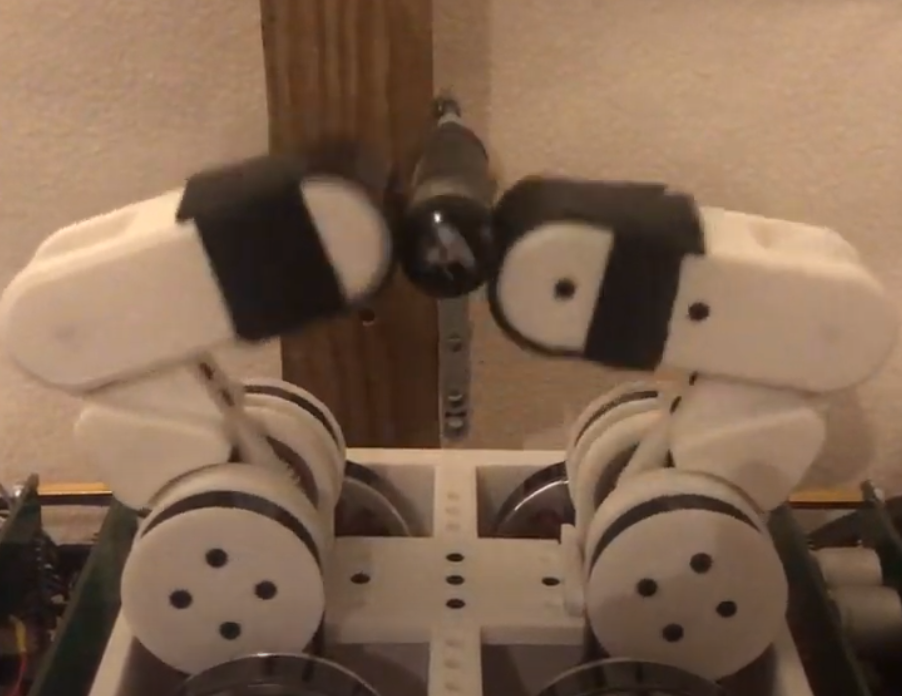}}
\subfloat[]{\includegraphics[height = 3.1cm]{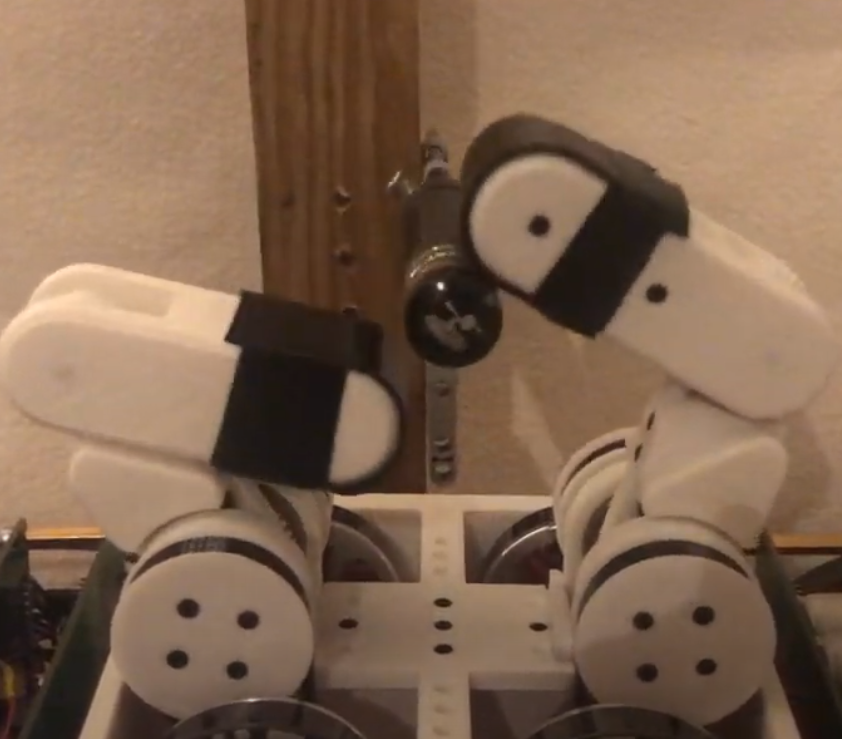}}
\caption{Turning a screwdriver.}
\label{fig:screw2}
\end{figure}

Figure~\ref{fig:screw2} shows our test setup for in-hand turning of the screwdriver. Being that the hand has only two fingers, the movement of the screwdriver is constrained to its axis by supports mounted behind the hand. In the future, it could be useful to add additional fingers to the hand so that it could accomplish this task unassisted by the environment. Nevertheless, this test shows that our QDD hand can use in-hand manipulation to perform some useful tasks. A video of this test being performed can be found here: \url{https://youtu.be/B7YEwJ3F6jc}.

\begin{figure}[!htbp]
\centering
\subfloat[]{\includegraphics[height = 3.4cm]{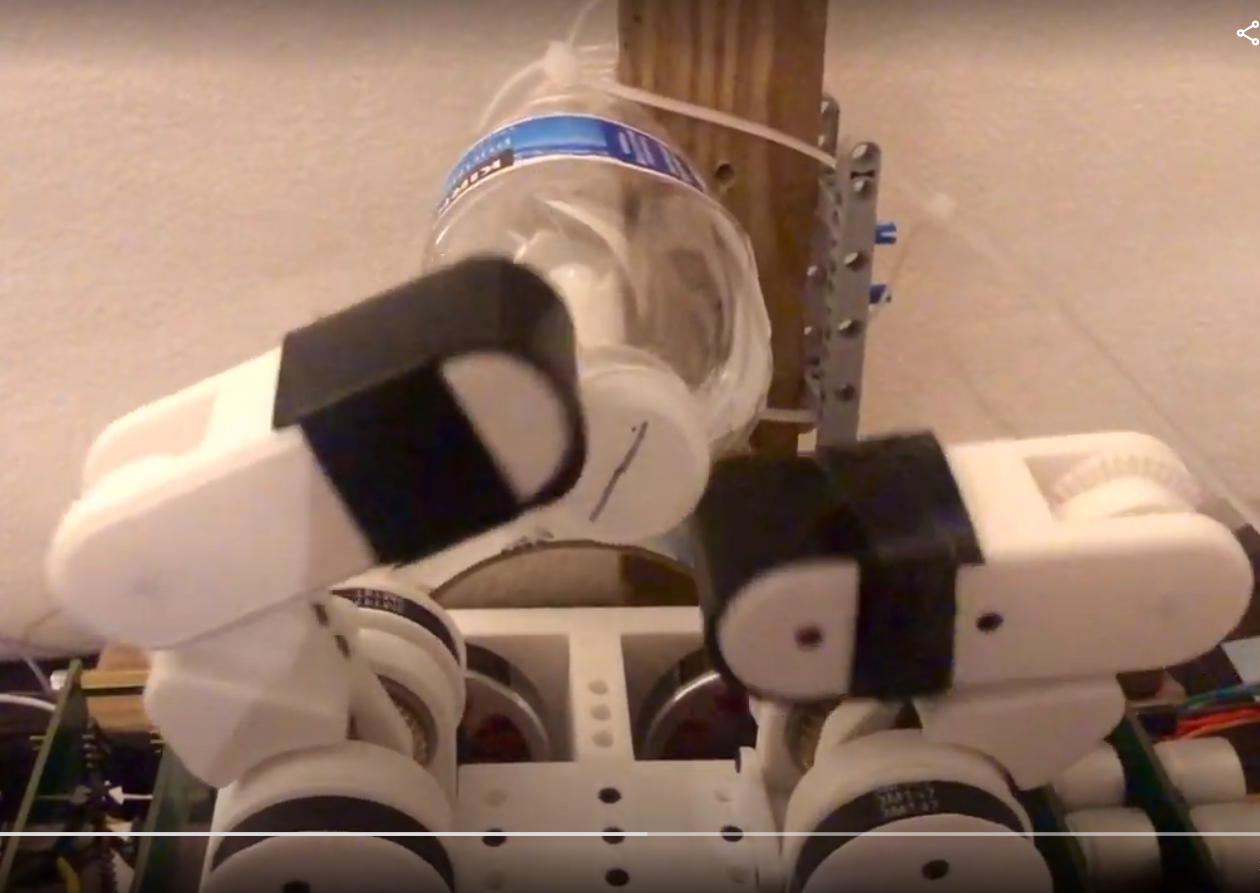}}
\subfloat[]{\includegraphics[height = 3.4cm]{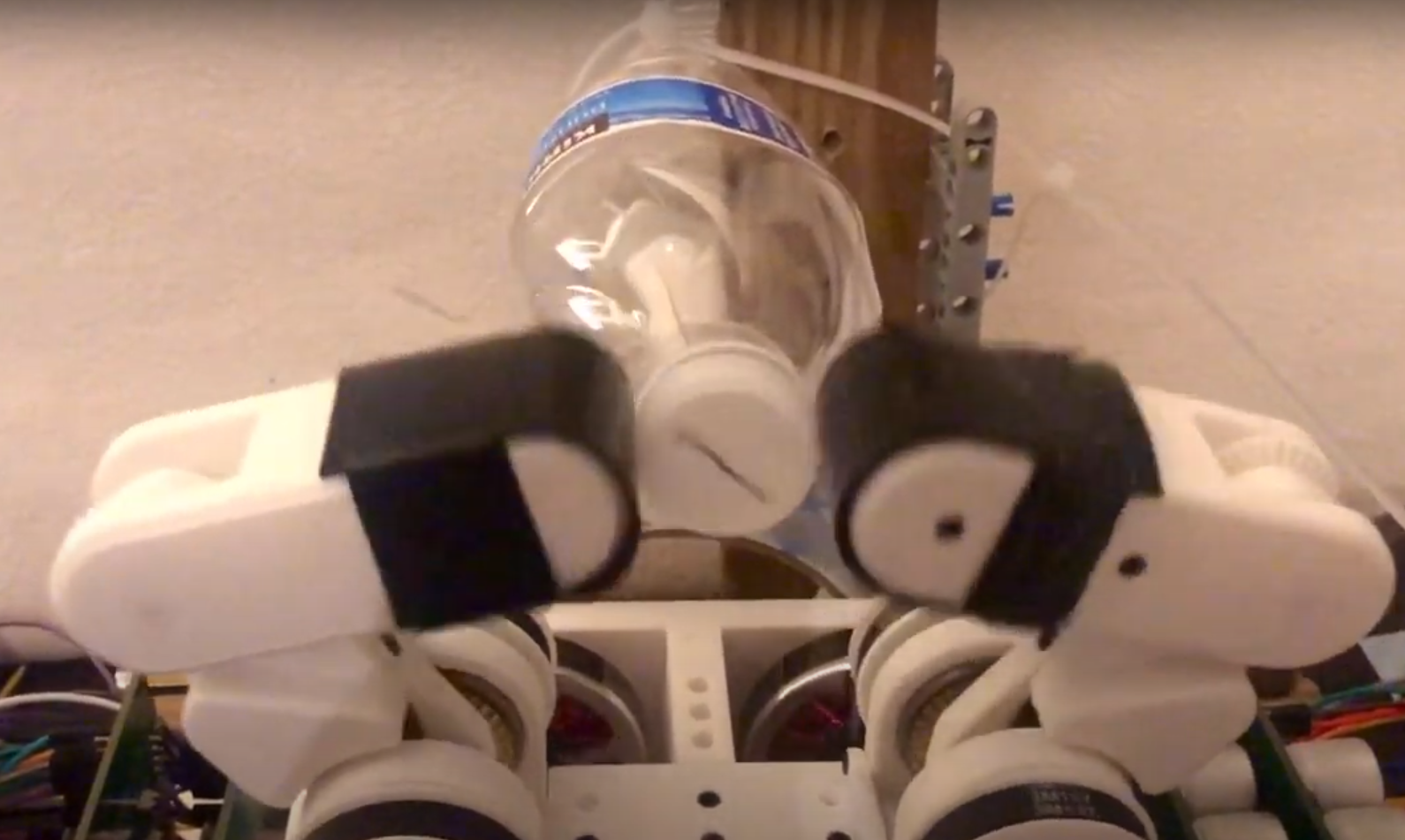}}
\subfloat[]{\includegraphics[height = 3.4cm]{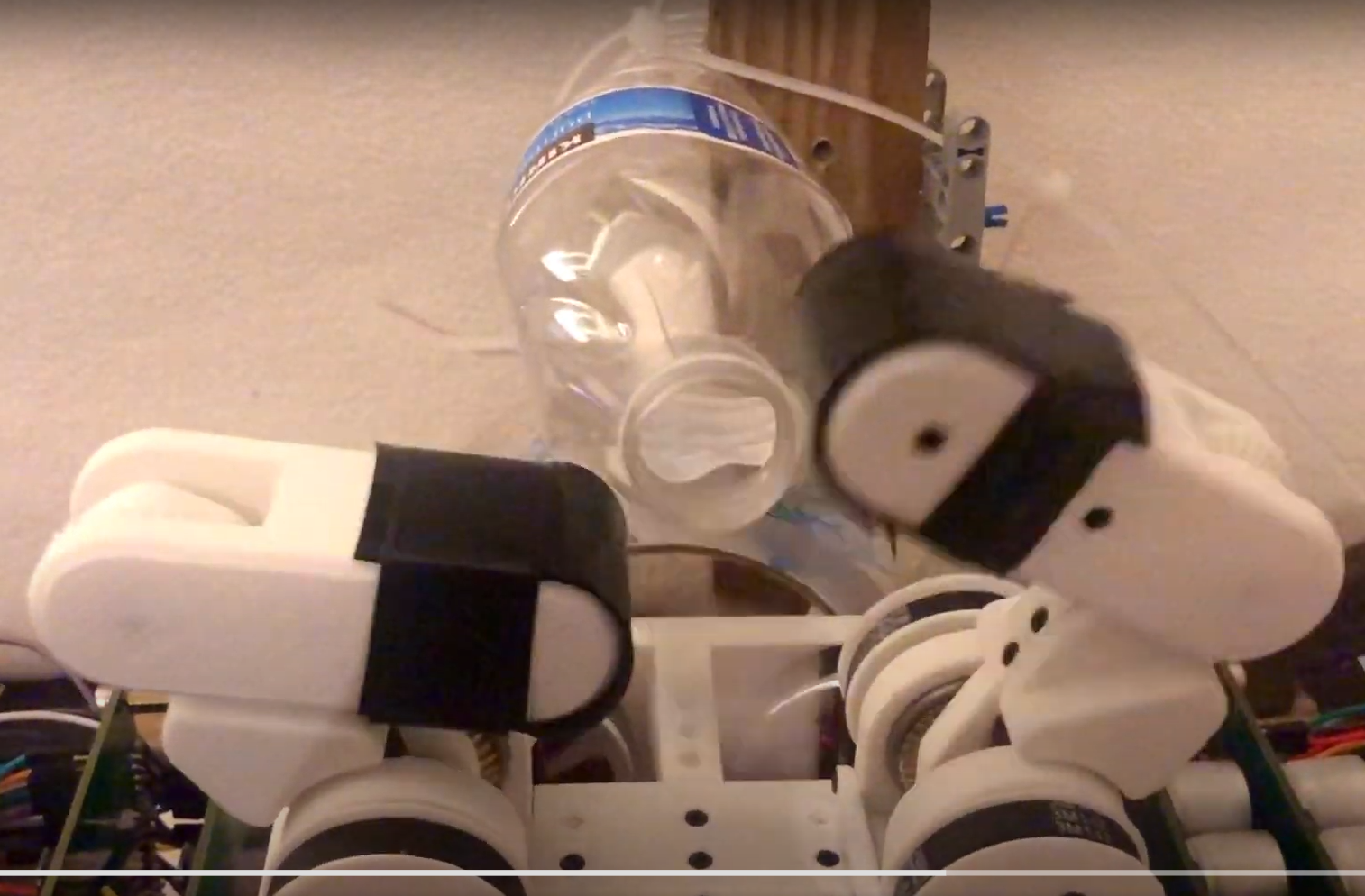}}
\caption{Opening a water bottle.}
\label{fig:water}
\end{figure}

Similarly, we can apply this same motion to the task of opening a plastic water bottle (Fig. \ref{fig:water}). Again, the bottle is constrained about its axis so the hand does not need to support the weight of the bottle. The bottle cap is screwed down about halfway on the bottle, as the fingers are not strong enough to unscrew a fully closed cap. This is something we may wish to improve in the future, by optimizing the selection of impedance parameters. A video of the hand opening a water bottle can be found here: \url{https://youtu.be/NlufcoPd1Ns}.

In recent years, there have been great advancements in dexterous in-hand manipulation leveraging machine learning and artificial intelligence. One example is solving a Rubik's Cube with one hand by researchers at OpenAI \cite{openai2019solving}. While our research features much simpler tasks, this would be a good focus for future work.

\subsection{In-Hand Manipulation Through Interaction with Environment}
\begin{figure}[!htbp]
\centering
\subfloat[]{\includegraphics[height = 4cm]{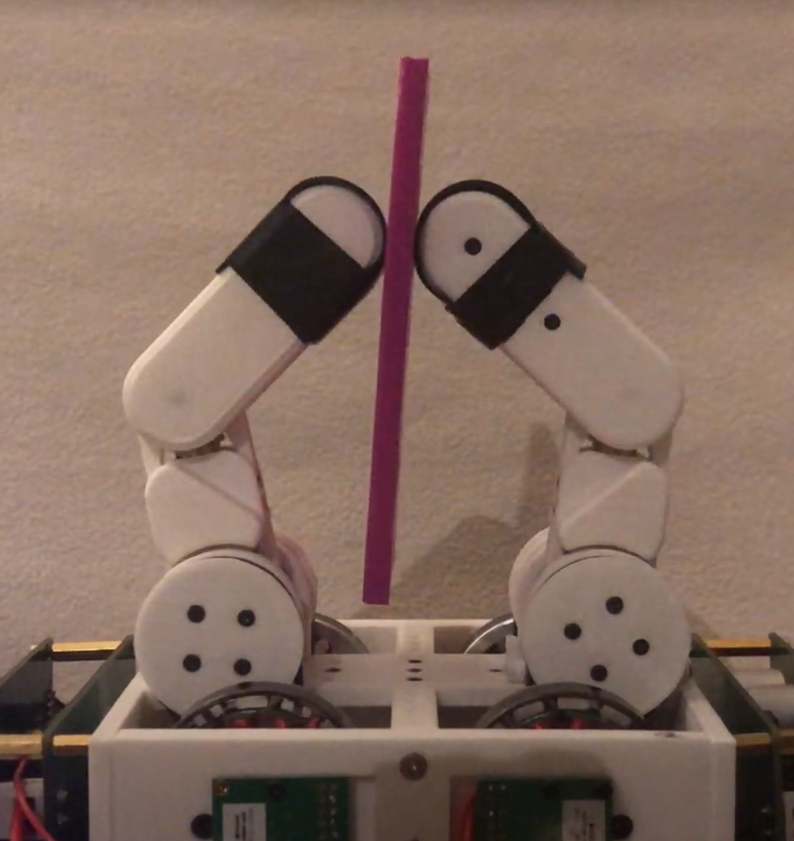}}
\subfloat[]{\includegraphics[height = 4cm]{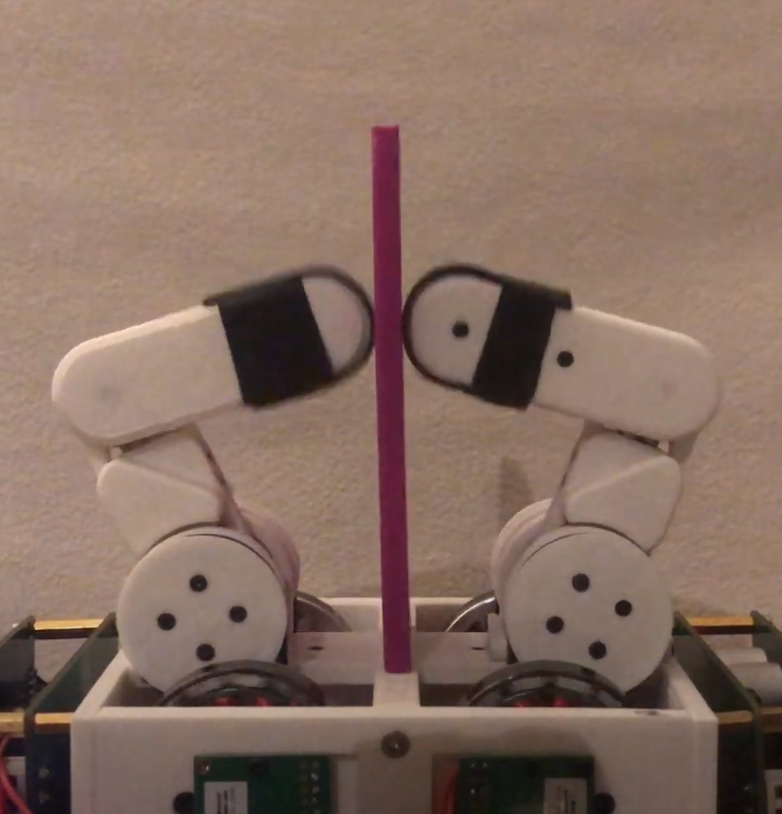}}
\subfloat[]{\includegraphics[height = 4cm]{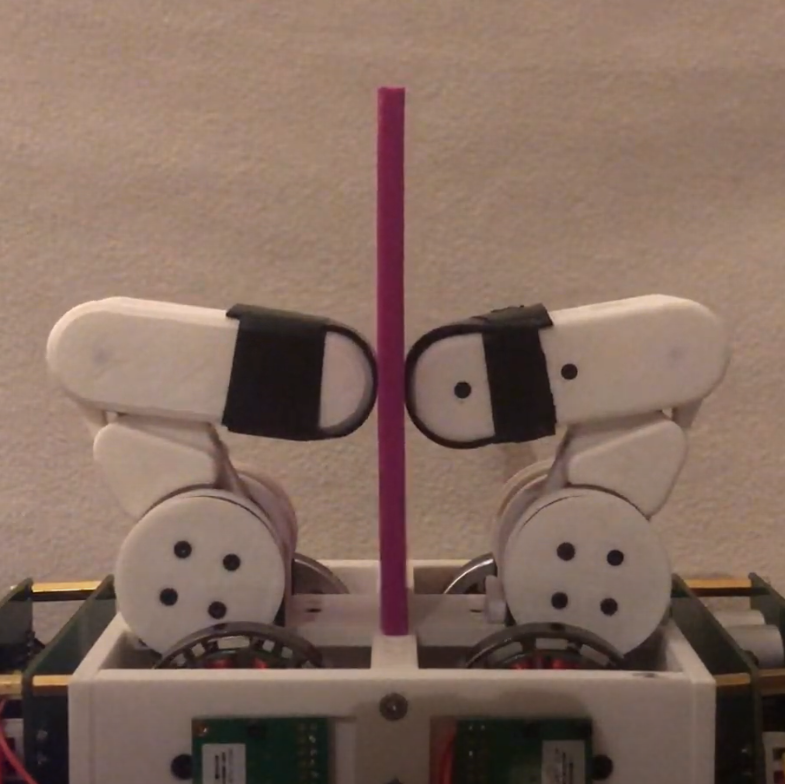}}
\subfloat[]{\includegraphics[height = 4cm]{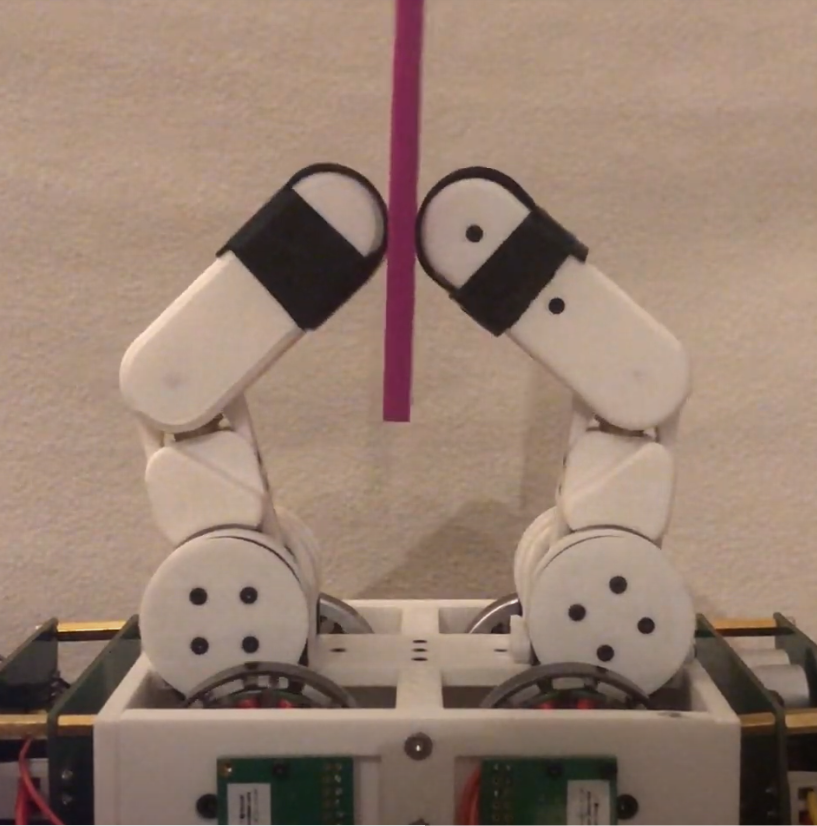}}
\caption{In-hand manipulation through pushing.}
\label{fig:push}
\end{figure}

A number of papers achieve in-hand manipulation with simple grippers by exploiting interaction with the environment, a concept known as ``extrinsic dexterity" \cite{6907062}. For example, Chavan-Dafle \textit{et al.} \cite{7354264} model the contact forces on a grasped object when it is manipulated by pushing it against its environment. Using this method, the object can be re-positioned in the two fingers of the gripper, however, since the gripper only has 1 DOF, it must be connected to a robot arm. Using our hand, we can perform a similar operation, but instead of using contact with the environment, we can exploit contact with the palm of the hand. When grabbing an object in the fingertips and moving them in a trajectory downward, the palm applies a force to the object, pushing the object up in relation to the fingertips as shown in Fig.~\ref{fig:push}. In order to facilitate the sliding contact between the fingertip and the object, the stiffness coefficient is reduced by half in the direction normal to the surface of the object ($X$-direction) while the fingers push the object against the palm. This was simply a heuristic chosen to reduce the frictional force to make sliding easier, while still maintaining a stable grasp. Using gravity, the fingers can reset the object's original position in the hand and the task can be repeated with great reliability as seen in this video:  \url{https://youtu.be/hE44xuGeTuk}.

This same concept was applied in practice to two other objects. The first is a pencil (video: \url{https://youtu.be/eLY_Iq33zP4}), and the second is a playing card (video: \url{https://youtu.be/KgPDP-Wg1zU}). Manipulation of the playing card demonstrates the light touch the fingers are capable of. The hand is able to provide enough force to manipulate the card but does not cause it to bend when it is pushed against the palm of the hand, showing that the QDD hand can handle delicate objects.

\section{Picking up a Coin}
\begin{figure}[!htbp]
\centering
\includegraphics[width = 12 cm]{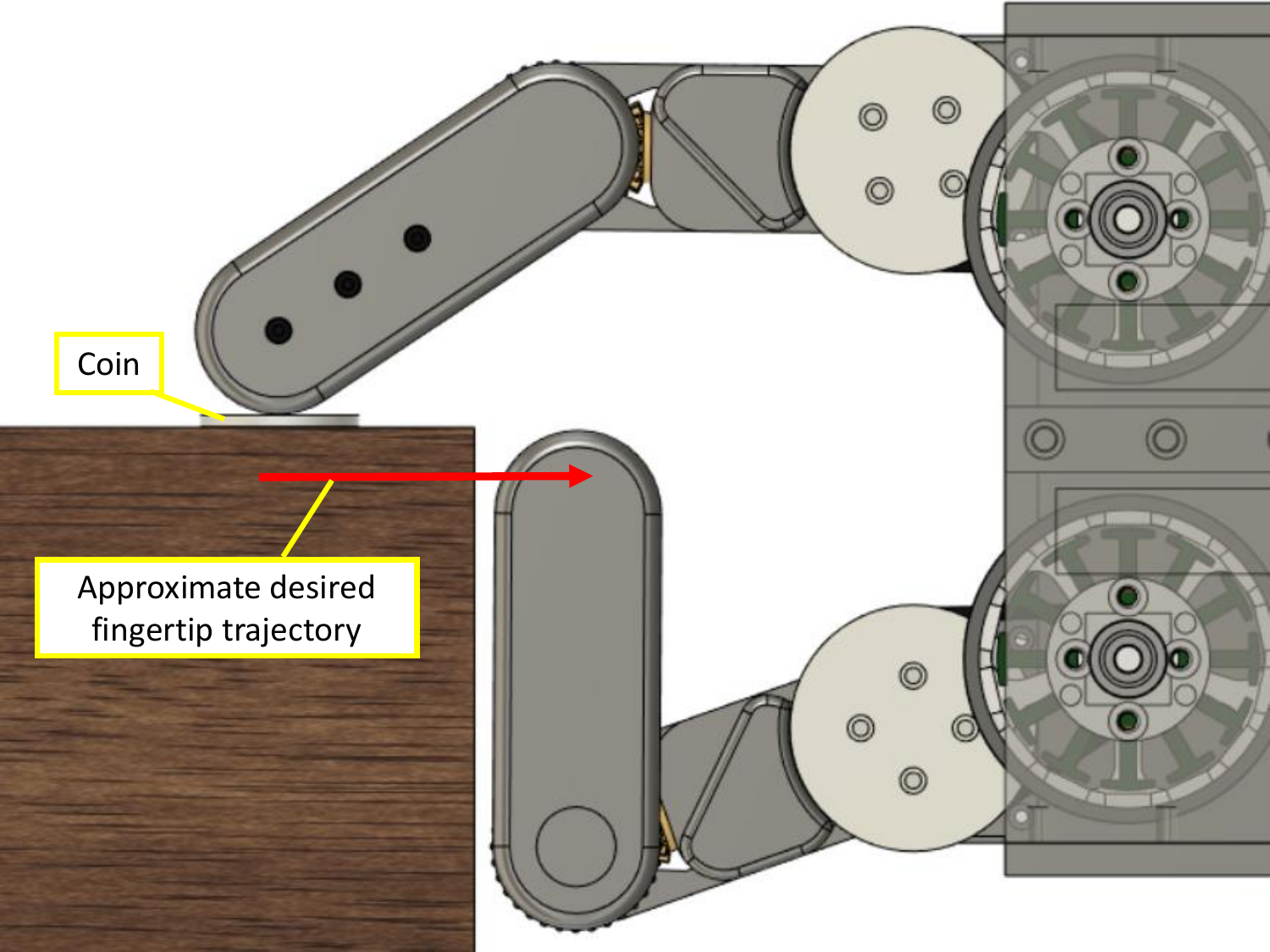}
\caption{Desired trajectory for picking up the coin.}
\label{fig:coin traj}
\end{figure}

\begin{figure}[!htbp]
\centering
\subfloat[]{\includegraphics[height = 4cm]{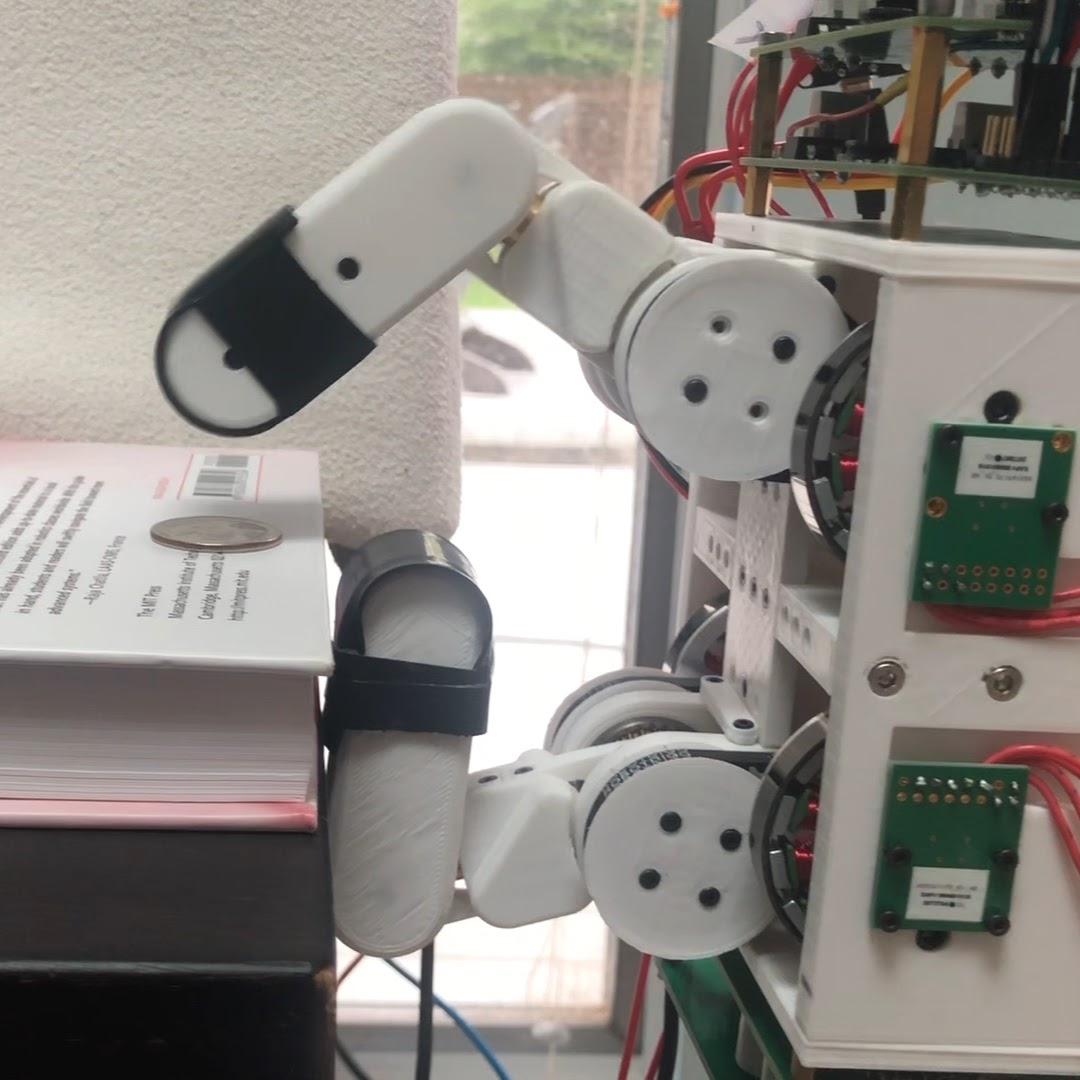}}
\subfloat[]{\includegraphics[height = 4cm]{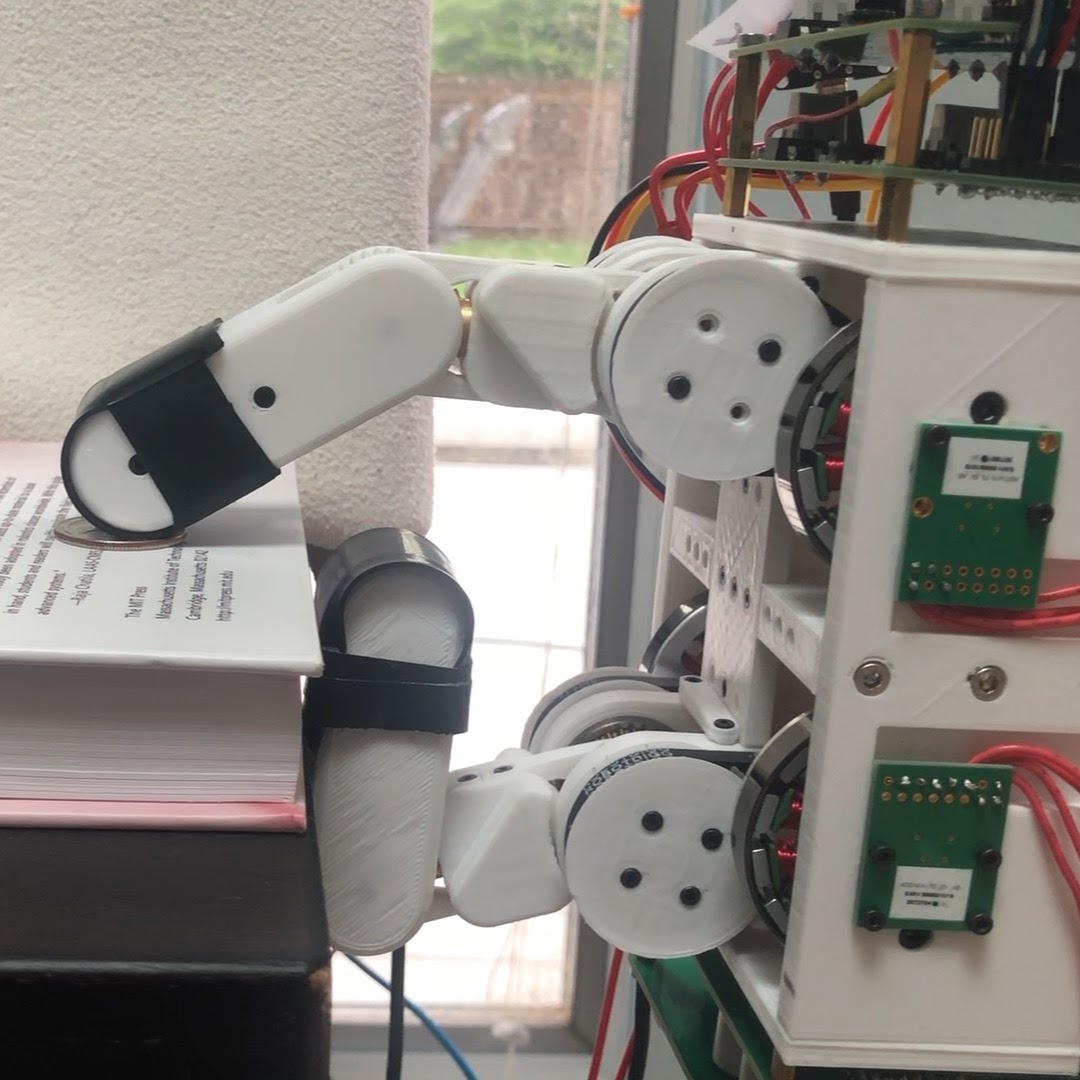}}
\subfloat[]{\includegraphics[height = 4cm]{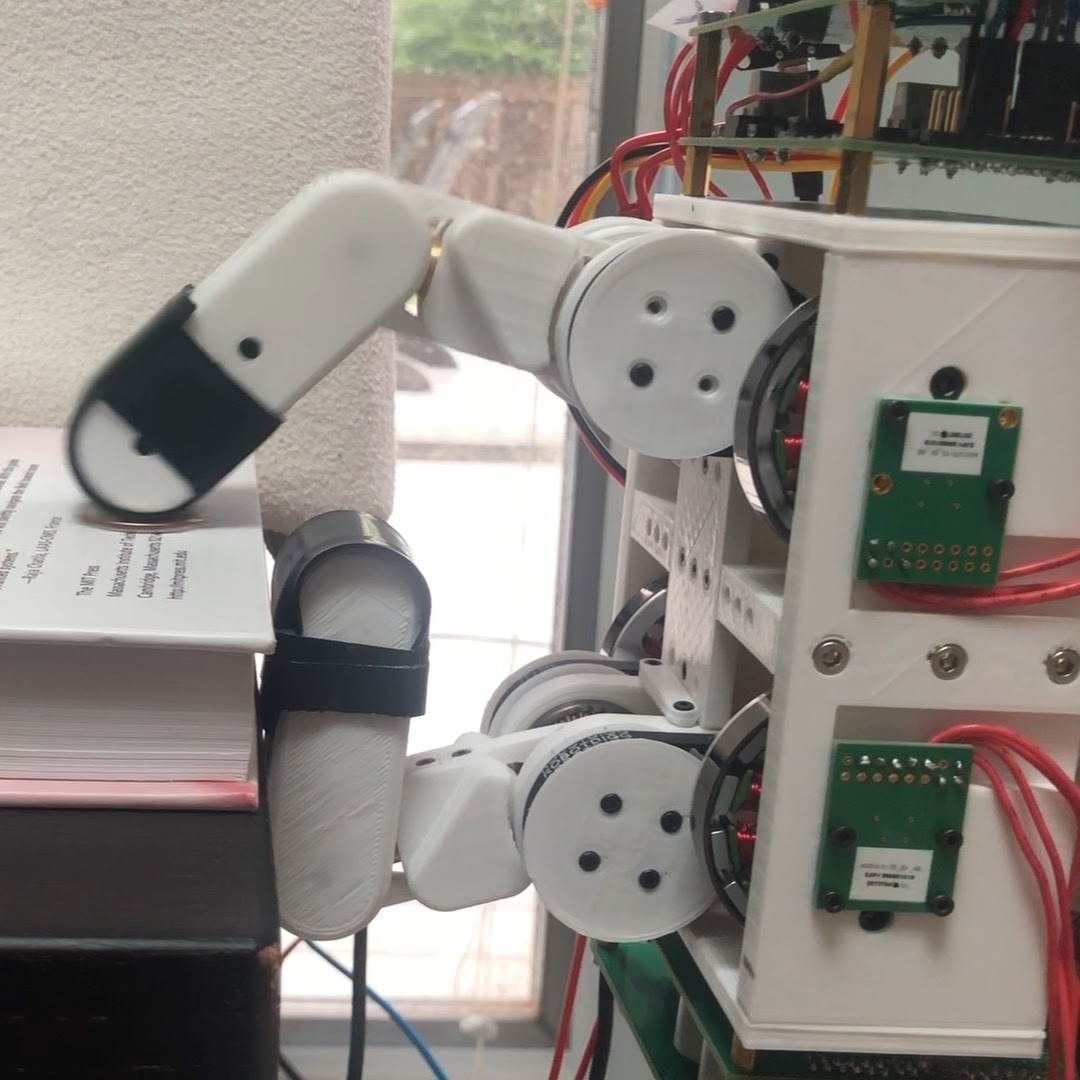}}
\subfloat[]{\includegraphics[height = 4cm]{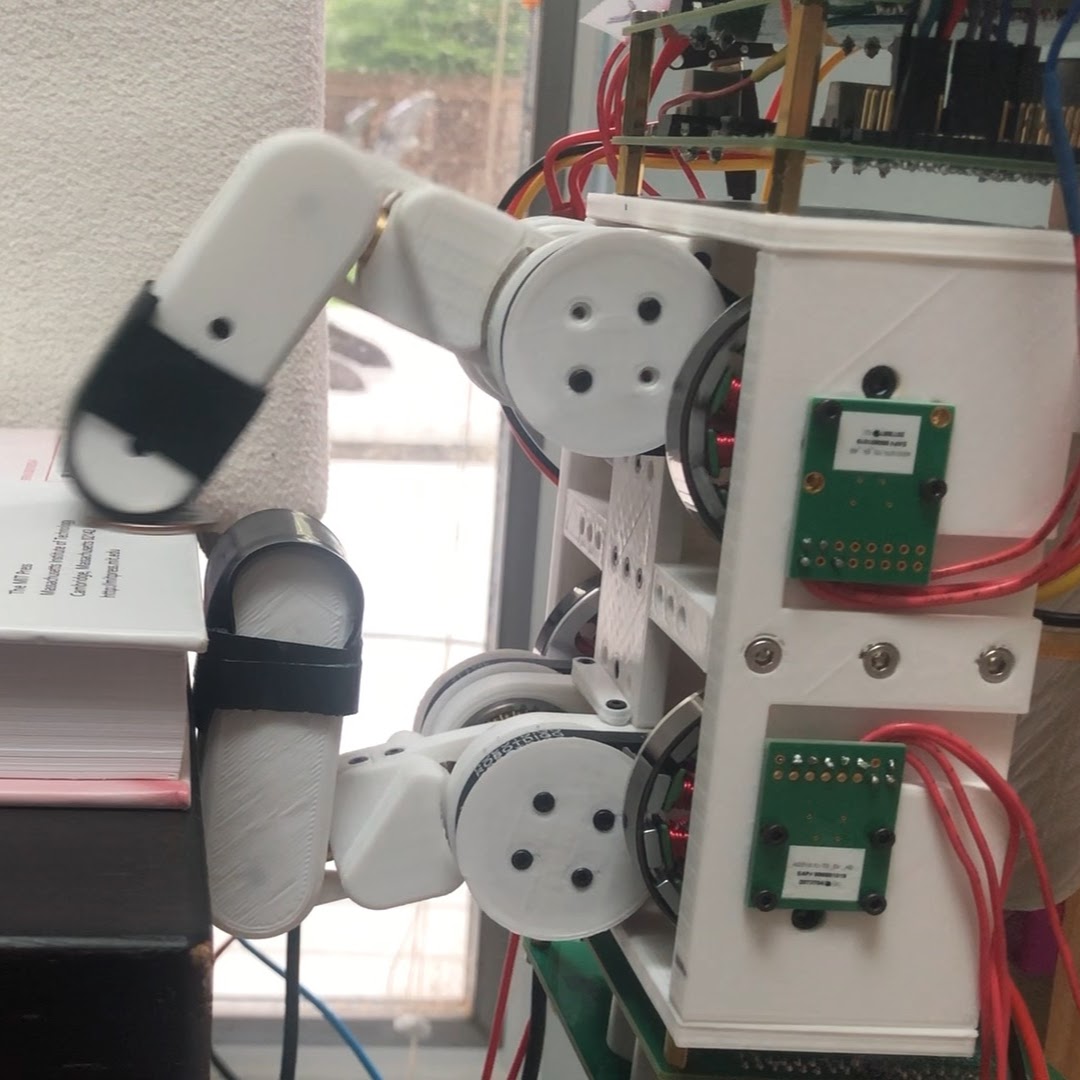}}\\
\subfloat[]{\includegraphics[height = 4cm]{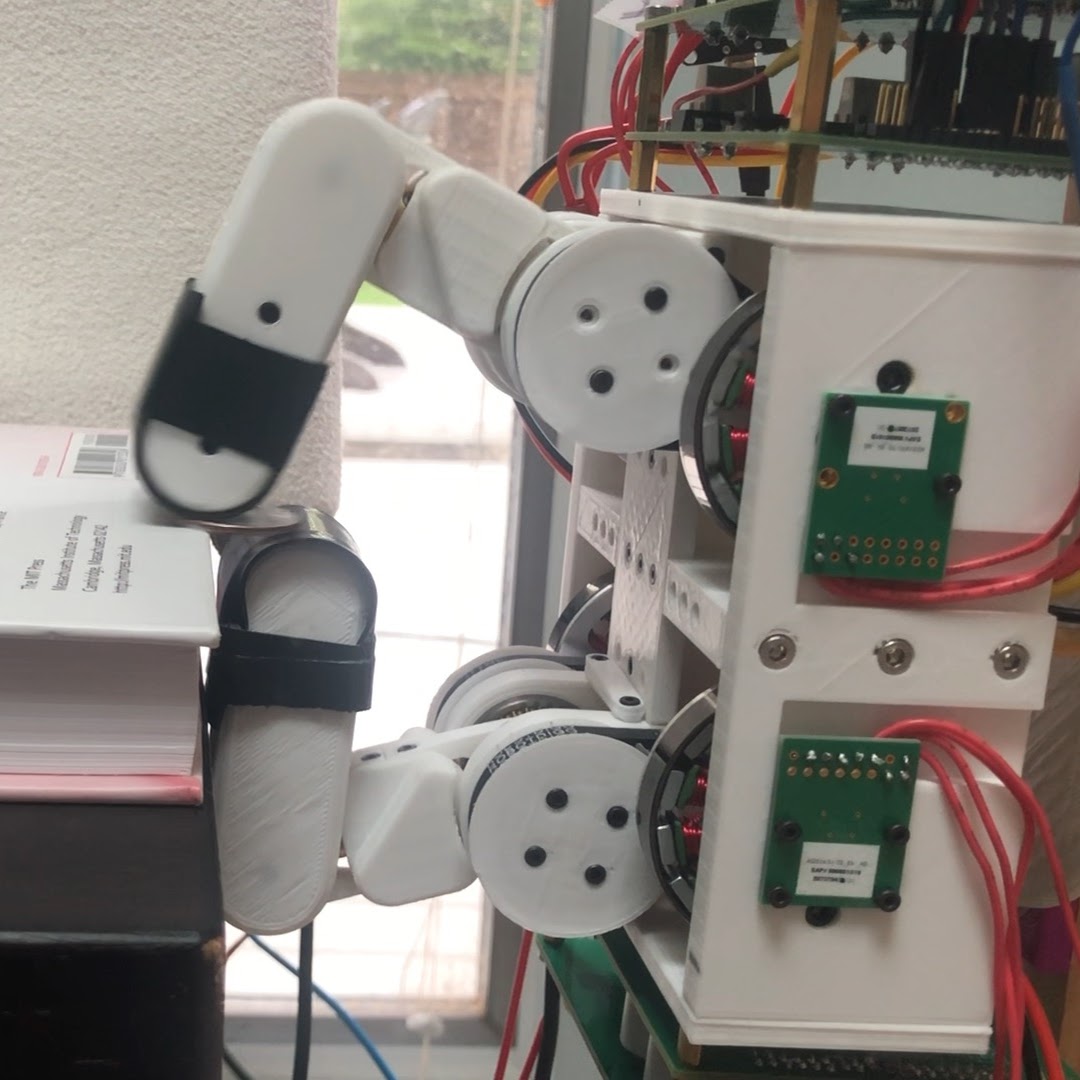}}
\subfloat[]{\includegraphics[height = 4cm]{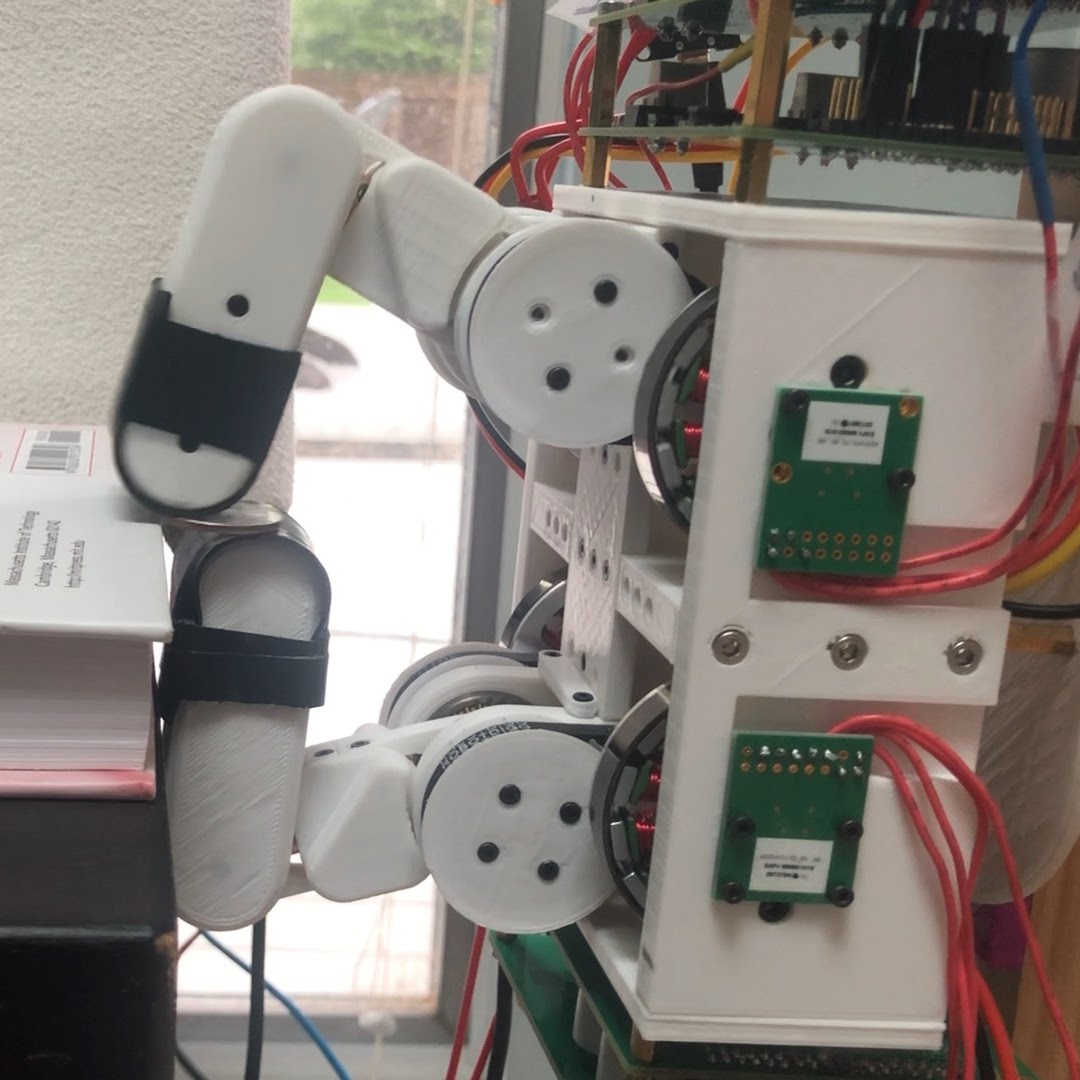}}
\subfloat[]{\includegraphics[height = 4cm]{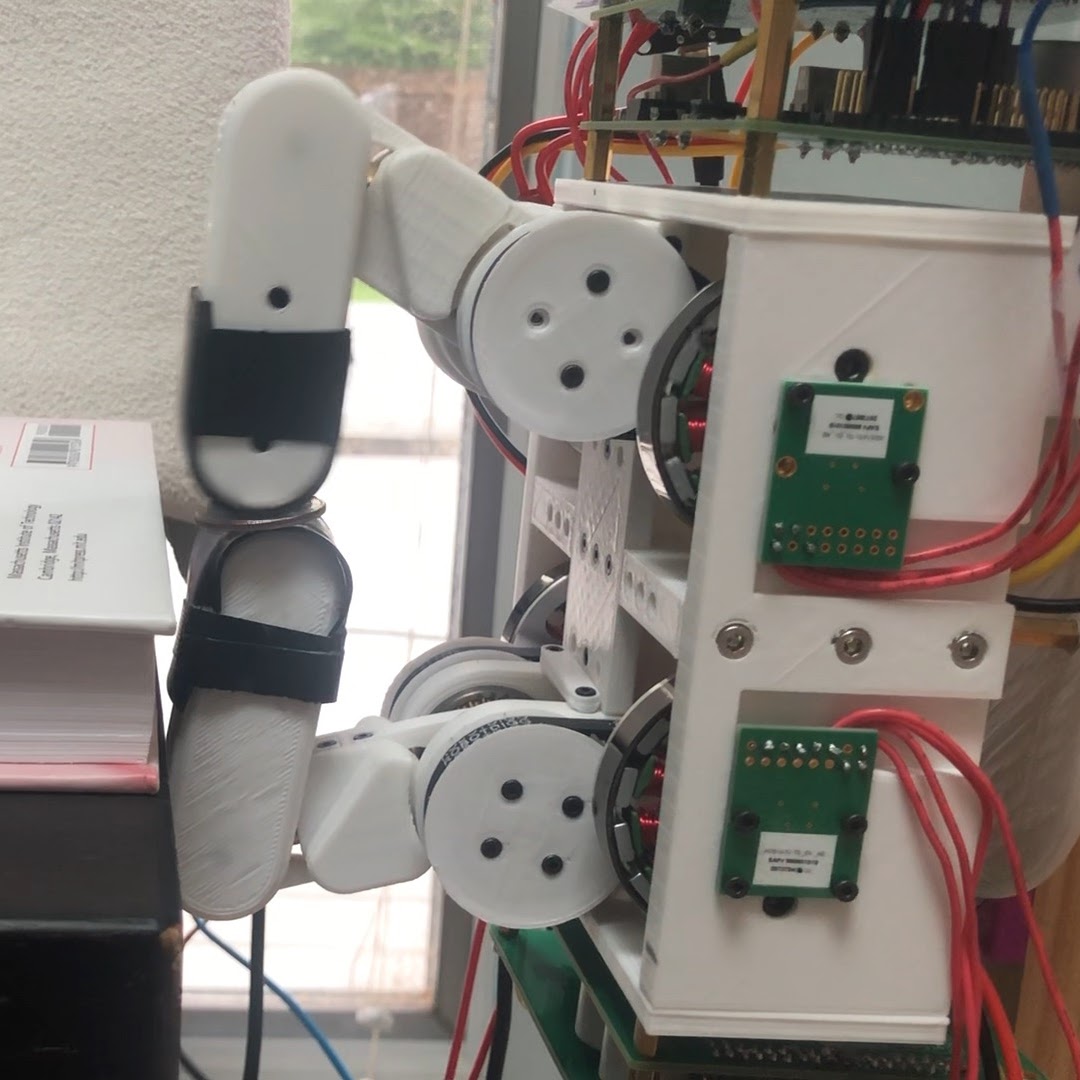}}
\subfloat[]{\includegraphics[height = 4cm]{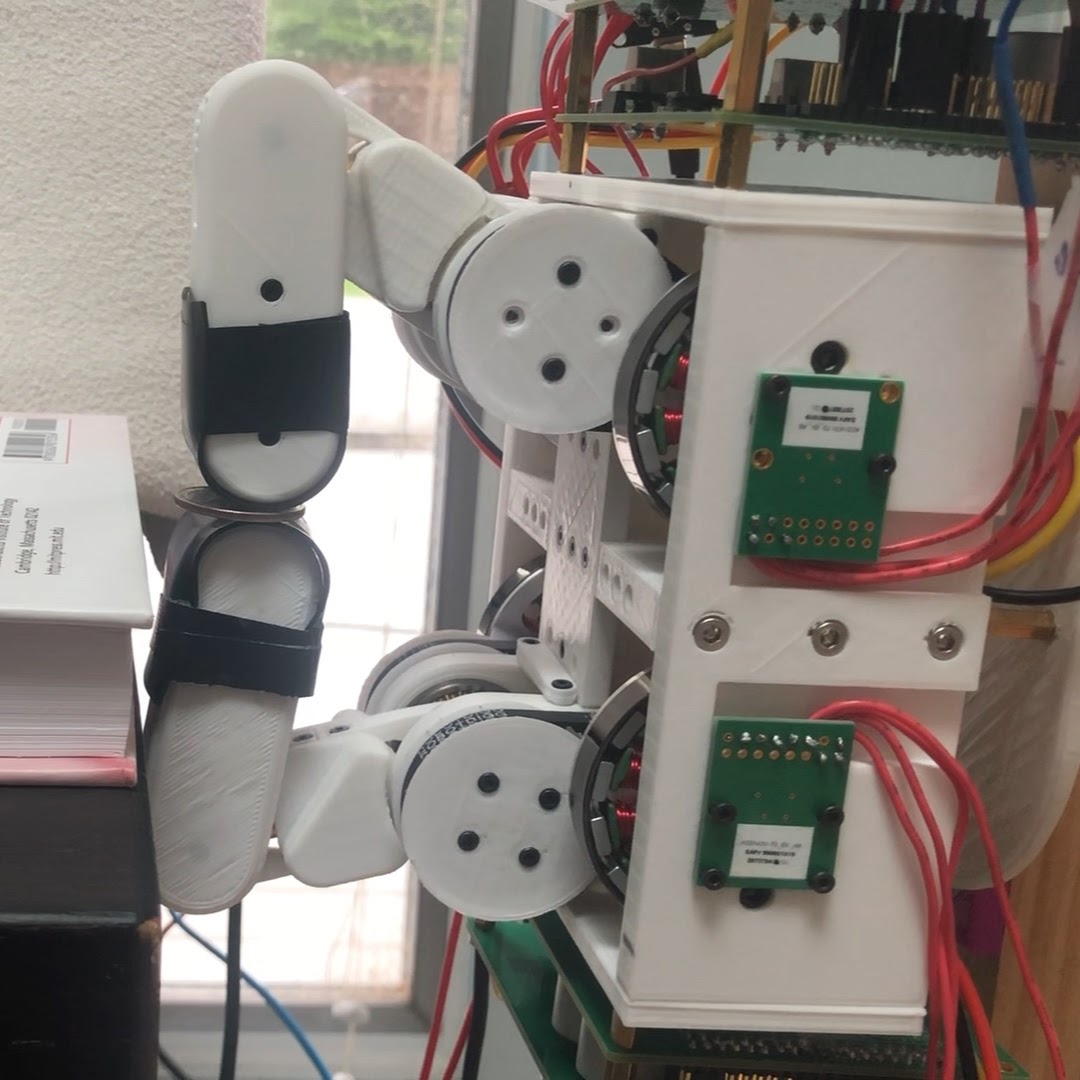}}
\caption{Picking up a coin.}
\label{fig:coin}
\end{figure}

One task robot hands and grippers may struggle with is picking up small flat objects such as a coin from a surface. Due to the geometry of the fingers in our robot hand, it would be very difficult to lift a coin directly from the surface of a table. However, if we position the hand near the edge of the table, the fingers can slide the coin along the surface and into a stable grasp between the fingertips. In order to accomplish this, the finger must apply a force normal to the coin while sliding towards the edge of the table. This force remains constant by maintaining the difference between the actual and desired positions in the direction normal to the table, in our case the $X$ direction. Figure~\ref{fig:coin traj} shows the planned finger trajectory. After reaching the edge of the table, the coin is transferred to the second finger, and the task is complete (Fig. \ref{fig:coin}). A video of the hand successfully picking up a coin can be found here: \url{https://youtu.be/U9F1Om37O4Y}.

\chapter{Conclusion and Future Work}
We believe that this work shows some of the benefits of using quasi-direct-drive actuators with variable impedance control for applications in general-purpose robotic hands and grippers. A transparent, backdrivable gear reduction effectively turns the motor into an additional force sensor which can aid in dexterous manipulation and highly dynamic tasks, while field-oriented control is implemented to accurately control motor torque with current sensing and two PI controllers. We define transparency as the ease to which force and motion can be applied not only from the finger mechanism to the environment, but also from the environment to the finger and actuator. This is essential for tasks such as \textit{Smack-and-snatch} manipulation, and grasping delicate objects with a light touch. Impedance control proves to be a simple, and intuitive way of adjusting motor position and compliance, as only three parameters are required, these being desired position, stiffness coefficient, and damping coefficient. Unlike series-elastic and sensored approaches, quasi-direct-drive requires only current sensing enabling higher bandwidth force control, while having more torque than direct-drive actuators.

Our design might provide inspiration for future direct and quasi-direct drive hand and gripper mechanisms. We showed kinematic analysis for the belt reduction/differential gear system. This method pairs well with the quasi-direct-drive actuators, as joint torque is shared between the two motors, thereby increasing the maximum possible torque at each joint. In this way, we retain some of the benefits of direct-drive actuation while increasing the possible gripping force.

Many robotic hands lack in-hand manipulation capability due to the complex motion planning and control of contact forces required with more traditional methods. Our hand achieves force control with minimal sensing (only current sensing) and simplifies motion planning by using impedance control as the relation between force and position. Perhaps this philosophy might lower the barrier for entry for performing more complicated tasks with robot hands.

While the QDD hand was successful in a lot of ways, it certainly has its limitations. Below is a list of potential design improvements for future iterations of the hand:

\begin{itemize}[noitemsep,topsep=0pt]
  \item Reduce the size of the base structure housing the motors.
  \item Reduce the size of electronic hardware (mainly motor drivers) so they can be integrated into the structure of the hand. Ideally, we would make a custom motor driver with integrated MCU, and encoder to limit the amount of wires needed and reduce the size.
  \item Add a visual perception system, e.g., a camera for object detection/position estimation.
  \item Add a third finger to the hand to increase capability and increase grasp stability.
\end{itemize}

In the future, we plan on mounting the hand to the Franka Emika Panda robot arm \cite{franka}. This will allow us to further test the ability of the hand for more complicated tasks where the hand does not need to be stationary. For example, it would be useful to know the success rate of the \textit{Smack-and-Snatch} manipulation task over the course of many trials with different objects.

The primary limitation of our design is that the torque output of the QDD actuators is much lower than other actuator types like higher gear ratio motors. However, we believe this limitation can be mitigated in a number of ways. Due to the structure of the fingers, they are compliant in the $X-Y$ plane between the fingers, but stiff in the $Z$ direction. This anisotropic stiffness of the fingers could be exploited in motion planning of the robot arm to lower the required gripping force of the fingers. Additionally in the future, we may experiment with better thermal management of the motors to increase their torque output without overheating. This includes active cooling of the motors by using fans, or by supplying the motors with high current when needed but only for a short period of time to ensure the motors do not overheat.

That being said, with further development, we believe impedance-controlled QDD actuators show great promise in enabling more capable robot hands of the future.

{
\begin{singlespace}
\setlength\bibitemsep{2\itemsep}
\printbibliography[heading=bibintoc]
\end{singlespace}
}


\appendix 

\chapter{Video Links}

\begin{table}[hbp!]
\caption{Organized list of all test videos.}
\label{videos}
\centering
\fontsize{10}{14}\selectfont
\begin{tabular}{ |c|c| } 
\hline
 \textbf{Description} & \textbf{Link} \\
\hline
 Unmodified Impedance Controller & \url{https://youtu.be/HuPaaSSRv9Q} \\
 Modified Cartesian Impedance Controller & \url{https://youtu.be/MrbVwZfQA1o} \\
 Grasping a Playing Card & \url{https://youtu.be/jcZhpDPOkac} \\
 Grasping a Tortilla Chip & \url{https://youtu.be/nhsiFj69y4k} \\
 Grasping an Egg & \url{https://youtu.be/4gP3ush-n-A}\\
 Form Closure & \url{https://youtu.be/Sd9FdDblNxo} \\
 Grasp in Response to Disturbance &  \url{https://youtu.be/bYJElUFPpNE} \\
 Smack-And-Snatch Manipulation & \url{https://youtu.be/0FjuOXXyo-8} \\
 Smack-And-Snatch - Egg & \url{https://youtu.be/0b7juuTGyyA}\\
 In-Hand manipulation of rubber ball &  \url{https://youtu.be/NOzzvAztOfY} \\
 In-Hand manipulation - Translation & \url{https://youtu.be/ytpdwM4pvCM}\\
 Turning a Screwdriver & \url{https://youtu.be/B7YEwJ3F6jc}\\
 Opening a Water bottle & \url{https://youtu.be/NlufcoPd1Ns}\\
 In-Hand Manipulation Through Pushing & \url{https://youtu.be/hE44xuGeTuk}\\
 In-Hand Manipulation - Pencil & \url{https://youtu.be/eLY_Iq33zP4}\\
 In-Hand Manipulation - Card & \url{https://youtu.be/KgPDP-Wg1zU}\\
 Picking up a Coin & \url{https://youtu.be/U9F1Om37O4Y}\\
 
 \hline
\end{tabular}
\end{table}

\end{document}